%% file: main.tex
\documentclass[10pt,twocolumn,letterpaper]{article}

\usepackage{cvpr}            % To produce the CAMERA-READY version

\input{preamble}

\definecolor{cvprblue}{rgb}{0.21,0.49,0.74}
\usepackage[pagebackref,breaklinks,colorlinks,citecolor=cvprblue]{hyperref}
\def\paperID{XXXXX} % *** Enter the Paper ID here
\def\confName{CVPR}
\def\confYear{2026}

\title{CARD: A Multi-Modal Automotive Dataset for Dense 3D Reconstruction in Challenging Road Topography}

\author{
Gasser Elazab$^{1,2}$ \quad Frank Neuhaus$^{3}$ \quad Tilman Koß$^{3}$ \quad Malte Splietker$^{3}$ \\
Aditya Date$^{1}$ \quad Michael Unterreiner$^{1}$ \quad Maximilian Jansen$^{1}$ \quad Olaf Hellwich$^{2}$ 
\vspace{0.45em} \\
$^{1}$CARIAD SE \qquad $^{2}$Technische Universit{\"a}t Berlin \qquad $^{3}$Vision \& Robotics GmbH \\
{\tt\small gasser.elazab@cariad.technology} \\
\vspace{0.4em}
\textbf{\url{https://card.content.cariad.digital}}
}
\begin{document}
\maketitle
\input{sec/0_abstract}

\input{sec/1_intro}

\input{sec/2_formatting}

\input{sec/3_finalcopy}

\input{sec/X_suppl}
\clearpage        % <-- force bibliography to start on a new page

{
    \small
    \bibliographystyle{ieeenat_fullname}
    \bibliography{main}
}

\end{document}

%% file: preamble.tex
\usepackage{tabularx,booktabs}

\newcommand{\xmark}{\texttimes}

\usepackage{tikz}
\usepackage{enumitem} % in the preamble
\usepackage{makecell}
\newcommand{\best}[1]{\textbf{#1}}
\newcommand{\secbest}[1]{\underline{#1}}

\usetikzlibrary{arrows.meta,positioning,fit,shapes.geometric,calc}
\usepackage{xcolor}
\definecolor{stepbg}{RGB}{245,247,250}
\usepackage{graphicx}
\usepackage{amsmath}
\usepackage{siunitx}
\usepackage{multirow}
\usetikzlibrary{arrows.meta,positioning,fit,shapes.geometric}
\usepackage{IEEEtrantools}
\usepackage{float}
\newcolumntype{Y}{>{\centering\arraybackslash}X}
\makeatletter
\newenvironment{AutoWideTable}[1][t]{%
  \if@twocolumn\begin{table*}[#1]\else\begin{table}[#1]\fi
  \centering
}{%
  \if@twocolumn\end{table*}\else\end{table}\fi
}
\makeatother

\newcolumntype{Y}{>{\centering\arraybackslash}X}

%% file: sec/0_abstract.tex
\begin{abstract}
Autonomous driving must operate  across diverse surfaces to enable safe mobility. However, most driving datasets are captured on well-paved flat roads. Moreover, recent driving datasets primarily provide sparse LiDAR ground truth for images, which is insufficient for assessing fine-grained geometry in depth estimation and completion. To address these gaps, we introduce CARD, a multi-modal driving dataset that delivers quasi-dense 3D ground truth across continuous sequences rich in speed bumps, potholes, irregular surfaces and off-road segments. Our sensor suite includes synchronized global-shutter stereo cameras, front and rear LiDARs, 6-DoF poses from LiDAR-inertial odometry, per-wheel motion traces, and full calibration. Notably, our multi-LiDAR fusion yields \(\sim 500\text{K}\) valid depth pixels per frame, about \(6.5\times\) more than KITTI Depth Completion and \(10\times\) more on average than other public driving datasets. The dataset spans \(\sim\)110\,km and 4.7\,hours across Germany and Italy. In addition, CARD provides 2D bounding boxes targeting road-topography irregularities, enabling accurate benchmarking for both geometry and perception tasks. Furthermore, we establish a standardized evaluation protocol for road surface irregularities on CARD and benchmark state-of-the-art depth estimation models to provide strong baselines. The CARD dataset is hosted on \url{https://huggingface.co/CARD-Data}.

\end{abstract}

%% file: sec/1_intro.tex
\begin{figure}[t]

\includegraphics[width=\linewidth]{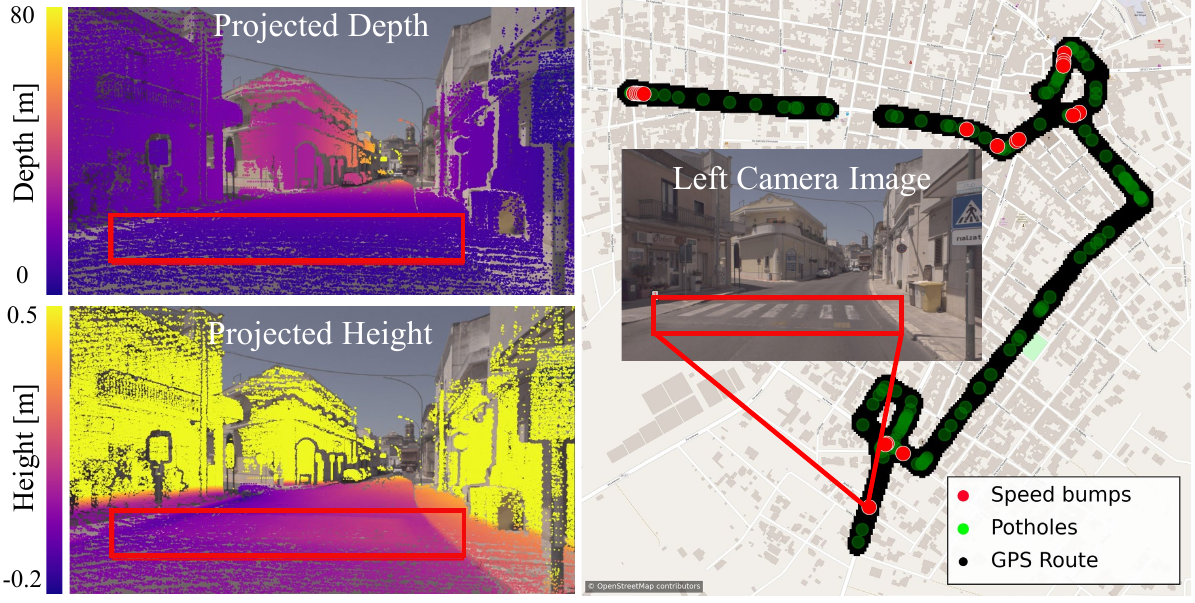}
\caption{CARD example from Carmiano, Italy. Right: map with annotated positive irregularities (speed bumps) and negative irregularities (potholes) along the driven routes, with one speed bump highlighted. Left: camera image with depth ground truth projected into the image (top) and height relative to the road (bottom).}
\label{fig:main_fig}
\end{figure}

\section{Introduction}
\label{sec:intro}

Road accidents are responsible for approximately 1.19 million deaths per year~\cite{WHO2023_GlobalStatusReportRoadSafety}. Several studies~\cite{Pavement_effect_on_road_accident_2022,Review_Classificaiton_of_road_obstacles_2025,impact_roadway_conditions_accident_malaysia_2020} show that poor road surface conditions, such as potholes, bumps, and low skid resistance, are among the major factors in road accidents~\cite{Road_condition_accident_seveirity_2025}. Since one goal of autonomous driving is to reduce crash risk~\cite{deepLearning_for_safe_autonomous_driving}, high-quality datasets that reflect the full diversity of real-world surfaces are required. However, most public autonomous driving datasets were collected on well-paved, largely flat roads, with limited coverage of degraded or irregular topographies, even though such surfaces materially affect perception, planning, and safety~\cite{Comfortable_speedcontorl_based_on_roughness_ref1,Adaptive_lane_change_based_on_road_frictions_ref2}. This gap is particularly critical for vision-centric driving stacks, where reliable depth estimation and scene understanding remain challenging under non-ideal road surface conditions~\cite{bumps_and_potholes_detection_zedCamera_2023}. In addition, it is reported that a primary trigger for driver takeovers from the autonomous driving stack is the system’s inconsistent response to irregular road geometry, such as speed bumps and potholes~\cite{Tesla_disengagement_survey}.

Synthetic data and simulators, such as CARLA~\cite{Dosovitskiy2017CARLA}, AirSim~\cite{Shah2018AirSim}, and Synscapes~\cite{Wrenninge2018Synscapes} help alleviate scarcity, but the sim-to-real gap persists~\cite{zuo2025ralad}, and natural road irregularities are hard to model faithfully. Large-scale driving datasets, such as KITTI~\cite{KITTI_dataset}, Waymo Open~\cite{waymodataset}, Argoverse~\cite{argoverse1,argoverse2}, nuScenes~\cite{caesar2020nuscenes}, ONCE~\cite{ONCE_dataset}, PandaSet~\cite{PANDAset_dataset}, A2D2~\cite{geyer2020a2d2}, and ZOD~\cite{ZODDATASET}, primarily target urban and highway settings with well-paved surfaces and therefore under-represent potholes, speed bumps, irregular roads, and off-road conditions. 

Efforts such as the RSRD dataset~\cite{zhao2023rsrd} partially address this gap by targeting surface irregularities. However, it contains only $\sim$16\,k stereo pairs captured with downward-facing cameras that are atypical of production vehicles. By primarily capturing the road surface and not the broader urban scene, RSRD offers limited support for training or evaluating forward-looking perception at scale. Similarly, the TartanDrive dataset~\cite{TartanDrive} provides multi-modal recordings of off-road terrain, yet all sequences were collected with an all-terrain vehicle in a single location, leading to limited scene diversity and a focus on optimizing for that specific off-road environment. Moreover, aside from densified datasets, such as DrivingStereo~\cite{DrivingStereo} and KITTI~\cite{KITTI_dataset,uhrig2017sparsity}, most benchmarks provide sparse LiDAR as depth ground truth, limiting fine-scale geometric evaluation.
%Moreover, only 2.8\,k stereo pairs include dense 3D ground truth.

Addressing these limitations, we introduce CARD, a complementary multi-modal dataset with quasi-dense 3D ground truth that extends existing autonomous driving benchmarks to include diverse road topographies. Unlike prior datasets, which were primarily collected in limited or relatively flat environments, CARD was deliberately recorded to capture urban, suburban, and rural roads under diverse, non-flat conditions, spanning 12 cities across Germany and 9 across southern Italy. This makes CARD particularly suitable for training and evaluating models in safety-critical scenarios where flat road assumptions no longer hold. Finally, CARD is released under two licensing scheme: sequences recorded in Germany are distributed under CC BY 4.0~\cite{CCBY4}, which requires attribution and permits commercial use for broad academic and industrial reuse, while sequences recorded in Italy are distributed under CC BY-NC 4.0~\cite{CCBYNC4} for non-commercial use in compliance with local restrictions. The main novel aspects of CARD are:

\begin{itemize}

\item \textbf{Road topography driving dataset.}
We target the 3D geometry of road surfaces, such as speed bumps, potholes, uneven and off-road segments, providing a dedicated benchmark for road topography in driving scenes.

\item \textbf{Quasi-dense 3D ground truth.}  
We provide $\sim500$\,k fused depth points per frame, corresponding to $\sim18\%$ coverage of the scene-relevant structure in the image.

\item \textbf{Rich multi-modal sequences.}  
We offer continuous multi-sensor sequences (from $\sim1$ to $10$ minutes) with $6$ DoF per-sensor poses, sensor–road calibration, bounding-box labels for road irregularities, and per-wheel trajectories capturing the road-excitation profile.

\end{itemize}

%% file: sec/2_formatting.tex
\begin{table*}[t]
\centering
\scriptsize
\setlength{\tabcolsep}{4pt}
\renewcommand{\arraystretch}{1.15}
\resizebox{\textwidth}{!}{%
\begin{tabular}{|l|c|c|c|c|c|c|c|c|c|}
\hline
\textbf{Dataset} & \textbf{Cities} & \textbf{Hours} & \textbf{Vision Stack} & \textbf{Average GT pts/cam} & \textbf{OR} & \textbf{SP} & \textbf{UE} & \textbf{Urban} \\
\hline
% KITTI~\cite{KITTI_dataset}
% & 1 city (DE) & 1.5  & 2 front-facing cams & 20K* & \xmark & L & L & \checkmark\\
% \hline
KITTI-DC~\cite{KITTI_dataset,uhrig2017sparsity}
& 1 city (DE) & 1.5  & 2 front-facing cams & 75K & \xmark & L & L & \checkmark \\
\hline
Waymo~\cite{waymodataset}
& 3{+} cities (USA) & 11.3  & 5 surround cams (360\textdegree) & 24K* & \xmark & L & L & \checkmark \\
\hline
DrivingStereo~\cite{DrivingStereo}
& N/A (CH) & 5.1 & 2 front-facing cams & 75K & \xmark & L & L & \checkmark \\
\hline
nuScenes~\cite{caesar2020nuscenes}
& 2 cities (US/SG) & 5.5  & 6 surround cams (360\textdegree) & 7.6k & \xmark & L & L & \checkmark \\
\hline
ZOD Sequences~\cite{ZODDATASET}
& 6 cities (EU) & 8.2& 6 surround cams (360\textdegree) & 45K* & \xmark & L & L & \checkmark \\
\hline
A2D2~\cite{geyer2020a2d2}
& 3 cities (DE) & $\sim$ 9.3  & 6 cams (360\textdegree) & 15K* & \xmark & \xmark & L & \checkmark \\
\hline
RSRD-Dense~\cite{zhao2023rsrd}
& 1 city (CH) & $\leq$ 1  & 2 down-facing cams & 90K* & \xmark & \checkmark & \checkmark & L \\
\hline
DDAD~\cite{PackNet_DDAD}
& 5 cities (US) / 2 cities (JP) & $\leq 1$ & 6 surround cams (360\textdegree) & 38K* & \xmark & L & L & \checkmark \\
\hline
Argoverse\,2~\cite{argoverse2}
& 6 cities (US) & 4.2  & 7 surround cams (360\textdegree) & 15K* & \xmark & L & L & \checkmark \\
\hline
ONCE~\cite{ONCE_dataset}
& N/A (CH) & 144  & 7 surround cams (360\textdegree) & 12K* & \xmark & L & L & \checkmark \\
\hline
% PandaSet~\cite{PANDAset_dataset}
% & 2 routes (US) & $\leq 1$  & 6 surround cams (360\textdegree) & 15K* & \xmark & L & L & \checkmark \\
% \hline
TartanDrive\,2.0~\cite{TartanDrive}
& 1 site (US) & 7  & 2 front-facing cams & 34K* & \checkmark & L & \checkmark & \xmark \\
\hline
CARD~(ours)
& 12 cities (DE) / 9 cities (IT) & 4.7 & 2 front-facing cams & \textbf{500K} & \checkmark & \checkmark & \checkmark  & \checkmark \\
\hline
\end{tabular}%
}
\caption{Datasets comparison. Entries marked with * indicate values we averaged over subsamples of the respective datasets. Abbreviations: OR=off-road, SP=speed bumps/potholes, UE=uneven roads, Urban=city scenes. L denotes low or not explicitly reported.}

\label{tab:dataset_comparison_1}
\end{table*}

\section{Related work}
\label{sec:related_work}

Over the past decade, progress in autonomous driving has been driven largely by large-scale, multi-sensor datasets emphasizing scene diversity, geographic coverage, and richly annotated traffic actors~\cite{AD_survey_2024}. These datasets have underpinned breakthroughs in 3D object detection, tracking, segmentation, scene understanding, motion forecasting, and end-to-end driving policies \cite{caesar2020nuscenes, argoverse2, ZODDATASET, ONCE_dataset, PANDAset_dataset, geyer2020a2d2,KITTI_dataset}. In parallel, synthetic and simulated environments~\cite{Shah2018AirSim,CARLA_RARE_OBJECT_DETECTION} have played a complementary role, enabling edge-case generation, domain randomization, and scalable data augmentation, which are critical for robust model training beyond the real-world data~\cite{Dosovitskiy2017CARLA, Shah2018AirSim, Wrenninge2018Synscapes}. Since the sim-to-real gap remains a major challenge~\cite{sim-2-real_for_testing_autonomousdriving}, synthetic data still fails to capture the full complexity of real-world textures, dynamics, and lighting conditions~\cite{sim-2-real_for_testing_autonomousdriving}. Consequently, large-scale real-world datasets remain essential for developing and validating perception models under real driving conditions. To provide a structured overview, we summarize and compare the key properties of CARD and existing datasets in~\Cref{tab:dataset_comparison_1}.

The KITTI dataset remains one of the most influential multi-modal autonomous driving datasets, offering stereo, LiDAR, and high-precision GNSS/IMU, catalyzing progress in 3D object detection, tracking, visual odometry, and depth estimation \cite{KITTI_dataset}. Its depth benchmarks, based on sparse and semi-dense supervision with standardized protocols and an active leaderboard, have become the standard baselines for depth estimation. In addition, it provides densely annotated depth images for depth completion evaluation~\cite{uhrig2017sparsity}, with one of the highest ground truth densities among outdoor datasets. Despite its lasting impact, KITTI’s single-city coverage limits its geographic and topographic diversity, motivating larger-scale datasets with denser geometry and broader environmental variation. Consequently, DrivingStereo~\cite{DrivingStereo} emerged as a subsequent benchmark that adopts a similar densification approach by fusing multiple LiDAR steps and using a stereo network to obtain semi-dense disparity labels, scaling to over 180\,k images across diverse scenarios and weather conditions.

% In addition, they provided densely annotated depth images for depth completion evaluation~\cite{uhrig2017sparsity}, establishing semi-dense ground truth via multi-scan accumulation that remains, to the best of our knowledge, among the densest real-world outdoor ground truths. Subsequent benchmarks such as Argoverse Stereo, DrivingStereo, and ApolloScape also provide denser-than-single-sweep supervision through temporal LiDAR fusion or rendered maps~\cite{argoverseStereo,drivingStereo,apolloscape}. Despite its lasting impact, KITTI’s single-city coverage limits its geographic and topographic diversity, motivating larger-scale datasets with denser geometry and broader environmental variation.

Subsequent autonomous driving datasets further expanded scale and sensing coverage, such as ApolloScape~\cite{Apolloscape_dataset_2019} and H3D~\cite{H3D_dataset}, which provide large-scale multi-sensor urban driving sequences. In addition, nuScenes~\cite{caesar2020nuscenes} was among the first to integrate a complete autonomous vehicle sensor suite, incorporating six cameras, five radars, and a roof-mounted LiDAR, with rich annotations, temporal continuity, and semantic road maps, setting a new standard for multi-sensor fusion and tracking. Moreover, the Waymo Open Dataset~\cite{waymodataset} scaled up coverage across multiple U.S. cities, combining dense labeling with long-range sensing to enable robust perception in diverse urban and suburban settings. Moreover, Argoverse~\cite{argoverse1, argoverse2} added rich long-range annotations and HD map priors, bridging perception and motion forecasting.

Building on these foundations, subsequent datasets began to emphasize industrial-grade realism and large-scale diversity. ONCE~\cite{ONCE_dataset} broadened scene variation through over one million frames of labeled and unlabeled data. PandaSet~\cite{PANDAset_dataset} introduced the dual-LiDAR, production-grade multimodal dataset with high-quality semantic annotations, while A2D2~\cite{geyer2020a2d2} contributed densely annotated camera–LiDAR data collected under varied real-world conditions across Germany, together enriching sensor diversity and annotation fidelity in the AD landscape. Finally, the Zenseact Open Dataset (ZOD)~\cite{ZODDATASET} extended this large-scale paradigm across Europe, emphasizing fine-grained annotations for traffic signs, lane geometry, and adverse weather. As of its release, it remains one of Europe’s largest and most geographically diverse public datasets.

The remarkable progress of large-scale autonomous driving datasets remains largely centered on well-paved urban roads, leaving challenging surface conditions underrepresented. From a systems perspective, several works have emphasized the importance of speed- and friction-aware control for maintaining stability and safety on rough pavements~\cite{Comfortable_speedcontorl_based_on_roughness_ref1, Adaptive_lane_change_based_on_road_frictions_ref2}. While such surface irregularities may occasionally appear in existing datasets, they are typically unlabeled or hard to retrieve, which limits their usefulness. Only a few datasets explicitly target road-surface geometry. Most efforts, such as pothole and crack detection sets captured only with cameras~\cite{SHREC_pothole_dataset1,RDD_pothole_dataset2}, focus primarily on classification and lack geometric fidelity. As an example, TartanDrive 2.0~\cite{TartanDrive} contributes valuable off-road sensing and a robust experimental platform for self-supervised learning, but its coverage is limited to repeated off-road tracks using an all-terrain vehicle. In addition, the RSRD dataset~\cite{zhao2023rsrd} represents a major step forward by providing semi-dense 3D ground truth of road surfaces with annotated bumps and potholes, establishing the first benchmark explicitly dedicated to road-surface reconstruction. With approximately 16\,k stereo pairs, it offers high-quality, curated examples that are particularly valuable for algorithm prototyping and controlled evaluation of surface-aware perception. However, its downward-facing viewpoint captures only the road surface itself, lacking the surrounding scene elements, such as vehicles, pedestrians, and roadside context, typically visible in forward-looking autonomous driving cameras. This narrow field of view, combined with its limited spatial scale, constrains generalization to diverse urban environments.

In contrast to large urban datasets that prioritize actors and traffic structure~\cite{caesar2020nuscenes, waymodataset, argoverse1, argoverse2, ONCE_dataset, PANDAset_dataset, geyer2020a2d2, ZODDATASET} and to specialized topography corpora constrained in scale~\cite{TartanDrive, zhao2023rsrd}, our work targets the missing regime: dense, real-world 3D road-surface geometry captured across diverse urban, suburban, and off-road segments, enabling rigorous evaluation and training of perception and planning under non-flat conditions. By pairing high-density depth with standard forward-looking sensors and topography-oriented labels, CARD complements existing benchmarks and provides the supervision that modern geometric estimation models increasingly require \cite{Ranftl2021MiDaS, DepthAnything2024, DepthAnythingV2_2025, Metric3D_2023, Metric3D_v2, UniDepth2024, UniDepth2025_V2, MoGe2025, MoGE_V2}.

%% file: sec/3_finalcopy.tex
\section{CARD Dataset}
\label{sec:card_dataset}

\begin{figure}[t]
  \centering
  \includegraphics[width=0.95\linewidth]{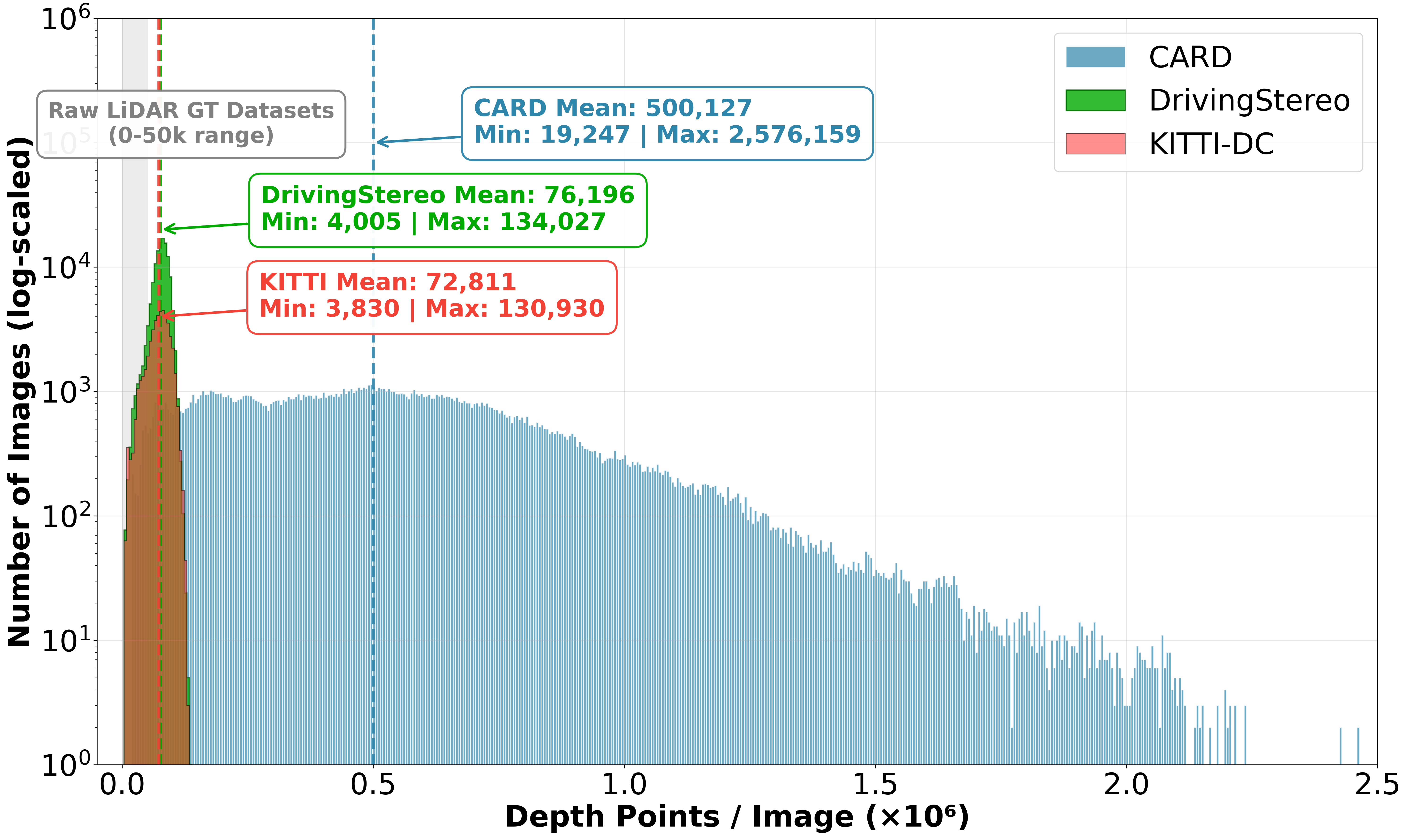}
  \caption{Ground truth points per image. CARD has more depth points per image than KITTI-DC~\cite{uhrig2017sparsity,KITTI_dataset} and DrivingStereo~\cite{DrivingStereo}.}
  \label{fig:KITTI_vs_CARD}
\end{figure}

CARD is a multi-modal driving dataset for perception tasks such as depth estimation and 3D understanding from images, focusing on underrepresented irregular road surface topography. Our sensor suite is designed for road level geometric fidelity with calibration tied to wheel contact points. To our knowledge, CARD provides the highest number of ground truth depth points per image among public driving datasets, as shown in~\cref{fig:KITTI_vs_CARD}, exceeding the semi-dense aggregated ground truth of KITTI-DC~\cite{uhrig2017sparsity} and the aggregated ground truth of DrivingStereo~\cite{DrivingStereo}.

\subsection{Sensor Setup}
\label{subsec:sensor_setup}

We record our dataset using a Porsche E-Macan (2024) with adaptive air suspension to stabilize chassis height across payload changes, reducing ride height drift that can corrupt sensor to road extrinsics, as reported in prior work on real-world datasets~\cite{koledic2024gvdepth}. \cref{fig:car_diagram} shows an example of sensor placement measurements for the depicted capture, and the sensor details are presented in~\cref{tab:sensor_rig_overview}.
Positions are approximate, as the precise calibrated pose of each sensor is provided per sequence in the released metadata.

\noindent\textbf{Per-sequence deliverables (10\,Hz).}
The entire rig is referenced to a right-handed East North Up (ENU) world frame with origin at the IMU beneath the front LiDAR \((X\) east right, \(Y\) north forward, \(Z\) up\()\). Each sequence is time synchronized and includes:

\begin{enumerate}[leftmargin=0pt, itemindent=*, itemsep=2pt, topsep=2pt]
  \item \textbf{Stereo RGB images} (\(2248\times2252\), \texttt{.jpg}). Left and right images are provided in their original distorted format.
  \item \textbf{Raw LiDAR scans} (two XT32 units, \texttt{.laz}). Scans are provided for each timestamp. Sensor-to-base and base-to-world transforms (trajectory) are provided in the calibration and pose bundles to allow global registration.

  \item \textbf{Calibration bundle} (\texttt{.json}). As explained in~\cref{subsec:calibration}, we provide camera intrinsics and extrinsics between all sensors, as well as sensor-to-vehicle transforms for the four wheel centers and the four wheel ground contact frames.

  \item \textbf{Base poses} (\texttt{.json}). As explained in~\cref{subsec:imu-lidar_fusion}, for each timestamp we provide 6-DoF world to base poses from LiDAR inertial fusion in a right handed ENU world frame with origin at the IMU under the front LiDAR.
  
  \item \textbf{3D quasi-dense ground truth (GT)} (\texttt{.npz}). As explained in~\cref{subsec:gT_processing}, for each left image we provide an array of shape \((N,3)\) in the left camera frame.
    
  \item \textbf{Wheel excitation} (\texttt{.json}). As explained in~\cref{subsec:wheel_trajectory}, we provide time-series contact-point trajectories \([t, x, y, z, q_w, q_x, q_y, q_z]\) in the world frame.
  
  \item \textbf{Image annotations} (\texttt{.txt}, YOLO format~\cite{varghese2024yolov8}). As explained in~\cref{subsec:road_topography_anontations}, labels are provided as \(\langle\)class, \(x_{\mathrm{c}}\), \(y_{\mathrm{c}}\), \(w\), \(h\rangle\) normalized to image size.

\begin{table}[t]
\centering
\footnotesize
\setlength{\tabcolsep}{3pt}
\renewcommand{\arraystretch}{1.0}
\begin{tabular}{|l|p{0.7\linewidth}|}
\hline
\textbf{Device} & \textbf{Model and key specifications} \\
\hline
2$\times$ LiDAR &
Hesai XT32~\cite{hesai_xt32_product_2025}, 32-channel spinning, HFOV $360^\circ$, VFOV $\approx31^\circ$, range $\sim120$\,m, accuracy $\pm1$\,cm, precision $0.5$\,cm (1$\sigma$), angular resolution $0.18^\circ$ H and $1^\circ$ V \\
\hline
2$\times$ RGB camera &
IDS U3-3990SE C-HQ Rev.\,1.2~\cite{camera_refernece_u3_3990_rev12}, global shutter, lens FOV $90^\circ$ diagonal and $70^\circ$ horizontal \\
\hline
\end{tabular}
\caption{Perception sensor specifications for CARD}
\label{tab:sensor_rig_overview}
\end{table}

\begin{figure}[t]
    \centering
    \includegraphics[width=\linewidth]{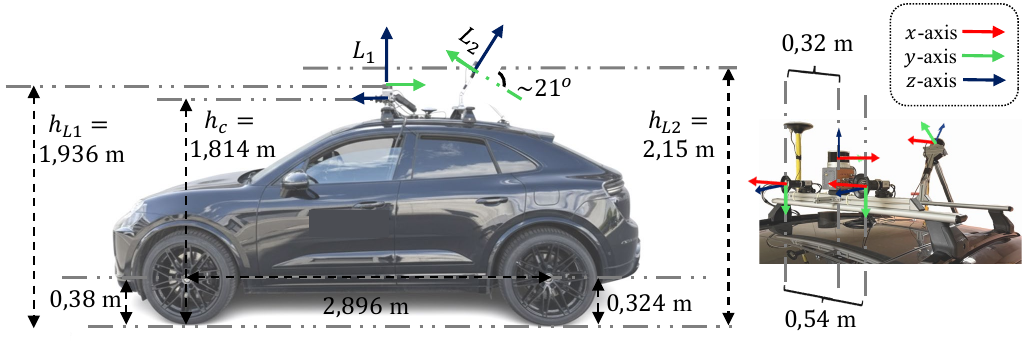}
    \caption{Rig visualization with sensor locations. Values are approximated and intended only to illustrate the setup. Legend: L1: front LiDAR, L2: rear LiDAR, and $h_c$: left-camera height.}

    \label{fig:car_diagram}
\end{figure}

\end{enumerate}

\begin{figure}[t]
  \centering
  \includegraphics[width=1\linewidth]{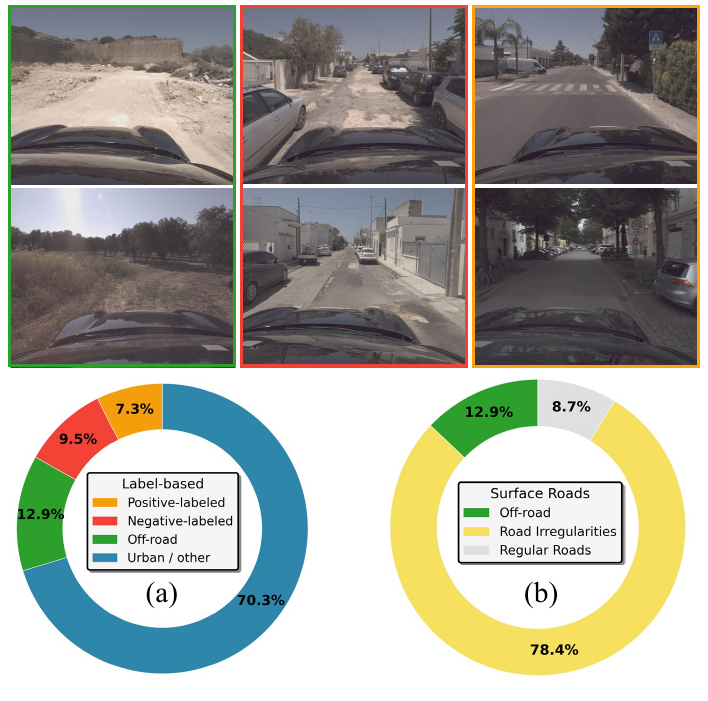}
\caption{(a) Image-level distribution of road-topography labels: \emph{positive}, such as speed bumps, \emph{negative}, such as potholes, and \emph{off-road}, such as non-asphalt cases. (b) Sequence-level statistics: a sequence is marked irregular if it contains at least one positive or negative instance. Example frames above, such as left: off-road, middle: negative, right: positive.}

  \label{fig:biggest_fig}
\end{figure}

\noindent\textbf{Dataset scale and splits.} CARD comprises 118 sequences and $\sim 350k$ images from left and right cameras, totaling $\sim$175k stereo pairs. The sequences range from 1 to 10 minutes. To reduce redundancy and support self-supervised methods, we sample stereo pairs at every third timestamp across all sequences. These sampled pairs are then partitioned into training, validation, and testing sets. This partitioning is stratified by road topography and location to ensure a similar distribution of these attributes across all sets, preserving the ratios of different road irregularities, which are shown in~\cref{fig:biggest_fig}. In addition, five full sequences are also held out for zero-shot evaluation of tasks requiring temporal consistency, such as visual odometry. This process yields $\sim$33k train, $\sim$11k validation, and 16k test stereo pairs.

\noindent\textbf{Privacy.}
To comply with the GDPR and protect personal privacy~\cite{GDPR_eu}, all released images are anonymized by a pretrained YOLOv8-based detector~\cite{varghese2024yolov8}. We also provide a dedicated contact address and an expedited takedown process for privacy concerns upon request by data subjects.

% \subsection{Calibration}
% \label{subsec:calibration}
% Drift-free cross-modal alignment is essential to achieve sub-centimeter accuracy, which is required for road-topography perception. Thus, all sensors are hardware-timestamped, time-synchronized, and jointly calibrated three times per recording day (start, mid, end), with per-sequence checks. The rig coordinate system is fixed at the IMU box, with axes $Z\uparrow$, $X\rightarrow$ (vehicle right), and $Y\rightarrow$ (forward).  We calibrate the two RGB cameras jointly with the front LiDAR using a calibration board. The rear LiDAR is registered to the front via SLAM-based alignment. Additionally, Leica survey instrumentation is used to measure eight points on the calibration board and, in the same session, the four wheel centers and wheel–ground contact points. These surveyed points are then rigidly aligned to the IMU frame, yielding consistent inter-sensor extrinsics and sensor-to-vehicle frames. Further implementation details are provided in the supplementary material.

\subsection{Calibration}
\label{subsec:calibration}

All sensors are time synchronized and jointly calibrated three times per recording day (start, mid, end). The rig coordinate system is fixed at the IMU box and all sensor relative poses are expressed relative to this frame. The two RGB cameras are calibrated jointly with the front LiDAR using a calibration board. The rear LiDAR is registered to the front by SLAM based alignment. Leica survey instrumentation measures \(N\) points on the calibration board that are observed by both the LiDAR and the camera, and in the same session surveys the four wheel centers and the wheel ground contact points. These surveyed points are rigidly aligned to the base frame, which yields consistent inter sensor extrinsics and sensor to vehicle transforms.

\subsection{IMU-Lidar Fusion}
\label{subsec:imu-lidar_fusion}

We estimate globally consistent 6 DoF world to rig poses in two stages. First, we run MC2SLAM~\cite{mc2slam_frank_2018} as tightly coupled LiDAR inertial odometry. Scans are deskewed with IMU data and each incoming sweep is registered against a local map with point-to-plane alignment. We apply minimal preprocessing with range limits and azimuth sector bounds. This real time stage produces odometry and a local map at the LiDAR spin rate, which is 10\,Hz. Second we refine the full trajectory offline with batch optimization that uses all IMU measurements the stage one odometry and loop closure constraints computed at revisits. The optimizer uses IMU preintegration factors that estimate gravity direction and time varying biases within a factor graph. Poses are reported in a right handed ENU frame with origin at the IMU beneath the front LiDAR and per sensor poses follow from the released extrinsics.

%original frank

% Our two-staged IMU-LiDAR fusion is based on a commercial solution provided by V\&R. A first stage is an implementation of the MC2SLAM algorithm \cite{mc2slam_frank_2018}. This is a real-time SLAM algorithm that tightly integrates IMU and LiDAR data in order to obtain an intermediate result for further processing. Though it achieves low drift and good registration performance already, it does not include loop closures and non-local optimization.
% To this end, a second stage focuses on global accuracy and map quality. The output of MC2SLAM is refined, peforming explicit loop closures and batch postprocessing. This is an offline optimization step includes \emph{all} available IMU data, loop closure constraints, and scan data.

% During both of these steps, laser data is only minimally filtered using range- and yaw-based cutoff filters.

% Crucial for the performance of real-time and offline SLAM are correct geometric calibration parameters for the LiDAR and IMU setup. The batch processing step sketched above can be used to perform this calibration prior to mounting the device on the car. This results in an individual characterization of the HESAI LiDARs calibration parameters, and the location of the IMU relative to the LiDAR.

\subsection{Ground-Truth Processing}
\label{subsec:gT_processing}
Most driving datasets provide single LiDAR scan as a depth ground truth, which is not sufficient for high-fidelity evaluation. Similar to the KITTI~\cite{KITTI_dataset,uhrig2017sparsity} and DrivingStereo~\cite{DrivingStereo} densification strategy, our goal is to aggregate LiDAR scans to obtain a dense scene representation while keeping ground truth strictly LiDAR-measured. However, applying stereo consensus may remove fine grained road geometry that is sometimes not detectable by stereo systems due to low parallax, especially at long range, which is why we built a novel multi LiDAR fusion strategy.

In brief, we motion-compensate, and aggregate all scans from a sequence into a voxelized grid. Then, we apply a filtering process to isolate the static environment by removing dynamic voxels. To generate the final ground truth for a specific evaluation timestamp, we re-insert its corresponding scan and perform hidden point removal to account for occlusions. For efficiency, this static voxel grid is computed once per sequence and cached for reuse across all frames. We illustrate each step of this pipeline next. Comprehensive details are provided in the supplementary material.

% In Kitti~\cite{KITTI_dataset}, Uhrig~\etal~\cite{uhrig2017sparsity} generate semi-dense KITTI GT by accumulating 11 LiDAR sweeps per frame and then using SGM stereo depth once after accumulation to remove inconsistent points from occlusions, dynamics, and measurement artifacts

\begin{figure*}
  \centering
  \includegraphics[width=1\linewidth]{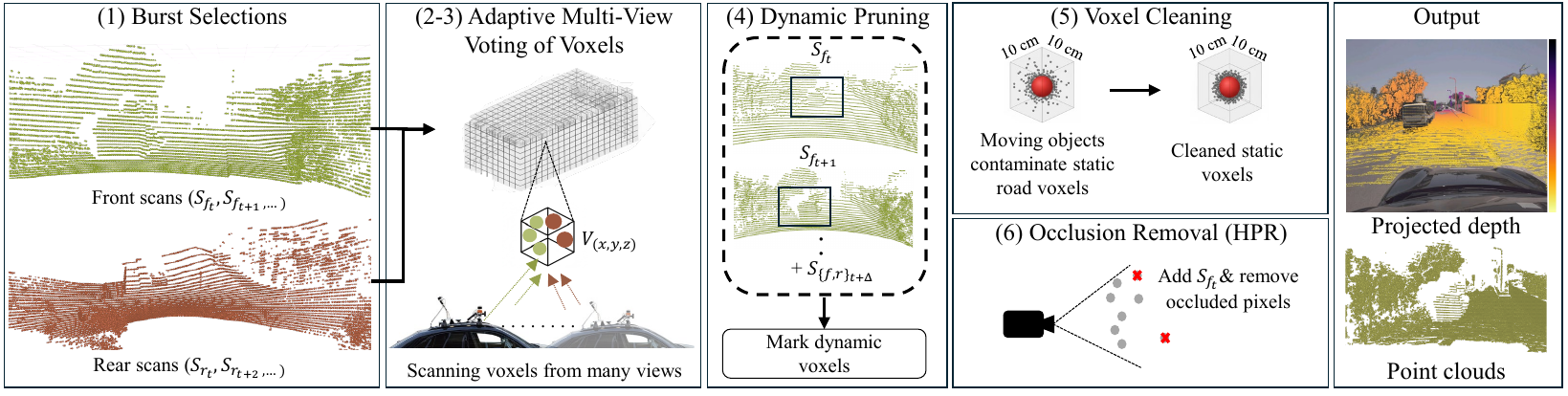}
\caption{Ground truth generation: motion-compensated LiDAR bursts are voxel-accumulated, dynamic-object filtered, then projected to the left camera to yield dense per-image GT, as explained in~\cref{subsec:gT_processing}.}

  \label{fig:data_processing}
\end{figure*}

\paragraph{Pipeline}
An overview of the ground truth construction is shown in \cref{fig:data_processing}. Full derivations, schedules, thresholds, and implementation details are provided in the supplementary material. The construction proceeds in the following stages.

\begin{enumerate}[leftmargin=0pt, itemindent=*, itemsep=2pt, topsep=2pt]
\item \textbf{Scans selection and alignment.}
For each camera timestamp we gather LiDAR bursts from both units within fixed time windows, cropping the front LiDAR to the forward \(180^{\circ}\) sector to remove simultaneous overlap with the rear unit. This reduces redundant observations for the subsequent multi view voting and prevents dynamic objects in the overlap region from receiving supporting votes from both sensors. Using provided poses, all points are placed in a common world frame.

 \item \textbf{Voxelization.}
We discretize space into cubic voxels of \(10\,\mathrm{cm}\) to aggregate evidence across views.

\item \textbf{Adaptive multi view voting with multi LiDAR redundancy.}
For each voxel we accumulate independent observations from the front and rear LiDAR. A view contributes a vote only if its sensor origin is at least \(3\,\mathrm{cm}\) away from already accepted origins, so each vote corresponds to a distinct motion baseline. We define the baseline \(b\) as the distance between accepted front origins. Voxels seen over a short baseline require more confirming votes, while voxels seen over a long baseline can be accepted with fewer votes. In practice the required votes from the front lidar decrease from \(4\) to \(1\) as \(b\) grows from \(0.20\,\mathrm{m}\) to \(0.90\,\mathrm{m}\), and the rear votes decrease from \(2\) to \(1\) over the same range. A voxel is kept as static when both the front and rear counts meet their baseline conditioned thresholds.

\item \textbf{Dynamic pruning with ICP flow.}
To remove residual dynamics, we register LiDAR scan pairs, already expressed in the common world frame, whose sensor origins are at least \(0.5\,\mathrm{m}\) apart using voxelized generalized ICP~\cite{koide2021voxelized}. After alignment, we compute a three dimensional residual displacement per voxel. A scan pair casts a moved vote for a voxel when the residual exceeds \(0.10\,\mathrm{m}\). A voxel is marked dynamic when it receives at least \(2\) moved votes, and these voxels are removed. In practice, this ICP flow stage is particularly effective at suppressing slow movers, such as walking pedestrians, that may survive the earlier filtering stage.

% We register burst pairs whose sensor origins are at least \(0.5\,\mathrm{m}\) apart in the world frame using a voxelized Generalized ICP method \cite{koide2021voxelized}.
% After alignment we compute a 3D residual per voxel.
% If the residual exceeds a fixed threshold \(t=0.10\,\mathrm{m}\) the pair casts a moved vote.
% A voxel is dynamic when moved votes \(\ge 2\) and moved fraction \(\ge 0.7\). Then, we eliminate these dynamic voxels. This ICP-based check follows the adaptive multi-view voting and clears motion that remains after the first step.

\item \textbf{Voxel cleaning by median absolute deviation.}
For each surviving voxel we compute the voxel center as the mean of its points and the radii \(r_i\) from each point to this center. Let \(\tilde r\) be the median of \(\{r_i\}\) and define the median absolute deviation (MAD) as \(\mathrm{MAD} = \mathrm{median}\bigl(|r_i - \tilde r|\bigr)\). We keep points that satisfy \(|r_i - \tilde r| \le 1.5\,\mathrm{MAD}\) and drop the voxel if fewer than \(2\) points remain. This robust per voxel trimming removes outlier returns from residual movers, for example wheel points from passing vehicles that can otherwise be fused into static road voxels.

% Moving objects can leave points inside road voxels (e.g., tire or skirt hits at the \(10\,\mathrm{cm}\) grid).
% For voxel \(v\) with points \(P_v\), let \(\mu_v\) be the mean of \(P_v\) and let \(r=\{\|p_j-\mu_v\|\!:\,p_j\in P_v\}\) be the set of point–to–center radii.
% We keep points using a MAD threshold with \(\sigma=1.5\):
% {\abovedisplayskip=2pt \belowdisplayskip=2pt
% \begin{gather}
% \mathrm{MAD}_v = \mathrm{median}\bigl(|\,r - \mathrm{median}(r)\,|\bigr), \label{eq:mad_def}\\
% \text{keep}(p_i) \iff r_i \le \sigma\,\mathrm{MAD}_v. \label{eq:mad_keep}
% \end{gather}}

% If fewer than two points remain we drop the voxel.
% This removes dynamic points, which contaminates the static road voxels.

\item \textbf{Occlusion removal.}
For each timestamp, we append the time aligned front LiDAR sweep. We then apply the hidden point removal algorithm in Open3D~\cite{zhou2018open3d}, using the left camera center as the viewpoint. The visibility radius is set to one hundred times the scene diameter, where the scene diameter is the maximum distance between any two points in the cloud. This keeps points that are directly visible from the camera and discards points that are occluded~\cite{katz2007direct}.

% \item \textbf{Monocular consensus as rejector only.}
% We run DepthAnyThing~\cite{DepthAnything2024} to predict a depth map from image and fit the predicted points clouds to LiDAR on overlapping valid pixels. This step is optional and used only where manual checks reveal residual dynamics.
% It filters gross inconsistencies, no predicted points are added.
% We keep a LiDAR point only if
% {\abovedisplayskip=2pt \belowdisplayskip=2pt
% \begin{gather}
% |z_{\mathrm{pred}}-z_{\mathrm{LiDAR}}| \le \tau_{\mathrm{rel}}\, z_{\mathrm{LiDAR}}, \label{eq:mono_consensus_err}\\
% z_{\mathrm{LiDAR}} \le r_{\max}. \label{eq:mono_consensus_range}
% \end{gather}}
% \noindent
% Here, $\tau_{\mathrm{rel}}$ is a tunable relative-error tolerance and $r_{\max}$ is the near-range cap.
% In most of experiments, $\tau_{\mathrm{rel}}=0.14$ and $r_{\max}=40\,\mathrm{m}$, chosen conservatively to avoid over-rejecting due to monocular inconsistencies.

\end{enumerate}
% \begin{table}[t]
% \centering
% \footnotesize
% \setlength{\tabcolsep}{2pt}
% \renewcommand{\arraystretch}{1.03}
% \begin{tabular}{@{}l*{5}{c}@{}}
% \toprule
% Method &
% RMSE [m] $\downarrow$ &
% AbsRel $\downarrow$ &
% $\delta_1 \uparrow$ &
% $\delta_3 \uparrow$ \\
% \midrule
% Pipeline
%   & 4.833 & 0.115 & 0.933 & 0.971 \\
% Pipeline (+MDE rejector)
%   & 2.582 & 0.051 & 0.984 &  0.997 \\
%   w/o multi-votes (single consensus)
%   & 6.368 & 0.165 & 0.897 & 0.956 \\
% w/o ICP pruning
%   & 5.274 & 0.129 & 0.920 &  0.965 \\
% voxel size 15 cm
%   & 7.495 & 0.197 & 0.857 &  0.946 \\
% \bottomrule
% \end{tabular}
% \caption{Ablation of the aggregation pipeline using MoGeV2~\cite{MoGE_V2} as a fixed monocular probe.}
% \label{tab:agg_ablation_moge}
% \end{table}

In sequences where manual inspection reveals residual dynamics, we apply a monocular depth estimator (MDE) consensus filter. DepthAnything~\cite{DepthAnything2024} predicts a depth map, and we fit it to the sparse front LiDAR sweep. For each projected LiDAR point, we compute the absolute relative error, and remove points whose absolute relative error exceeds \(15\%\). This threshold is set slightly above the typical absolute relative error of DepthAnything~\cite{DepthAnything2024}, so that only the worst inconsistencies are rejected. In practice, these large errors usually represent residual movers. In addition, we provide a quantitative ablation of our pipeline in the supplementary material. As a qualitative example, \cref{fig:ablation_ex} illustrates the effect of our voxel cleaning and adaptive voting strategies in a sequence containing dynamic objects. Without these refinements, moving agents are incorrectly registered into the static road geometry. Our full pipeline effectively suppresses these artifacts to recover only static points.

\begin{figure}[t]
  \centering
  \includegraphics[width=\linewidth]{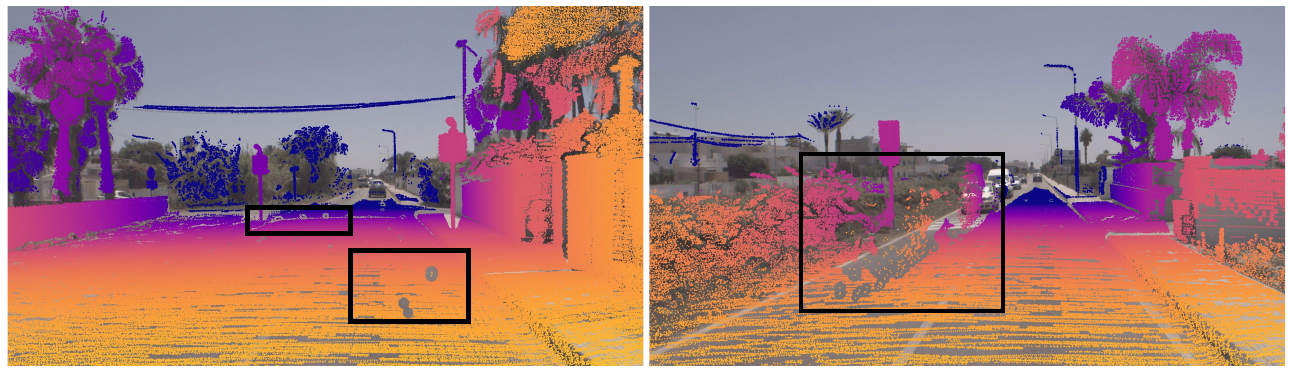}
    \caption{
    Qualitative ablation of voxel cleaning and adaptive voting. 
    Left: projected depth without voxel cleaning, clearly showing dynamic points registered into static road voxels. 
    Right: projected depth with our pipeline using a single consensus across both LiDARs, without the adaptive voting strategy.
    }
  \label{fig:ablation_ex}
\end{figure}

% In addition, ablation results in~\cref{tab:agg_ablation_moge} evaluate the main aggregation components using MoGeV2~\cite{MoGE_V2} as a fixed depth estimator, so that only aggregation and cleaning choices vary. The ablations are run on three sequences with pronounced dynamic content, where monocular depth is required to remove residual movers, we use monocular depth only as a rejector, since making it a core aggregation component may bias the process. Adaptive voxel voting and increasing voxel size yield the largest quantitative gains. MAD-based voxel cleaning, which operates at roughly the \(10\,\mathrm{cm}\) level, has negligible effect on the numerical metrics but clearly sharpens the geometry and suppresses outliers, as shown in~\cref{fig:ablation_ex}.

\begin{figure*}

  \centering
  \includegraphics[width=0.95\linewidth]{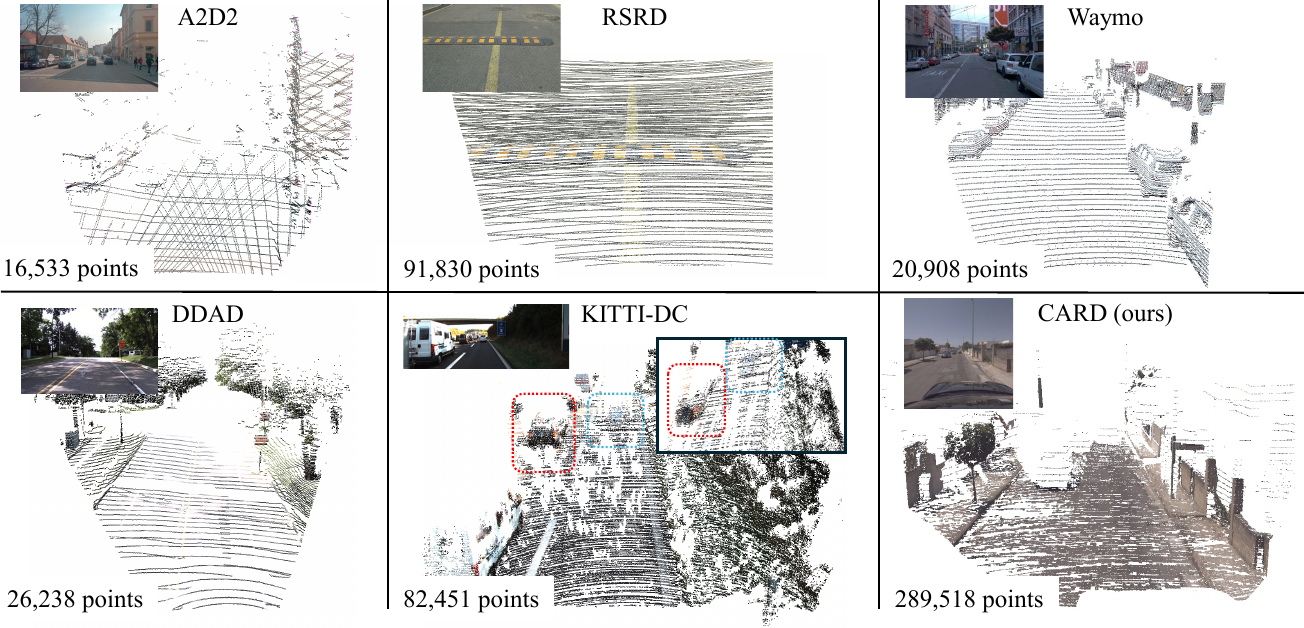}
    \caption{Colored point cloud comparison. We show CARD alongside densified ground truth from KITTI-DC~\cite{uhrig2017sparsity,KITTI_dataset} and RSRD~\cite{zhao2023rsrd}, and LiDAR based ground truth from A2D2~\cite{geyer2020a2d2}, Waymo~\cite{waymodataset}, and DDAD~\cite{PackNet_DDAD}. Each panel shows the input image at top left and the point cloud count in bottom left, and only the KITTI-DC panel includes a zoom-in illustrating distortions from dynamic objects~\cite{KITTI_dataset,uhrig2017sparsity}.}

  \label{fig:data_comparison}
\end{figure*}

% \begin{table}[t]
%     \centering
%     \scriptsize

%     \setlength{\tabcolsep}{4pt}
%     \label{tab:agg_ablation}
%     \begin{tabular}{lccccc}
%         \toprule
%         Configuration 
%         & RMSE $\downarrow$ 
%         & Abs.\ Rel.\ $\downarrow$ 
%         & $\delta_1$ $\uparrow$ 
%         & $\delta_2$ $\uparrow$ 
%         & $\delta_3$ $\uparrow$ \\
%         \midrule
%         Full pipeline 
%         & 0.000 & 0.000 & 0.000 & 0.000 & 0.000 \\

%         Voxel size 15\,cm 
%         & 0.000 & 0.000 & 0.000 & 0.000 & 0.000 \\

%         Voxel size 20\,cm 
%         & 0.000 & 0.000 & 0.000 & 0.000 & 0.000 \\

%         No front LiDAR cropping 
%         & 0.000 & 0.000 & 0.000 & 0.000 & 0.000 \\

%         No MAD cleaning 
%         & 0.000 & 0.000 & 0.000 & 0.000 & 0.000 \\

%         No ICP flow 
%         & 0.000 & 0.000 & 0.000 & 0.000 & 0.000 \\
%         \bottomrule
%     \end{tabular}
%         \caption{Ablation results of the depth aggregation pipeline. }
% \end{table}

% Figure (force placement here; requires \usepackage{float})
% \begin{figure}[t]
%   \centering
%   \includegraphics[width=\linewidth]{sec/figures/example_prof.pdf}
%   \caption{Aggregation examples. (a) Top row: scene with a dynamic car. (b) Bottom row: rough off-road terrain. Left: input image. Right: processed GT projected onto the image (color map: yellow = near, purple = far).}
%   \label{fig:agg_example}
% \end{figure}

% Preamble (ensure these)
% \usepackage{booktabs,tabularx,subcaption,array}

\subsection{Wheel Excitation}
\label{subsec:wheel_trajectory}

We define wheel excitation as the trajectory of each tire’s ground contact point. Given a rigid calibration at standstill between the IMU and each wheel contact point, as introduced in \ref{subsec:calibration}, we apply this transform to the ego poses, which yields an approximate ground path for each wheel under a no-suspension assumption. We then collect LiDAR returns inside a narrow corridor around this path, limited by the tire footprint and expected suspension travel. Each point is reduced to a 1D sample by computing its distance along the path and its height relative to the path using the suspension axis, assuming parallel suspension. Subsequently, we fit a smooth cubic spline to the distance–height samples using the Ceres Solver~\cite{agarwal2012ceres_solver} with a Tukey robust loss to down-weight outliers and read off the spline at each timestamp. The resulting vertical offset is added to the approximate path to yield the wheel excitation trajectory, which provides a time varying extrinsic transform from the perception rig to the road surface. This yields sensor–road extrinsics at any given timestamp and serves as an online local road-level reference to express depth or point clouds as height relative to the current road surface, with full interpolation details provided in the supplementary material.

\subsection{Road Topography Annotations}
\label{subsec:road_topography_anontations}

Our dataset emphasizes the 3D understanding of road topography. Since these features induce minimal geometric deviations relative to the global environment, specialized annotations are required to evaluate these fine-grained variations, as shown in~\cref{sec:benchmark}. We define positive topography as elements protruding above the road plane (e.g., speed bumps) and negative topography as depressions (e.g., potholes). To ensure high-quality ground truth, we employed a semi-automated pipeline: a \(40\%\) manually annotated subset was used to train a YOLOv8~\cite{varghese2024yolov8} model, which assisted in labeling the remaining \(60\%\). Crucially, these annotations are designed to benchmark 3D reconstruction accuracy in these critical regions rather than standard 2D detection. Additionally, for off-road segments where the entire drivable surface is intrinsically irregular, we provide per-sequence labels rather than local bounding boxes. Comprehensive analysis and qualitative results are provided in the supplementary material.

\section{Benchmark}
\label{sec:benchmark}

\begin{table*}[t]
\captionsetup{font=small}
\scriptsize
\setlength{\tabcolsep}{1pt}
\renewcommand{\arraystretch}{1.05}
% \label{tab:card_depth_est} % <-- moved label to after \caption

% ===== (a) Depth Estimation =====
\centering
\captionsetup{justification=raggedright,singlelinecheck=false} % avoid tiny caption overfulls

% Use flexible centered columns so width fits exactly
\begin{tabularx}{\linewidth}{@{} >{\raggedright\arraybackslash}p{1.65cm} *{18}{Y} @{}}
\toprule
\multirow{2}{*}{Method} &
\multicolumn{2}{c}{AbsRel $\downarrow$} &
\multicolumn{2}{c}{SqRel $\downarrow$} &
\multicolumn{2}{c}{RMSE $\downarrow$} &
\multicolumn{2}{c}{RMSElog $\downarrow$} &
\multicolumn{2}{c}{$\delta_1 \uparrow$} &
\multicolumn{2}{c}{$\delta_2 \uparrow$} &
\multicolumn{2}{c}{$\delta_3 \uparrow$} &
\multicolumn{2}{c}{Height AbsDiff $\downarrow$} &
\multicolumn{2}{c}{Height $\delta_{10}$ $\uparrow$} \\
\cmidrule(lr){2-3}\cmidrule(lr){4-5}\cmidrule(lr){6-7}\cmidrule(lr){8-9}
\cmidrule(lr){10-11}\cmidrule(lr){12-13}\cmidrule(lr){14-15}
\cmidrule(lr){16-17}\cmidrule(lr){18-19}
 & F & B & F & B & F & B & F & B & F & B & F & B & F & B & F & B & F & B \\
\midrule
DAV2~\cite{DepthAnythingV2_2025}
& 0.096 & 0.055 & 0.548 & 0.084 & 4.107 & 0.683 & 0.154 & 0.060 & 0.895 & 0.976 & 0.969 & 0.989 & 0.990 & 0.992 & 0.181 & 0.106 & 0.578 & 0.620 \\
DepthPro~\cite{DepthPro}
& 0.100 & 0.045 & 0.830 & 0.075 & 3.270 & 0.602 & 0.142 & 0.050 & 0.908 & 0.974 & 0.971 & 0.989 & 0.986 & \best{0.993} & 0.249 & 0.086 & 0.630 & 0.726 \\
Metric3D2~\cite{Metric3D_v2}
& 0.060 & 0.039 & \best{0.165} & \secbest{0.034} & 1.836 & 0.474 & \secbest{0.085} & 0.042 & \secbest{0.976} & \best{0.990} & \best{0.997} & \best{0.993} & \best{0.999} & \best{0.993} & \best{0.117} & \secbest{0.074} & \secbest{0.675} & \secbest{0.754} \\
UniDepth2L~\cite{UniDepth2025_V2}
& \secbest{0.046} & \secbest{0.029} & 0.204 & \best{0.022} & \secbest{1.825} & \secbest{0.382} & \best{0.074} & \secbest{0.032} & \best{0.981} & \best{0.992} & 0.995 & \best{0.993} & \best{0.999} & \best{0.993} & \best{0.117} & \secbest{0.055} & \best{0.802} & \secbest{0.860} \\
MoGe2L~\cite{MoGE_V2}
& \best{0.045} & \best{0.027} & \secbest{0.177} & \best{0.022} & \best{1.807} & \best{0.337} & \best{0.074} & \best{0.029} & \best{0.980} & \secbest{0.989} & \secbest{0.996} & \best{0.993} & \best{0.999} & \best{0.993} & \secbest{0.119} & \best{0.051} & \best{0.801} & \best{0.893} \\
\cmidrule[0.3pt]{1-19} 

MoGe2L$^\dagger$~\cite{MoGE_V2}
& 0.029 & 0.018 & 0.039 & 0.005 & 0.894 & 0.209 & 0.042 & 0.020 & 0.996 & 0.993 & 0.999 & 0.993 & 0.999 & 0.993 & 0.051 & 0.034 & 0.893 & 0.981 \\

\cmidrule[0.3pt]{1-19} 

FS~\cite{foundationstereo_2025}
& 0.040 & 0.014 & 0.411 & 0.006 & 2.162 & 0.185 & 0.088 & 0.016 & 0.976 & 0.994 & 0.989 & 0.994 & 0.994 & 0.994 & 0.177 & 0.025 & 0.892 & 0.986 \\
\bottomrule
\end{tabularx}

\caption{CARD benchmark for monocular and stereo depth estimation. All results employ per-image median scaling to evaluate relative geometry. (F) and (B) denote full-image and bounding-box restricted evaluations, respectively. The top rows evaluate monocular models in a zero-shot setting, followed by MoGe2L$^\dagger$ fine-tuned on CARD. The final row presents the FoundationStereo~\cite{foundationstereo_2025} zero-shot stereo baseline. A comprehensive analysis is provided in the Supplementary Material.}
\label{tab:card_depth_est} 

\vspace{-1mm}
\end{table*}
In this section, we evaluate state-of-the-art zero-shot depth estimators on CARD and report results for depth completion models trained and tested on our benchmark. We consider two protocols: full-image evaluation and a focused evaluation restricted to our irregularity bounding boxes. In addition, we introduce a modified variant of a leading depth completion model that further improves the result. All evaluations are capped at 80\,m.

\subsection{Depth Estimation}
\label{subsec:depth_prediction}

In~\cref{tab:card_depth_est}, we evaluate strong depth estimation models in a zero-shot setting on CARD, and include a stereo baseline, which is FoundationStereo~\cite{foundationstereo_2025}. Following standard practice, we report AbsRel, SqRel, RMSE, RMSElog, and $\delta_{1/2/3}$, as defined in prior work~\cite{MonoPP,Monodepth2,saxena2008make3d}. In addition to full-image evaluation, we introduce a per-box protocol that measures performance only inside bounding boxes of annotated road irregularities.

Beyond depth, we report height metrics derived from depth using our calibration kit. Using the continuous trajectory of the wheel-road contact point together with the vehicle pose, we convert depth to height relative to the road. We summarize height with Absolute Difference (Abs. Diff) and $\delta_{10}$\,cm, which is the fraction of pixels within 10\,cm of the ground-truth height. We provide more results in the supplementary material.

\begin{figure}[t]
  \centering
  % Example layout: two rows, four columns. Replace with your own files.
  \includegraphics[width=\linewidth]{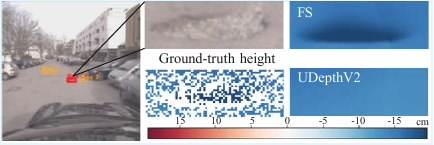}
  \caption{Height ground truth and predictions for a pothole: GT (left), UniDepthV2~\cite{UniDepth2025_V2}, FoundationStereo~\cite{foundationstereo_2025} (right).}
  \label{fig:height_estimation_qual}
\end{figure}

Our focus is the recovery of relative road geometry, so monocular results are reported with median scaling. Full results with and without median scaling are provided in the supplementary material. Overall, zero shot monocular depth estimators underperform on road surface irregularities, while performing strongly on global depth metrics. In contrast, the stereo baseline shows strong generalization on centimeter-level features, benefiting from geometry-driven disparity-to-depth conversion, as shown in~\cref{fig:height_estimation_qual}.

\paragraph{Fine-tuning results. } We conducted fine-tuning experiments on CARD using the released splits. Our preliminary results indicated that training with a standard $L_1$ loss yielded negligible gains, as the model prioritized global scale and structure over local road topography, as shown in~\cite{gfm2025}. Conversely, we found that fine-tuning MoGe2L \cite{MoGE_V2} by combining its affine-invariant loss with an $L_1$ height-space loss led to substantial improvements, as presented as MoGe2L$^\dagger$ in~\cref{tab:card_depth_est}.

\subsection{Depth Completion}
\label{subsec:depth_completion}

We benchmark recent depth completion models on CARD using the provided train, validation, and test splits. Specifically, we train and evaluate BP-NET~\cite{BP-NET_2024} and DMD3C~\cite{DMD3c_2025}, reporting RMSE, iMAE, and \(\delta_1\) under both full and boxes protocols, as shown in~\cref{tab:card_depthCompletion}.

To better preserve fine road-surface irregularities, we introduce a variant of DMD3C~\cite{DMD3c_2025} that keeps the original distillation pipeline but replaces the monocular teacher with FoundationStereo~\cite{foundationstereo_2025} and swaps the scale-invariant loss function for a direct L1 loss in metric depth, since the stereo teacher provides metric depth natively. This change yields consistent improvements over the DMD3C baseline in our benchmark, as shown in~\cref{tab:card_depthCompletion}. Results on CARD indicate that stereo-distilled depth completion with denser LiDAR input provides a viable means to densify dynamic objects in the ground truth depth while avoiding temporal aggregation artifacts. However, as seen in the per-box results in~\cref{tab:card_depthCompletion,tab:card_depth_est}, FoundationStereo~\cite{foundationstereo_2025} still outperforms the DMD3C+FS variant in box-relevant metrics. This suggests that, even with a strong stereo teacher, LiDAR–image depth completion may oversmooth or suppress valid road topographies, potentially degrading the target regions that CARD is designed to evaluate.

% ===== Depth Completion (compact: RMSE, iMAE, AbsRel, δ1 only) =====
\begin{table}[t]
\centering
\scriptsize
\setlength{\tabcolsep}{2pt}
\renewcommand{\arraystretch}{1.03}
\begin{tabular}{@{}l
*{2}{c}  % RMSE F/B
*{2}{c}  % iMAE F/B
*{2}{c}  % AbsRel F/B
*{2}{c}  % delta1 F/B
@{}}
\toprule
\multirow{2}{*}{Method} &
\multicolumn{2}{c}{RMSE [m] $\downarrow$} &
\multicolumn{2}{c}{iMAE [$1$/m] $\downarrow$} &
\multicolumn{2}{c}{AbsRel $\downarrow$} &
\multicolumn{2}{c}{$\delta_1 \uparrow$} \\
\cmidrule(lr){2-3}\cmidrule(lr){4-5}\cmidrule(lr){6-7}\cmidrule(lr){8-9}
 & F & B & F & B & F & B & F & B \\
\midrule
BP\text- NET~\cite{BP-NET_2024}
  & 0.7975 & \secbest{0.1939}
  & \best{0.0044} & \best{0.0028}
  & \best{0.0211} & 0.0170
  & \best{0.9853} & \best{0.9948} \\
DMD3C~\cite{DMD3c_2025}
  & \secbest{0.7742} & 0.1950
  & \secbest{0.0050} & \best{0.0028}
  & 0.0225 & \secbest{0.0157}
  & 0.9804 & 0.9900 \\
DMD3C (+FS~\cite{foundationstereo_2025})
  & \best{0.7510} & \best{0.1918}
  & 0.0057 & 0.0029
  & \secbest{0.0219} & \best{0.0155}
  & \secbest{0.9805} & \best{0.9984} \\
\bottomrule
\end{tabular}
\caption{Evaluation results on depth completion task.}
\label{tab:card_depthCompletion}
\end{table}

\section{Conclusion}
Most autonomous driving datasets are recorded in well paved urban settings, so diverse road topography is under represented and hard to use for evaluating safety critical behavior. Estimating fine road geometry is challenging, and, to our knowledge, there is no dedicated benchmark that targets road topography in realistic forward driving. CARD addresses this gap by focusing on irregular road geometry, including off road segments, varied pothole shapes and sizes, and speed bumps with diverse profiles and placements. The dataset provides quasi-dense ground truth derived entirely from LiDAR with visibility and motion checks. Alongside depth ground truth, CARD includes calibrated sensor poses, wheel excitation traces, and bounding boxes that target road irregularities, enabling both full image and box focused evaluation. Evaluations for monocular and stereo depth estimation and for depth completion, together with height based metrics, establish a dedicated benchmark for these under represented scenarios.

\paragraph{Limitations.} Despite manual review, the anonymization stage can blur benign content as false positives. Furthermore, while the 3D ground truth undergoes manual inspection, residual dynamic artifacts may persist in rare cases of slowly moving objects. We provide a dedicated reporting channel for such cases. In addition, our irregularity annotations are intended for 3D reconstruction evaluation rather than standard object detection. Given the inherent visual ambiguity in defining the boundaries of local depressions, some surface irregularities may remain unlabeled.

\section*{Acknowledgment} 
We would like to thank Dr. Matthias Zeller for his invaluable guidance and technical oversight, which significantly contributed to the structured presentation and clarity of this work. We also extend our gratitude to Dr. Marc Fessler for his strategic support throughout the open-sourcing procedure and dataset release. Special thanks are due to Jurij Wolf for his assistance in sensor calibration and logistics management. Finally, we sincerely thank the legal team at CARIAD SE for their extensive efforts in reviewing the legal and privacy aspects of this dataset. Furthermore, we would like to acknowledge Florian Jaumann, Lorenz Ott, and Francesco Maldonato for their valuable assistance.

%% file: sec/X_suppl.tex
\clearpage
\setcounter{page}{1}
\setcounter{section}{0}
\maketitlesupplementary

\section{Sensors Calibration}
\label{sec:calibration}

Accurate road geometry requires consistent calibration of cameras, LiDARs, the IMU, and the wheel contact points. We therefore use a two stage procedure: Camera-LiDAR calibration, followed by calibrating the rig, which is referenced at the IMU, to each wheel.

\textbf{Camera-LiDAR calibration.}
For each recording day, we perform three calibration sessions. In each session, we record $\sim500$ synchronized LiDAR and stereo image samples while observing a calibration board at different distances, orientations, and positions. Camera intrinsics and extrinsics are estimated from board detections. The front LiDAR is registered to the cameras by fitting a common plane between LiDAR returns and the calibration board in the images. The rear LiDAR and the IMU are then calibrated with respect to the front LiDAR using the batch optimization stage of our two stage IMU–LiDAR fusion pipeline based on MC2SLAM~\cite{mc2slam_frank_2018}. This step yields individual calibration parameters for both Hesai LiDARs and the pose of the IMU in the front LiDAR frame.

\begin{figure}[t]
    \centering
    \begin{subfigure}[t]{\linewidth}
        \centering
        \includegraphics[width=0.5\linewidth,trim=0 25 0 0,clip]{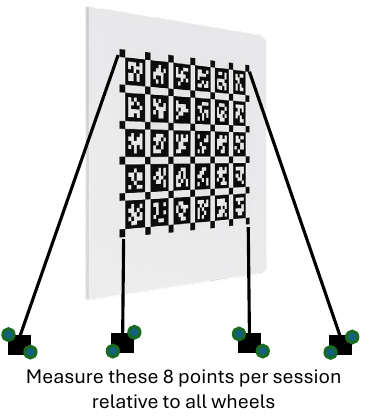}
        \caption{Calibration board with eight Leica measured points (circles).}
        \label{fig:wheel_calibration_a}
    \end{subfigure}

    \vspace{2pt}

    \begin{subfigure}[t]{\linewidth}
        \centering
        \includegraphics[width=\linewidth,trim=0 0 0 14,clip]{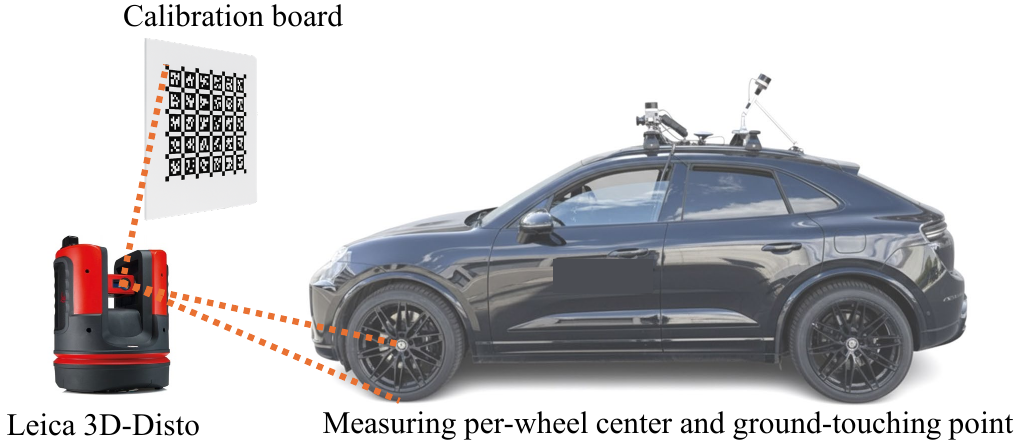}
        \caption{Leica 3D~DISTO setup measuring board points, wheel centers, and wheel contact points.}
        \label{fig:wheel_calibration_b}
    \end{subfigure}

    \caption{Overview of the rig and wheel calibration procedure. (a) Leica measurements of calibration board points. (b) Combined measurement of board points and wheel geometry, used to link the wheels, vehicle base frame, and sensor rig with sub centimeter accuracy.}
    \label{fig:wheel_calibration}
\end{figure}

\textbf{Rig-Wheel calibration with Leica.}
After calibrating cameras, LiDARs, and IMU, we perform an additional rig and wheel calibration once per recording day with a Leica 3D~Disto device. The calibration board is mounted on flat ground in three distinct poses. For each pose, the Leica device measures the three dimensional coordinates of eight marked points on the board, as shown in~\cref{fig:wheel_calibration_a}. With the vehicle and the board kept fixed, we also measure for each wheel the wheel center and the corresponding ground contact point, as shown in~\cref{fig:wheel_calibration_b}. In the same session, the stereo cameras observe the board, which allows us to register Leica measurements with the board points reconstructed in the camera frame. The combined set of board and wheel measurements defines a full calibration between the wheels, the vehicle base frame, and the sensor rig. We repeat the Leica session three times to ensure consistency and jointly refine all parameters to minimize reprojection error.

% Our two-staged IMU-LiDAR fusion is based on a commercial solution provided by V\&R. A first stage is an implementation of the MC2SLAM algorithm \cite{mc2slam_frank_2018}. This is a real-time SLAM algorithm that tightly integrates IMU and LiDAR data in order to obtain an intermediate result for further processing. Though it achieves low drift and good registration performance already, it does not include loop closures and non-local optimization.
% To this end, a second stage focuses on global accuracy and map quality. The output of MC2SLAM is refined, peforming explicit loop closures and batch postprocessing. This is an offline optimization step includes all available IMU data, loop closure constraints, and scan data.

% During both of these steps, laser data is only minimally filtered using range- and yaw-based cutoff filters.

% Crucial for the performance of real-time and offline SLAM are correct geometric calibration parameters for the LiDAR and IMU setup. The batch processing step sketched above can be used to perform this calibration prior to mounting the device on the car. This results in an individual characterization of the HESAI LiDARs calibration parameters, and the location of the IMU relative to the LiDAR.

\section{Wheel Excitation}
\label{supplementary_sec:wheel_excitation}
Wheel excitation is the trajectory of each wheel's contact point along the ground. The transformations between the IMU frame and wheel-ground contact points are calibrated for the vehicle in standstill, as described in~\cref{sec:calibration}. Transforming the optimized SLAM trajectory to the wheel contact points yields a rigid-body baseline trajectory. To recover the true wheel excitation, which is induced from the terrain profile, we must account for the suspension travel relative to this rigid baseline.

Limited by the tire footprint and suspension travel, a subset of the global point cloud $P_w$ is extracted along the approximate trajectory for wheel $w$. Projecting these ground points onto the 1D manifold defined by the wheel center trajectory and the up vector, under a parallel-suspension assumption, produces a 1D representation of the ground height relative to the statically calibrated wheel contact point.

For every point $p \in P_w$, we find the two closest adjacent trajectory points at timestamps $i-1$ and $i$ with accompanying rotation matrices and translation vectors $(R_{i-1}, t_{i-1}), (R_{i}, t_{i}) \in (\mathbb{R}^{3 \times 3}, \mathbb{R}^3)$. Let $d_i$ denote the cumulative arclength of trajectory pose $i$ from the start of the wheel trajectory. To interpolate the correct position on the manifold, the relative offset to the first trajectory pose is computed as
\begin{equation}
  \alpha = \left\langle \frac{t_i - t_{i-1}}{\|t_i - t_{i-1}\|},\, p - t_{i-1} \right\rangle \Big/ \|t_i - t_{i-1}\|.
\end{equation}
The projected distance of $p$ along the trajectory is then
\begin{equation}
d_p = (1 - \alpha)\, d_{i-1} + \alpha\, d_i
\end{equation}
where $d_{i-1}$ and $d_i$ are the arc lengths from the start of the trajectory for poses $i-1$ and $i$, respectively. Finally, the relative height of the point w.r.t. the trajectory is
\begin{equation}
h_p = (1 - \alpha)\,\left(R_{i-1}^{\intercal}(p - t_{i-1})\right)_z
    + \alpha\,\left( R_{i}^{\intercal}(p - t_{i}) \right)_z.
\end{equation}

To obtain a smooth 1-D excitation profile, we use Ceres to fit a cubic spline in arclength–height space over the set $\{(d_p, h_p) \mid p \in P_w\}$. Applying a robust Tukey loss ensures good outlier suppression, such that spurious measurements from dynamic objects or raised dust are effectively ignored. The final wheel-excitation signal is obtained by evaluating this spline at the trajectory arclengths $\{d_i\}$ and translating the nominal wheel trajectory by the corresponding relative heights.

\section{Ground-Truth Aggregation Details}
\label{supp:gt_pipeline}

This section gives the full specification of the LiDAR aggregation pipeline used to construct the quasi-dense ground truth. All steps are implemented in our released script and are applied per driving sequence. Unless otherwise stated, all coordinates are in metres.

We denote coordinate frames by $a,b\in\{w,r,c,\ell_f,\ell_r\}$, corresponding to world ($w$), vehicle/rig ($r$), camera ($c$), front LiDAR ($\ell_f$), and rear LiDAR ($\ell_r$). A 3D point expressed in frame $a$ is written as ${}^a\mathbf{p}\in\mathbb{R}^3$, and its homogeneous counterpart as ${}^a\tilde{\mathbf{p}} = [({}^a\mathbf{p})^\top,\,1]^\top$.  

A rigid transform ${}^aT_b\in\mathrm{SE}(3)$ maps coordinates from frame $b$ to frame $a$:
\begin{equation}
{}^a\tilde{\mathbf{p}} = {}^aT_b\,{}^b\tilde{\mathbf{p}}, \qquad {}^aT_c = {}^aT_b\,{}^bT_c.
\end{equation}
Thus, ${}^wT_r(t)$ is the rig pose in the world at time $t$, which is the base pose always referenced in the IMU center below the front LiDAR, and ${}^rT_s$ is the time-invariant extrinsic calibration from sensor $s$ to the rig.  

Each LiDAR burst $b$ from sensor $s\in\{\ell_f,\ell_r\}$ provides a finite set of points
\begin{equation}
\mathcal{P}_b = \bigl\{{}^s\mathbf{p}^{(k)}\in\mathbb{R}^3\bigr\}.
\end{equation}

%---------------------------------------------
\subsubsection*{A. Motion compensation and voxel grid}
%---------------------------------------------

For burst $b$ from sensor $s$ with end timestamp $t_b^{\text{end}}$, we first form the sensor pose in the world:
\begin{equation}
{}^wT_s(b) \;=\; {}^wT_r(t_b^{\text{end}})\,{}^rT_s.
\end{equation}
Each LiDAR point ${}^s\mathbf{p}^{(k)}$ is then transformed to the world frame:
\begin{equation}
{}^w\mathbf{p}^{(k)} \;=\; \Pi\!\bigl({}^wT_s(b)\,{}^s\tilde{\mathbf{p}}^{(k)}\bigr),
\end{equation}
where $\Pi$ drops the homogeneous coordinate.  
For the front LiDAR ($s=\ell_f$), we additionally crop to the forward half-space in the sensor frame to avoid overlap with the rear unit and reduce redundant support.

\paragraph{Voxelization.}
We discretize the world into a regular grid of cubic voxels of edge length
\[
s_{\text{vox}} = 0.10~\text{m}.
\]
The voxel index of a world-frame point ${}^w\mathbf{p}$ is
\begin{equation}
\mathbf{i}({}^w\mathbf{p}) \;=\;
\bigl\lfloor {}^w\mathbf{p} / s_{\text{vox}}\bigr\rfloor \in \mathbb{Z}^3.
\end{equation}
In implementation, $\mathbf{i}$ is stored via a hash for efficiency, but conceptually we operate on the integer index $\mathbf{i}$.

%---------------------------------------------
\subsubsection*{B. Baseline-adaptive multi-LiDAR voting}
%---------------------------------------------

\paragraph{Viewpoints per voxel.}
For each voxel $v$ we keep:
\begin{itemize}[leftmargin=1.0em,itemsep=2pt,topsep=2pt]
  \item a list of distinct front-LiDAR sensor origins in world coordinates, $\mathcal{O}^F_v = \{{}^w\mathbf{o}_j\}$,
  \item a rear-LiDAR hit count $n^R_v$,
  \item an axis-aligned bounding box accumulated over all points assigned to $v$.
\end{itemize}
For a front burst with sensor pose ${}^wT_{\ell_f}(b)$, the sensor origin in the world is the translation part ${}^w\mathbf{o}$. We append ${}^w\mathbf{o}$ to $\mathcal{O}^F_v$ only if it is at least
\[
d_{\min} = 0.03~\text{m}
\]
away from all existing origins in $\mathcal{O}^F_v$, enforcing a minimal motion baseline between ``distinct'' views.

\paragraph{Per-voxel motion baseline.}
The effective motion baseline for voxel $v$ is the maximum pairwise distance between its front viewpoints:
\begin{equation}
b_v \;=\;
\begin{cases}
\max\limits_{j,\ell} \bigl\|{}^w\mathbf{o}_j - {}^w\mathbf{o}_\ell\bigr\|_2, & |\mathcal{O}^F_v|\ge 2,\\[2pt]
0, & \text{otherwise}.
\end{cases}
\end{equation}

\paragraph{Baseline-dependent vote quotas.}
Given a vote range $[v_{\min}, v_{\max}]$ and baseline anchors
\[
b_n = 0.20~\text{m},\qquad b_f = 0.90~\text{m},
\]
we define
\begin{gather}
\operatorname{clip}(b) = \min\bigl\{\max(b, b_n),\, b_f\bigr\},\\
\hat{v}(b) = v_{\max} - \frac{\operatorname{clip}(b) - b_n}{b_f - b_n}\,\bigl(v_{\max}-v_{\min}\bigr),\\
\operatorname{req}(b; v_{\min},v_{\max}) = \mathrm{round}\!\bigl(\hat{v}(b)\bigr).
\end{gather}

Small baselines ($b\!\approx\! b_n$) require many confirmations $v_{\max}$, whereas large baselines ($b\!\approx\! b_f$) allow fewer confirmations $v_{\min}$.

For the front and rear LiDAR we use
\[
(v_{\min}^F,v_{\max}^F) = (1,4),\qquad
(v_{\min}^R,v_{\max}^R) = (1,2),
\]
and obtain per-voxel quotas
\begin{align}
n^F_v &\ge \operatorname{req}\bigl(b_v; v_{\min}^F,v_{\max}^F\bigr),\\
n^R_v &\ge \operatorname{req}\bigl(b_v; v_{\min}^R,v_{\max}^R\bigr),
\end{align}
where $n^F_v = |\mathcal{O}^F_v|$ is the number of distinct front viewpoints that hit $v$.

\paragraph{Rigid voxel constraint.}
We also track the axis-aligned bounding box of all world-frame points assigned to voxel $v$ and require that its spatial extent remains small:
\begin{equation}
\bigl\|{}^w\mathbf{p}_{\max,v}-{}^w\mathbf{p}_{\min,v}\bigr\|_2
\;\le\; \varepsilon_{\text{rigid}}, \qquad
\varepsilon_{\text{rigid}} = 0.20~\text{m}.
\end{equation}
This constraint acts as a temporal consistency filter, rejecting voxels whose content moves or deforms over time.

\paragraph{Global static voxel set.}
A voxel $v$ belongs to the global static set $\mathcal{V}_{\text{static}}$ if and only if it satisfies all three conditions:
\begin{equation}
v\in\mathcal{V}_{\text{static}}
\iff
\begin{cases}
n^F_v \ge \operatorname{req}(b_v; v_{\min}^F,v_{\max}^F),\\[1pt]
n^R_v \ge \operatorname{req}(b_v; v_{\min}^R,v_{\max}^R),\\[1pt]
\bigl\|{}^w\mathbf{p}_{\max,v}-{}^w\mathbf{p}_{\min,v}\bigr\|_2 \le \varepsilon_{\text{rigid}}.
\end{cases}
\end{equation}
This static grid is computed once per sequence and cached for all frames.

%---------------------------------------------
\subsubsection*{C. Dynamic pruning with ICP flow}
%---------------------------------------------

Even with baseline voting and rigidity checks, sometimes moving objects can accumulate into apparently static voxels, such as pedestrians. We therefore add an ICP-based dynamic pruning stage.

\paragraph{LiDAR-scan pairing.}
Let $b_i$ denote a LiDAR scan with sensor pose ${}^wT_s(b_i)$ and origin ${}^w\mathbf{c}_i$.
We form scan pairs $(b_i, b_j)$ such that
\begin{equation}
\bigl\|{}^w\mathbf{c}_j - {}^w\mathbf{c}_i\bigr\|_2 \;\ge\; b_{\text{flow}},\qquad
b_{\text{flow}} = 0.50~\text{m}.
\end{equation}

\paragraph{Residual-based motion votes.}
For each pair $(b_i,b_j)$ we align their world-frame point clouds using voxelized generalized ICP, estimating a rigid transform ${}^{w}T^{\text{ICP}}_{w'}$ that maps the points from burst $b_j$ to the reference burst $b_i$ (both are expressed in $w$). For every point ${}^w\mathbf{p}\in\mathcal{P}_{b_i}$ we compute the post-alignment residual
\begin{equation}
\delta({}^w\mathbf{p}) \;=\; \min_{{}^w\mathbf{q}\in \mathcal{P}_{b_j}}
\bigl\|{}^w\mathbf{p} - {}^{w}T^{\text{ICP}}_{w'}\,{}^w\tilde{\mathbf{q}}\bigr\|_2.
\end{equation}
If $\delta({}^w\mathbf{p})$ exceeds
\[
\tau_{\text{flow}} = 0.10~\text{m},
\]
we cast a ``moved'' vote for the voxel $v$ with index $\mathbf{i}({}^w\mathbf{p})$. For voxel $v$ we track $m_v$ (number of moved votes) and $c_v$ (total comparisons). A voxel is flagged as dynamic if
\begin{equation}
m_v \;\ge\; V_{\text{dyn}}
\quad\text{and}\quad
\frac{m_v}{c_v} \;\ge\; \eta_{\text{dyn}},
\end{equation}
with $V_{\text{dyn}} = 2$ and $\eta_{\text{dyn}} = 0.7$. All such voxels are removed from $\mathcal{V}_{\text{static}}$.

%---------------------------------------------
\subsubsection*{D. Per-frame static point selection}
%---------------------------------------------

For each camera timestamp $t_{\text{cam}}$, we select LiDAR bursts whose time windows overlap a sensor-specific interval around $t_{\text{cam}}$. We utilize the specific intervals of $[-3, 13]\,\text{s}$ for the front LiDAR and $[0, 30]\,\text{s}$ for the rear LiDAR. Let $\mathcal{B}(t_{\text{cam}})$ denote the selected bursts.

\paragraph{Multi-LiDAR redundancy per frame.}
Within $\mathcal{B}(t_{\text{cam}})$, a voxel must be confirmed by the two distinct LiDAR units. For voxel $v$ we track the set of sensors
\begin{equation}
\mathcal{S}_v = \bigl\{s : \exists\, b\in\mathcal{B}(t_{\text{cam}}),\,{}^w\mathbf{p}\in\mathcal{P}_b \text{ with } \mathbf{i}({}^w\mathbf{p})=v\bigr\},
\end{equation}
and define the per-frame redundant set
\begin{equation}
\mathcal{V}^{(t)}_{\text{multi}} = \{v : |\mathcal{S}_v|\ge 2\}.
\end{equation}

\paragraph{Static world cloud for the frame.}
Stacking all points from selected bursts,
\begin{equation}
\mathcal{P}^{(t)}_{\text{all}} = \bigcup_{b\in\mathcal{B}(t_{\text{cam}})} \mathcal{P}_b,
\end{equation}
we keep only those whose voxel belongs to both the global static set and the per-frame multi-LiDAR set:
\begin{equation}
\mathcal{P}^{(t)}_{\text{static}} =
\bigl\{{}^w\mathbf{p}\in\mathcal{P}^{(t)}_{\text{all}} :
\mathbf{i}({}^w\mathbf{p})\in \mathcal{V}_{\text{static}}
\cap \mathcal{V}^{(t)}_{\text{multi}}\bigr\}.
\end{equation}

\paragraph{Per-voxel MAD outlier cleaning.}
Residual dynamic points can still fall into otherwise static voxels. For each voxel $v$ with point set $P_v\subset\mathcal{P}^{(t)}_{\text{static}}$ we compute the centroid
\begin{equation}
{}^w\boldsymbol{\mu}_v = \frac{1}{|P_v|}\sum_{{}^w\mathbf{p}\in P_v} {}^w\mathbf{p},
\end{equation}
and radii $r_i = \bigl\|{}^w\mathbf{p}_i - {}^w\boldsymbol{\mu}_v\bigr\|_2$. We form a robust scale estimate (median absolute deviation) and keep only points that satisfy
\begin{equation}
r_i \;\le\; \sigma_{\text{vox}}\,\mathrm{MAD}_v,
\qquad
\sigma_{\text{vox}} = 1.5.
\end{equation}
Voxels with fewer than $N_{\text{vox}}^{\min}=2$ surviving inliers are dropped. This yields the cleaned static world cloud $\tilde{\mathcal{P}}^{(t)}_{\text{static}}$ used for projection.

For each timestamp, we additionally append a ``keyframe'' front-LiDAR burst, which is the first burst ending after $t_{\text{cam}}$, to guarantee a dense near-range sampling around the evaluation timestamp.

%---------------------------------------------
\subsubsection*{E. Projection and occlusion}
%---------------------------------------------

\paragraph{Projecting on the left camera.}
For camera $c$ with intrinsics $K_c$ and extrinsics ${}^rT_c$ we form the world-to-camera transform
\begin{equation}
{}^cT_w(t_{\text{cam}}) \;=\;
\bigl({}^wT_r(t_{\text{cam}})\,{}^rT_c\bigr)^{-1}.
\end{equation}
Every static world point is transformed to the camera frame,
\begin{equation}
{}^c\mathbf{p} = \Pi\!\bigl({}^cT_w(t_{\text{cam}})\,{}^w\tilde{\mathbf{p}}\bigr),
\end{equation}

\paragraph{Ego-vehicle mask.}
We precompute a 2D binary ego mask $M_{\text{ego}}(u,v)$ in the undistorted camera image space, marking pixels occupied by the hood. Any LiDAR point whose projection falls inside $M_{\text{ego}}$ is discarded before monocular consensus and visibility reasoning.

\paragraph{Hidden-point removal.}
To obtain only camera-visible points, we apply the hidden point removal (HPR) algorithm~\cite{katz2007direct} implemented in Open3D~\cite{zhou2018open3d}. Denoting by $\mathcal{P}_c$ the current camera-frame cloud, we run two passes of HPR with radii
\begin{equation}
R_1 = \alpha_1\,D,\qquad R_2 = \alpha_2\,D,
\end{equation}
where $D$ is the diagonal length of the cloud bounding box and $(\alpha_1,\alpha_2)=(100,210)$. We apply a two-pass HPR strategy. The first pass uses a strict radius ($R_1$) to aggressively remove static background points that are occluded by dynamic objects, which may appear as single scan lines. In the second pass, we re-introduce the keyframe burst and apply a relaxed radius ($R_2$). This restores valid sparse points at long range that were over-aggressively culled by the dense near-field points in the first pass, ensuring distant road geometry is preserved.

The resulting camera-frame point cloud for timestamp $t_{\text{cam}}$ is saved.

For a small subset of sequences with particularly challenging dynamics, we optionally run a monocular depth consensus step using a pre-trained DepthAnything model~\cite{DepthAnything2024}. This step is never used to densify or inpaint depth, it acts purely as a conservative rejector on top of the LiDAR-only pipeline above.

\paragraph{Monocular depth estimation (MDE) consensus.}
Given the undistorted RGB image, DepthAnything predicts a per-pixel depth map $\hat{z}_{\text{DA}}(u,v)$ up to an unknown scale. We align this scale to LiDAR by minimizing a robust one-dimensional least-squares problem over overlapping pixels:
\begin{align}
z_{\text{LiDAR}}(u,v) &: \text{LiDAR rasterization},\\
s^\star &= \operatorname*{argmin}_{s>0}
\sum_{(u,v)\in\Omega}
\bigl(z_{\text{LiDAR}}(u,v) - s\,\hat{z}_{\text{DA}}(u,v)\bigr)^2,
\end{align}
where $\Omega$ contains pixels where both signals are valid and we drop extreme percentiles for robustness. We then form a metrically aligned prediction
\begin{equation}
z_{\text{DA}}(u,v) = s^\star \hat{z}_{\text{DA}}(u,v).
\end{equation}

\paragraph{Per-point consistency test.}
For each remaining LiDAR point ${}^c\mathbf{p} = (x_c,y_c,z_c)$ we project to pixel coordinates $(u,v)$ and read the aligned DepthAnything depth $z_{\text{DA}}(u,v)$. If no prediction is available at that pixel or if $z_c > r_{\max}$ with
\[
r_{\max} = 40~\text{m},
\]
we keep the point unmodified. Otherwise, we compute absolute and relative errors
\begin{equation}
e_{\text{abs}} = \bigl|z_{\text{DA}}(u,v) - z_c\bigr|,
\qquad
e_{\text{rel}} = \frac{e_{\text{abs}}}{z_c}.
\end{equation}
A LiDAR point is kept if it satisfies the joint tolerance
\begin{equation}
e_{\text{rel}} \le \tau_{\text{rel}}
\quad\text{and}\quad
e_{\text{abs}} \le \max\!\bigl(\tau_{\text{abs}},\, \tau_{\text{rel}} z_c\bigr),
\end{equation}
with
\[
\tau_{\text{rel}} = 0.14,\qquad
\tau_{\text{abs}} = 0.10~\text{m}.
\]
Points that violate this condition are removed and treated as residual movers. Importantly, we never add monocular points to the ground truth, MDE is used exclusively as a sanity check on the LiDAR geometry and only on demand.

\begin{figure}[t]
    \centering
    \setlength{\abovecaptionskip}{4pt} 
    \setlength{\belowcaptionskip}{-10pt} 

    % Replace 'pipeline_composite.png' with your actual filename
    \includegraphics[width=1.0\linewidth]{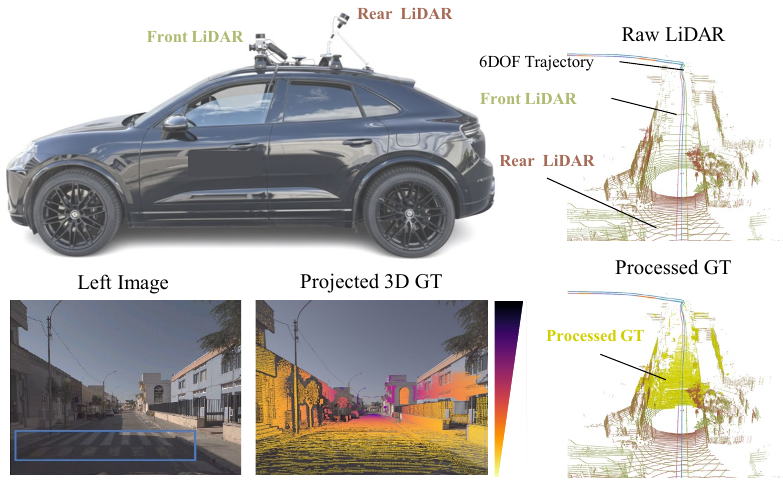}

    \caption{
    {Qualitative Overview of the Ground-Truth Generation Pipeline.} 
    The ego-vehicle (top left) captures data using front and rear LiDARs. 
    The right column compares the Raw LiDAR along the 6-DoF trajectory (top) against the final Processed GT (bottom), showing the densification effect. 
    The bottom row illustrates the downstream utility: a Left Image with a speed bump annotation and its corresponding Projected 3D GT depth map.
    }
    \label{fig:pipeline_overviewxx}
\end{figure}

\noindent
Finally, we present the qualitative output of our ground-truth generation pipeline in~\cref{fig:pipeline_overviewxx}. The figure visualizes the sensor setup, the intermediate point cloud aggregation, and the final dense geometry projected into the image frame for annotation and evaluation.
%---------------------------------------------
\subsubsection*{F. Numerical hyper-parameters}
%---------------------------------------------

Table~\ref{tab:gt_hparams} summarizes the numerical settings used across all sequences for the pipeline described above.

% \subsubsection*{G. Ablation results}
% We have conducted ablation studies, as shown in~\cref{tab:abb_agg_ablation}, to evaluate the effect of each main compoennt

% \begin{table}[t]
%     \centering
%     \scriptsize

%     \setlength{\tabcolsep}{4pt}
%     \label{tab:abb_agg_ablation}
%     \begin{tabular}{lccccc}
%         \toprule
%         Configuration 
%         & RMSE $\downarrow$ 
%         & Abs.\ Rel.\ $\downarrow$ 
%         & $\delta_1$ $\uparrow$ 
%         & $\delta_2$ $\uparrow$ 
%         & $\delta_3$ $\uparrow$ \\
%         \midrule
%         Full pipeline 
%         & 0.000 & 0.000 & 0.000 & 0.000 & 0.000 \\

%         Voxel size 15\,cm 
%         & 0.000 & 0.000 & 0.000 & 0.000 & 0.000 \\

%         Voxel size 20\,cm 
%         & 0.000 & 0.000 & 0.000 & 0.000 & 0.000 \\

%         No front LiDAR cropping 
%         & 0.000 & 0.000 & 0.000 & 0.000 & 0.000 \\

%         No MAD cleaning 
%         & 0.000 & 0.000 & 0.000 & 0.000 & 0.000 \\

%         No ICP flow 
%         & 0.000 & 0.000 & 0.000 & 0.000 & 0.000 \\
%         \bottomrule
%     \end{tabular}
%         \caption{Ablation results of the depth aggregation pipeline. }
% \end{table}

\vspace{0.5em}

\begin{table}[t]
\centering
\footnotesize
\setlength{\tabcolsep}{4pt}
\renewcommand{\arraystretch}{1.05}
\begin{tabular}{@{}llc@{}}
\toprule
Stage & Parameter & Value \\
\midrule
Voxel grid &
Voxel size $s_{\text{vox}}$ & $0.10$ m \\
& Min.\ front view spacing $d_{\min}$ & $0.03$ m \\
& Rigidity radius $\varepsilon_{\text{rigid}}$ & $0.20$ m \\
\midrule
Baseline voting &
Near / far baseline $b_n,b_f$ & $0.20 / 0.90$ m \\
& Front votes $v_{\min}^F,v_{\max}^F$ & $1 / 4$ \\
& Rear votes $v_{\min}^R,v_{\max}^R$ & $1 / 2$ \\
\midrule
Per-frame selection &
Min.\ LiDARs per voxel $|\mathcal{S}_v|$ & $\ge 2$ \\
\midrule
ICP pruning &
Flow baseline $b_{\text{flow}}$ & $0.50$ m \\
& Residual threshold $\tau_{\text{flow}}$ & $0.10$ m \\
& Moved votes $V_{\text{dyn}}$ & $2$ \\
& Moved fraction $\eta_{\text{dyn}}$ & $0.7$ \\
\midrule
Voxel MAD &
MAD multiplier $\sigma_{\text{vox}}$ & $1.5$ \\
& Min.\ inliers $N_{\text{vox}}^{\min}$ & $2$ \\
\midrule
Visibility &
HPR multiplier (1st / 2nd pass) & $100 / 210$ \\
\midrule
DepthAnything consensus &
Rel.\ error $\tau_{\text{rel}}$ & $0.15$ \\
& Abs.\ floor $\tau_{\text{abs}}$ & $0.10$ m \\
& Max.\ range $r_{\max}$ & $40$ m \\
\bottomrule
\end{tabular}
\caption{Numerical settings for the LiDAR aggregation pipeline.
All values are fixed across sequences. There are a few examples which required manual filtering by controlling the $\tau_{\text{rel}}$ of the MDE}
\label{tab:gt_hparams}
\end{table}

\begin{figure*}[t]
    \centering
    % Row 1 -----------------------------------------------------------
    \begin{subfigure}[t]{0.24\textwidth}
        \centering
        \includegraphics[width=\linewidth,trim=0 0 0 340,clip]{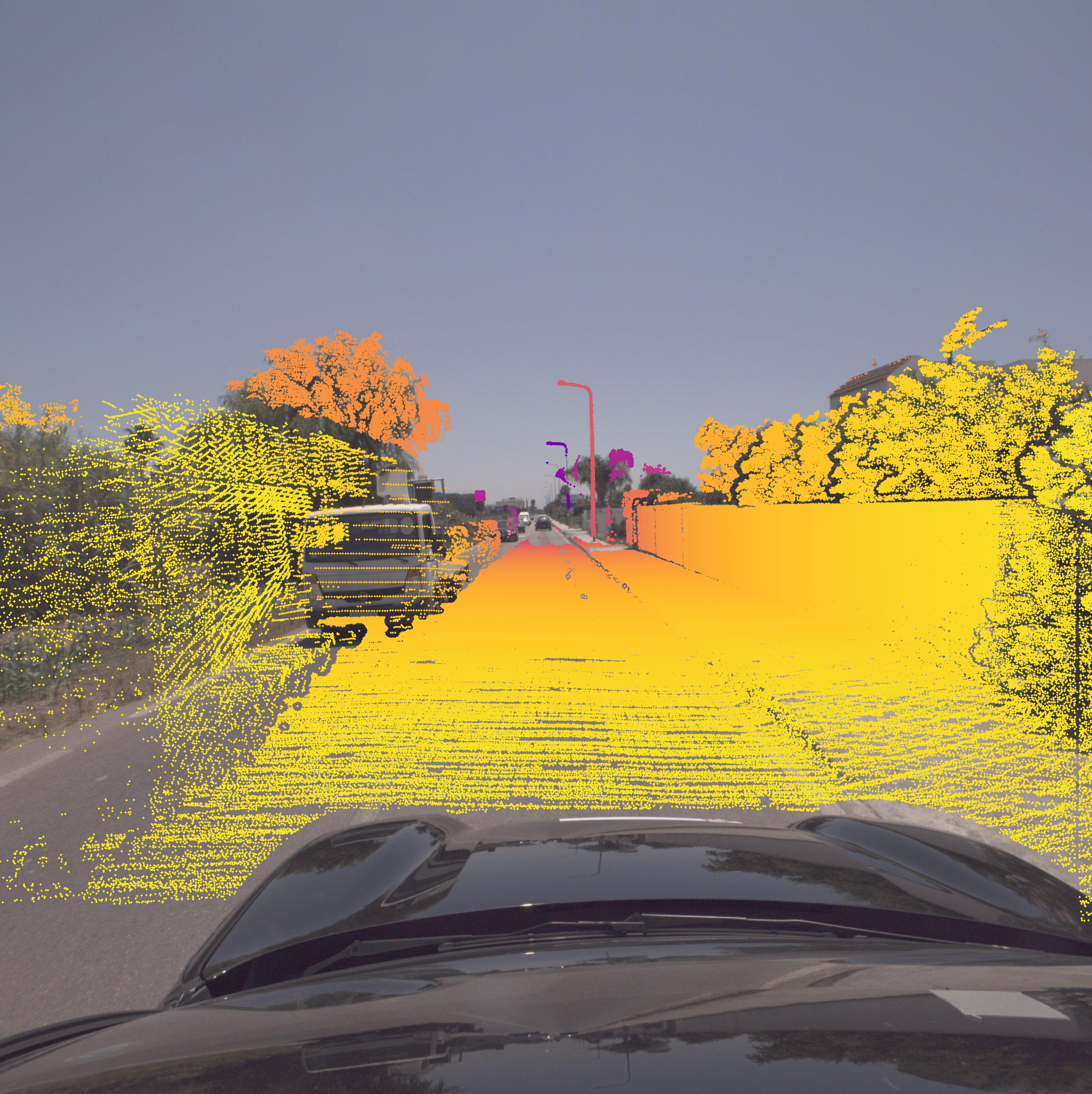}
        \caption{Single consensus between front and rear lidar, no cropping.}
        \label{fig:abl_3}
    \end{subfigure}\hfill
    \begin{subfigure}[t]{0.24\textwidth}
        \centering
        \includegraphics[width=\linewidth,trim=0 0 0 340,clip]{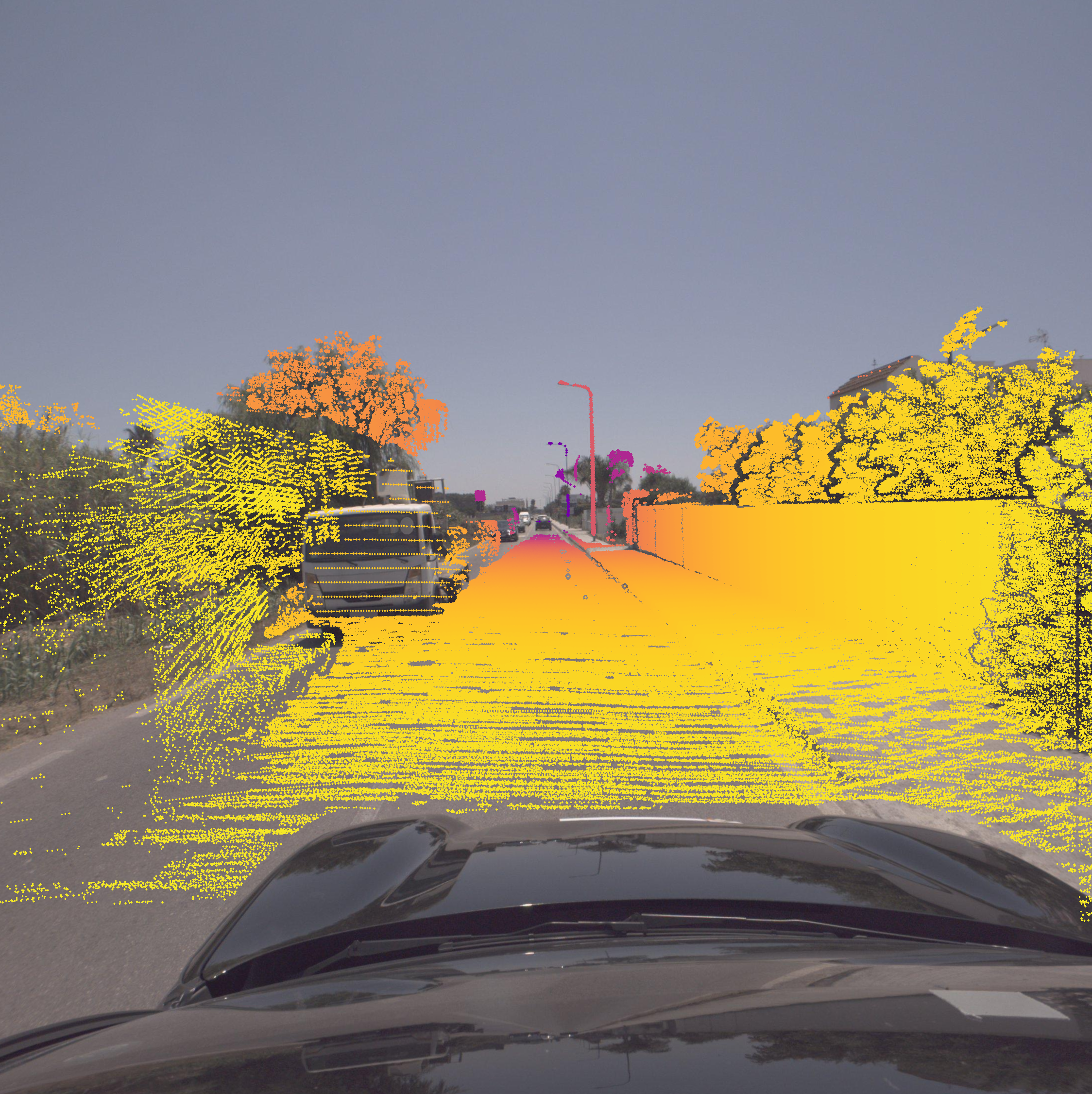}
        \caption{Addition of front LiDAR forward cropping.}
        \label{fig:abl_3_crop}
    \end{subfigure}\hfill
    \begin{subfigure}[t]{0.24\textwidth}
        \centering
        \includegraphics[width=\linewidth,trim=0 0 0 340,clip]{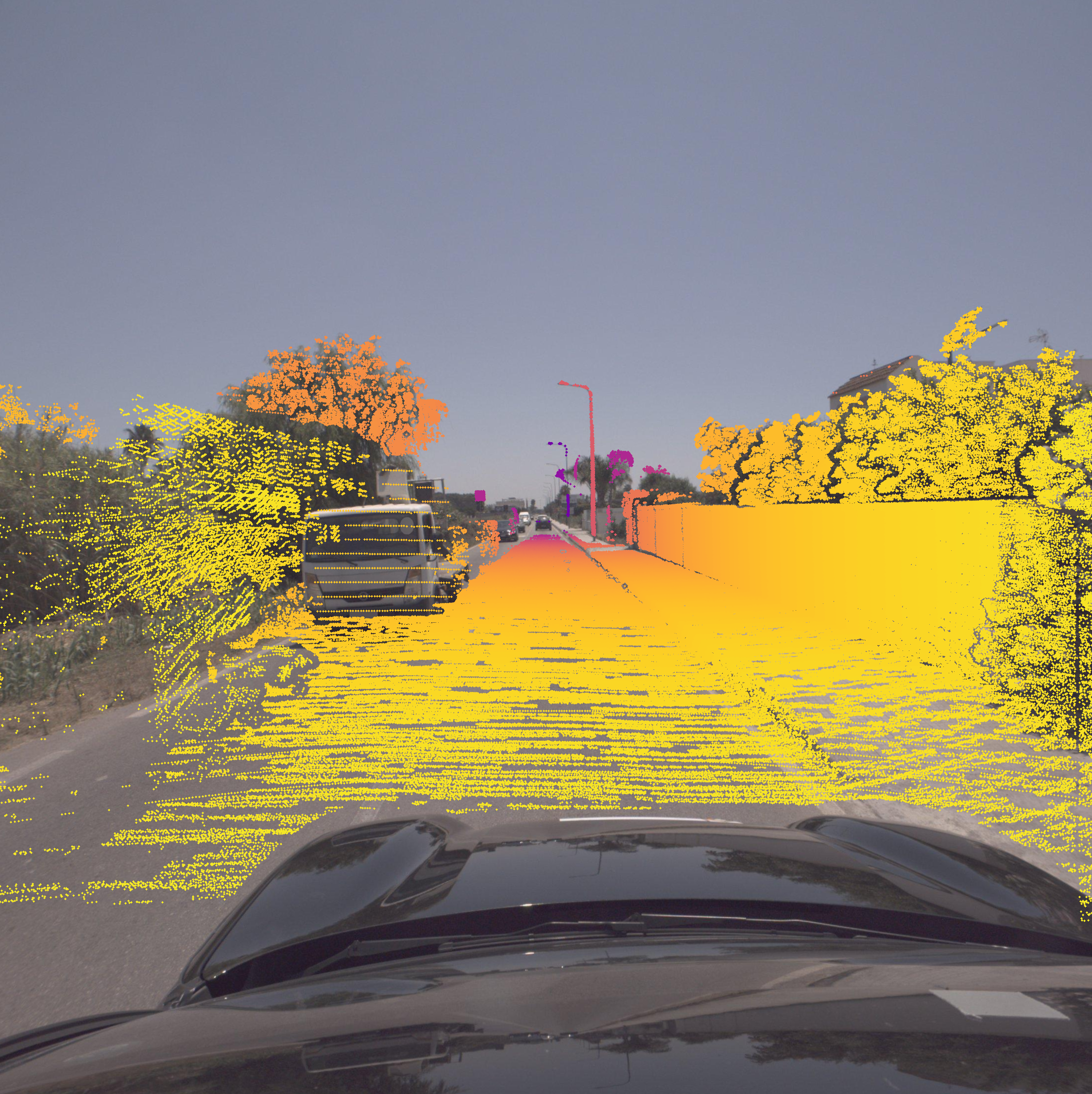}
        \caption{Fixed voting thresholds (Front: 2, Rear: 1).}
        \label{fig:abl_4}
    \end{subfigure}\hfill
    \begin{subfigure}[t]{0.24\textwidth}
        \centering
        \includegraphics[width=\linewidth,trim=0 0 0 340,clip]{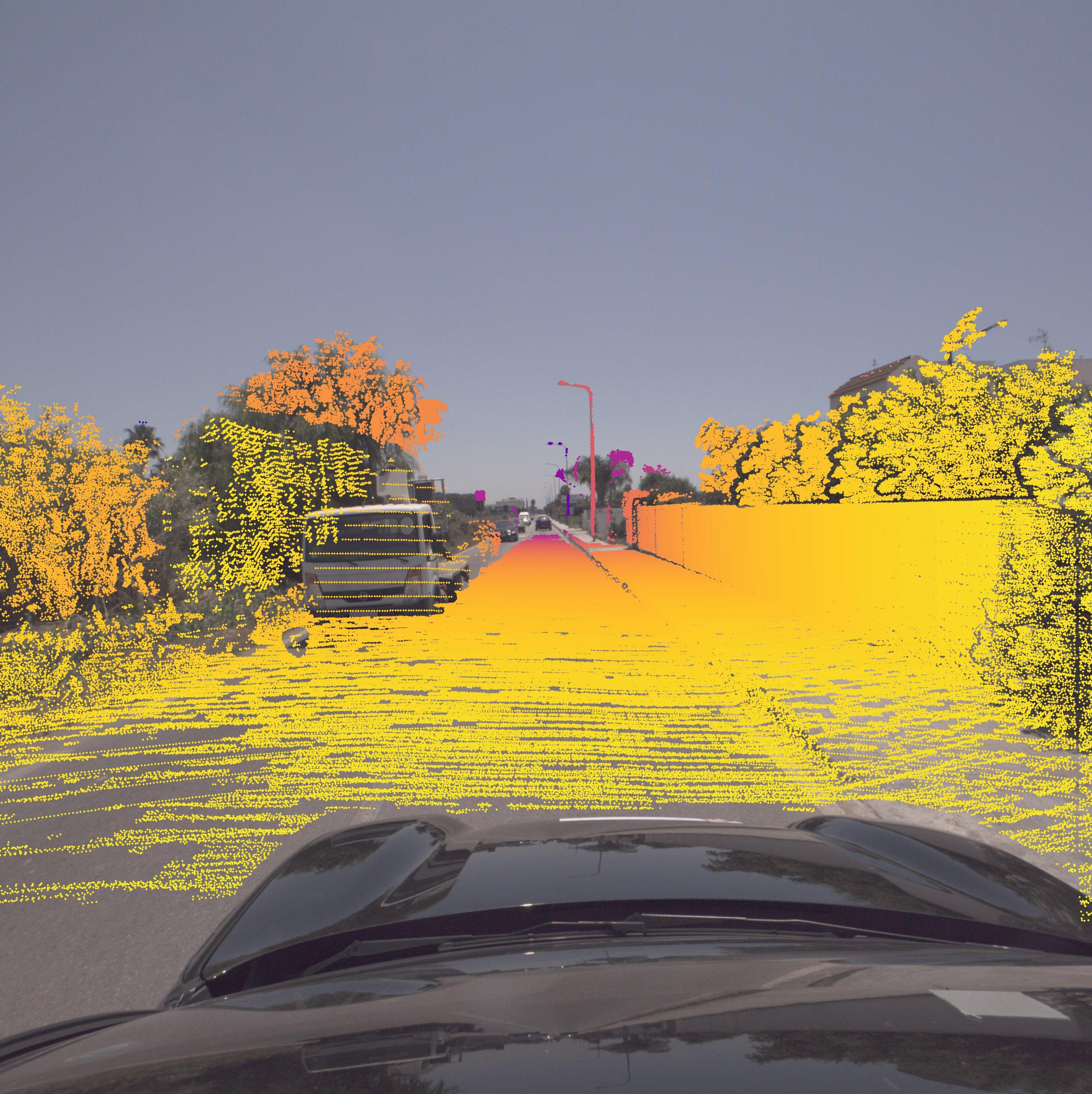}
        \caption{Baseline-adaptive consensus.}
        \label{fig:abl_5}
    \end{subfigure}

    \vspace{0.5em}

    % Row 2 -----------------------------------------------------------
    \begin{subfigure}[t]{0.24\textwidth}
        \centering
        \includegraphics[width=\linewidth,trim=0 0 0 340,clip]{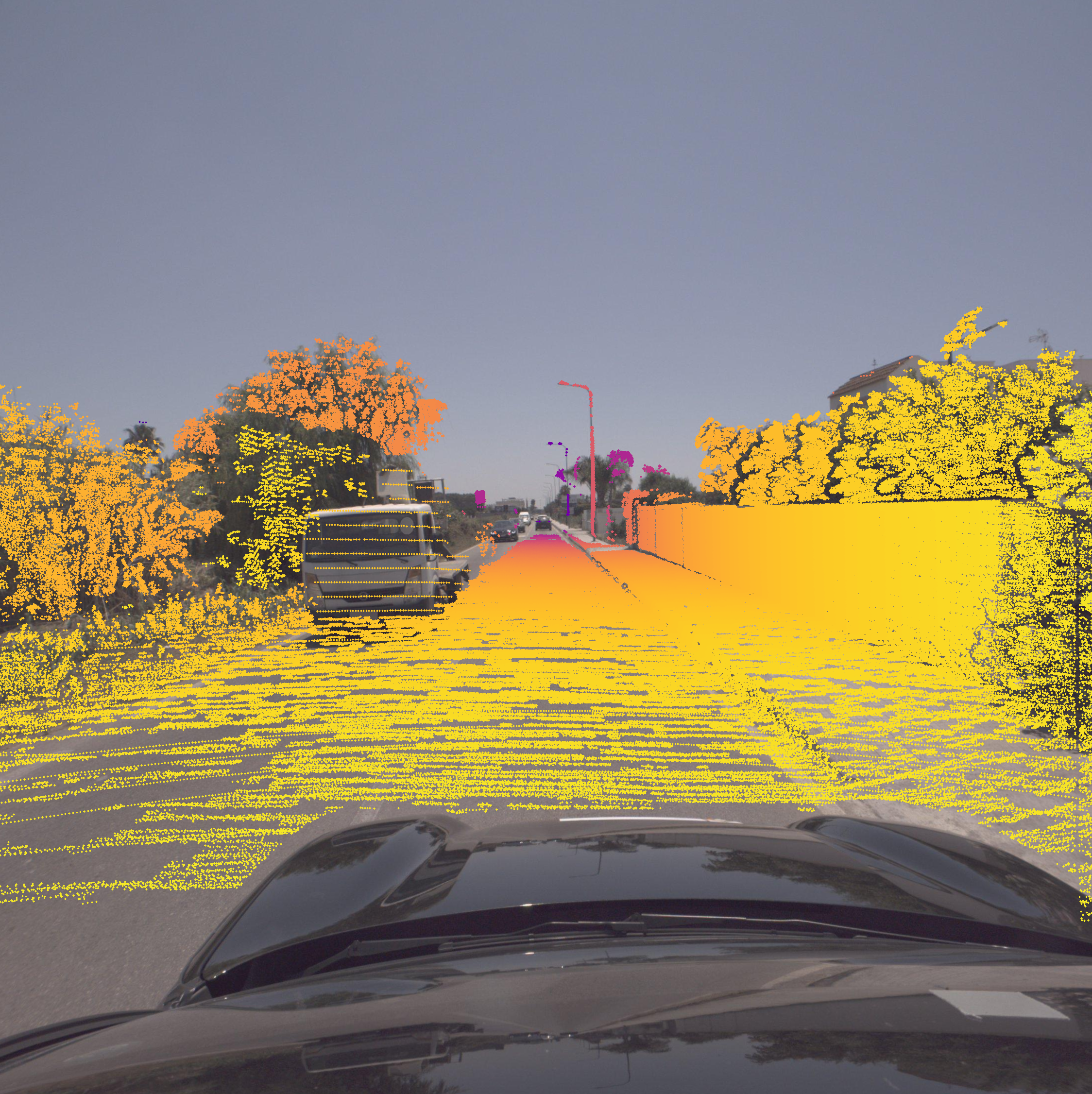}
        \caption{Addition of ICP-based dynamic pruning.}
        \label{fig:abl_6_icp}
    \end{subfigure}\hfill
    \begin{subfigure}[t]{0.24\textwidth}
        \centering
        \includegraphics[width=\linewidth,trim=0 0 0 340,clip]{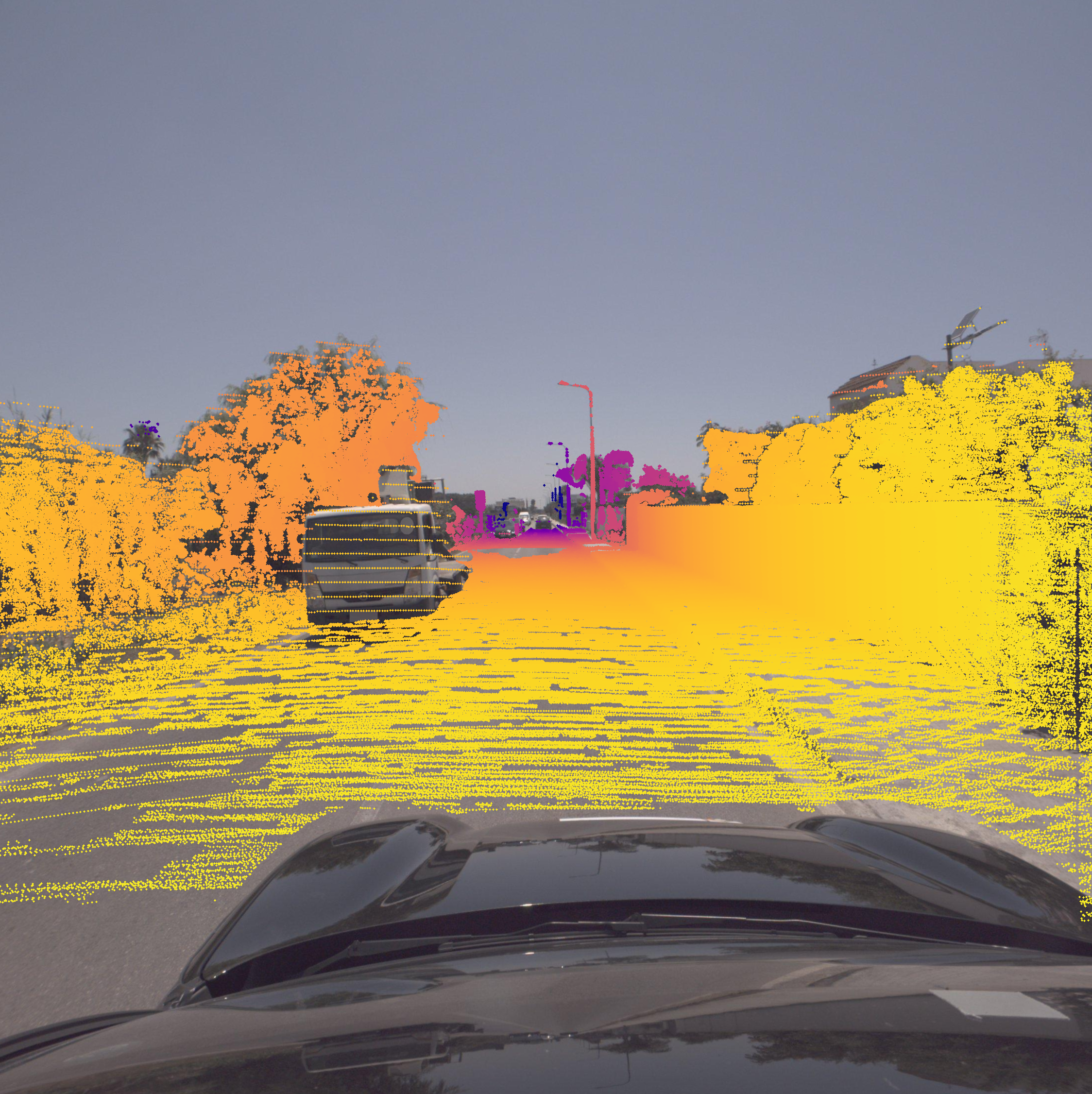}
        \caption{One-stage HPR to recover distant points with a multiplier of 210.}
        \label{fig:abl_6_icp_mono}
    \end{subfigure}\hfill
    \begin{subfigure}[t]{0.24\textwidth}
        \centering
        \includegraphics[width=\linewidth,trim=0 0 0 340,clip]{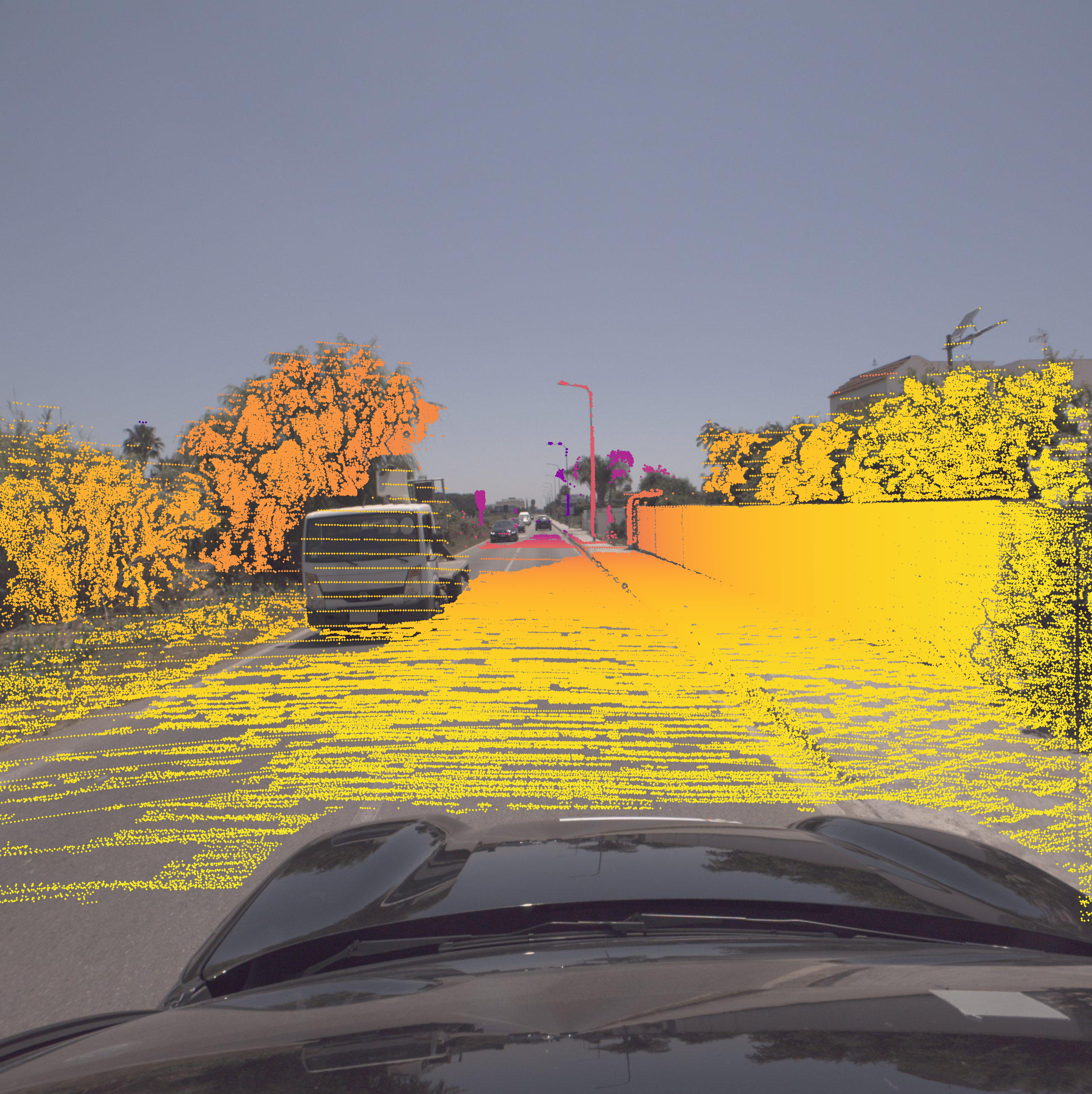}
        \caption{Example of aggressive filtering: Over-strict MDE ($\tau_{\text{rel}}=0.08$).}
        \label{fig:abl_orig_strict_mono}
    \end{subfigure}
    \begin{subfigure}[t]{0.24\textwidth}
        \centering
        \includegraphics[width=\linewidth]{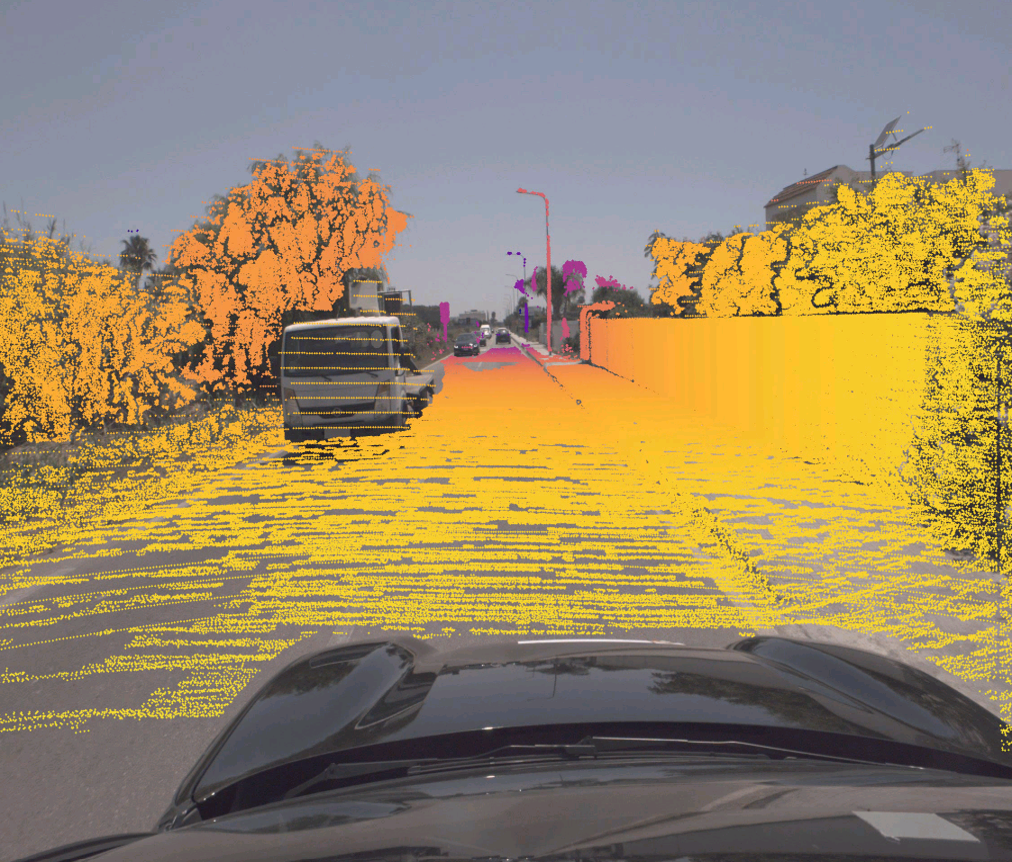}
        \caption{Final Pipeline: Standard output with MDE rejection ($\tau_{\text{rel}}=0.14$).}
        \label{fig:abl_orig}
    \end{subfigure}\hfill

    \caption{
    {Qualitative ablation of the ground-truth aggregation pipeline.} 
    (a)--(c) show the progressive cleanup using cropping and adaptive-voting thresholds. 
    (d) introduces our baseline-adaptive consensus to handle variable point density. 
    (e) adds ICP-based dynamic pruning to remove ghosting from moving objects. 
    (f) demonstrates the two-stage Hidden Point Removal (HPR), which successfully recovers sparse valid points at long range.  
    (g) illustrates a failure case where overly strict monocular consistency checks erode valid road geometry in the distance.
    (h) shows the final configuration used in our main experiments.
    }
    \label{fig:agg_ablation_grid}
\end{figure*}

\section{Annotation Pipeline}

Labeling road-surface irregularities from images is inherently ambiguous. In practice, we observe a variety of structures: speed bumps and humps with varying profiles and markings, elevated circular asphalt patches, and, on the other hand, potholes, cracks, subsidence, and other local depressions. Our main goal in CARD is not fine-grained categorization, but providing bounding boxes that enable height-based evaluation on regions where the road topography deviates from a locally planar surface.

Moreover, the perception of what constitutes a bump versus a mild elevation, or a pothole versus a shallow depression, is subjective and depends on context and surrounding texture. To reduce label noise and keep the annotation protocol tractable, we adopt a coarse but robust strategy with two image-based classes. The positive class represents speed bumps and other elevated structures from the road plane, excluding sidewalks or large structural irregularities. On the other hand, the negative class represents potholes and local depressions. These are annotated as bounding boxes at the image level to support per-object height evaluation.

In addition, CARD contains off-road segments, such as gravel, grass, or unpaved construction areas, where the entire drivable surface is intrinsically irregular and does not admit a stable reference plane. For such scenes, we provide a per-sequence off-road flag rather than per-image bounding boxes, since there is no single, well-defined defect region.

\subsection{Labeling Strategy}

To ensure high-quality ground truth and facilitate a targeted analysis of these road features, we employ a semi-automated pipeline that utilizes a dedicated object detector to verify and refine the irregularities across the dataset.

\subsection{Semi-Automated Dataset Annotation}

We employ a strategy based on YOLOv8~\cite{varghese2024yolov8}, combining a manually annotated seed set with large-scale model-assisted labeling.

\begin{enumerate}[leftmargin=0pt, itemindent=*, itemsep=2pt, topsep=2pt]
    \item \textbf{Initial frame selection.} We first perform a comprehensive manual review of all frames to identify images containing at least one visible road irregularity, positive or negative.
    
    \item \textbf{Manual annotation subset.} From this subset, we randomly sample 40\% for manual annotation. Using~\cite{labelImgRepo}, annotators draw bounding boxes for all clearly visible potholes and speed bumps in this subset.

    \item \textbf{Internal consistency split.} The manually annotated subset is randomly partitioned into training (70\%), validation (20\%), and test (10\%) splits. We use this split to train the verification model and evaluate the consistency of our labeling criteria.

    \item \textbf{Model-assisted labeling.} We train a YOLOv8 detector on the annotated training set and run it on the remaining 60\% of the master list to obtain preliminary bounding boxes. All predictions are subsequently reviewed, and a custom script is used to efficiently correct errors. This semi-automated loop substantially accelerates the annotation process while ensuring the final labels adhere to the definitions established in the seed set.
\end{enumerate}

\subsection{Training Protocol}

We fine-tune YOLOv8~\cite{varghese2024yolov8} on our two-class road-irregularity labels. The model is trained for 150 epochs with an input resolution of 1024$\times$1024 pixels on an NVIDIA RTX 6000 Ada Generation GPU. Training completes in approximately 8.1 hours.

\subsection{Annotation Consistency Analysis}

To validate the quality and learnability of our ground truth, we evaluate the detector on the validation and test splits using standard metrics. High detection scores in this context indicate that the visual definitions of "positive" and "negative" irregularities are distinct and consistently applied across the dataset.

\begin{table*}[t]
    \centering
    \setlength{\tabcolsep}{4pt}
    \renewcommand{\arraystretch}{1.05}

    % (a) Validation set
    \begin{subtable}[t]{\textwidth}
        \centering
        \begin{tabular}{@{}lcccccc@{}}
            \toprule
            Class &  Instances &
            Precision & Recall & mAP@.50 &
            \shortstack{mAP@\\.50--.95} \\
            \midrule
            All classes            & 1{,}725 & 0.932 & 0.903 & 0.952 & 0.804 \\
            Negative        &   970   & 0.897 & 0.839 & 0.919 & 0.793 \\
            Positive     &   755   & 0.968 & 0.967 & 0.986 & 0.814 \\
            \bottomrule
        \end{tabular}
        \caption{Validation set.}
        \label{tab:yolo_val_performance}
    \end{subtable}

    \vspace{0.8em}

    % (b) Test set
    \begin{subtable}[t]{\textwidth}
        \centering
        \begin{tabular}{@{}lcccccc@{}}
            \toprule
            Class  & Instances &
            Precision & Recall & mAP@.50 &
            \shortstack{mAP@\\.50--.95} \\
            \midrule
            All classes              &   870  & 0.878 & 0.912 & 0.935 & 0.793 \\
            Negative       &   521  & 0.793 & 0.833 & 0.876 & 0.755 \\
            Positive    &   349  & 0.964 & 0.991 & 0.994 & 0.831 \\
            \bottomrule
        \end{tabular}
        \caption{Test set.}
        \label{tab:yolo_test_performance}
    \end{subtable}

    \caption{Verification of label consistency: YOLOv8 detection metrics on (a) validation and (b) test splits.}
    \label{tab:yolo_overall_performance}
\end{table*}

The results on the held-out test set (Table~\ref{tab:yolo_test_performance}) demonstrate a high degree of label consistency. The model achieves an mAP@.50 of 0.994 for the positive class, confirming that elevated structures (speed bumps) are annotated with high precision and low ambiguity. 

For the negative class (potholes), the model achieves an mAP@.50 of 0.876. This performance confirms that the distinct visual features of potholes are learnable, the slight gap compared to the positive class reflects the inherent subjectivity in defining the boundary between a pothole and a minor depression, as noted in our annotation protocol. 
\begin{figure}[t]
    \centering

    % Top row: pothole detections
    \begin{subfigure}[t]{0.48\columnwidth}
        \centering
        \includegraphics[width=\linewidth]{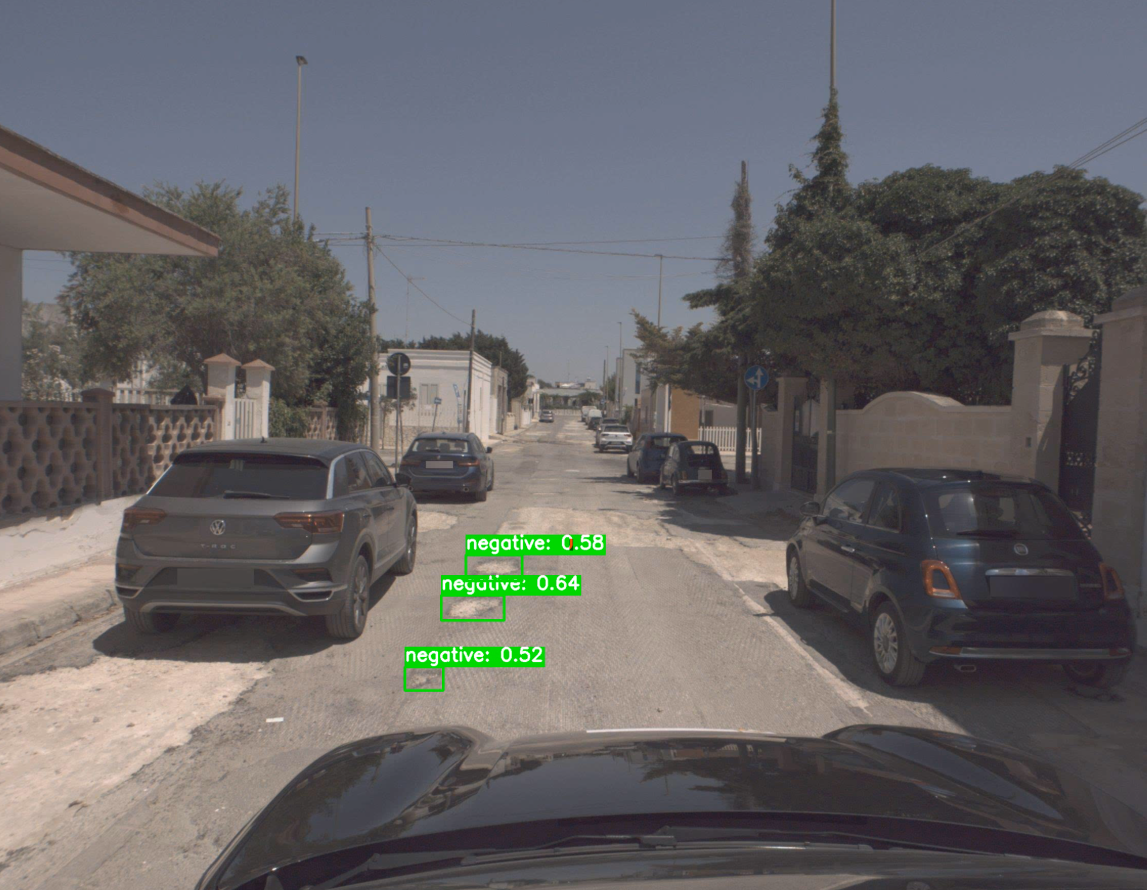}
        \caption{Potholes in sunny conditions.}
        \label{fig:yolo_pothole_1}
    \end{subfigure}
    \hfill
    \begin{subfigure}[t]{0.48\columnwidth}
        \centering
        % Crop from the top: trim = left bottom right top
        \includegraphics[width=\linewidth,trim=0 0 0 198,clip]{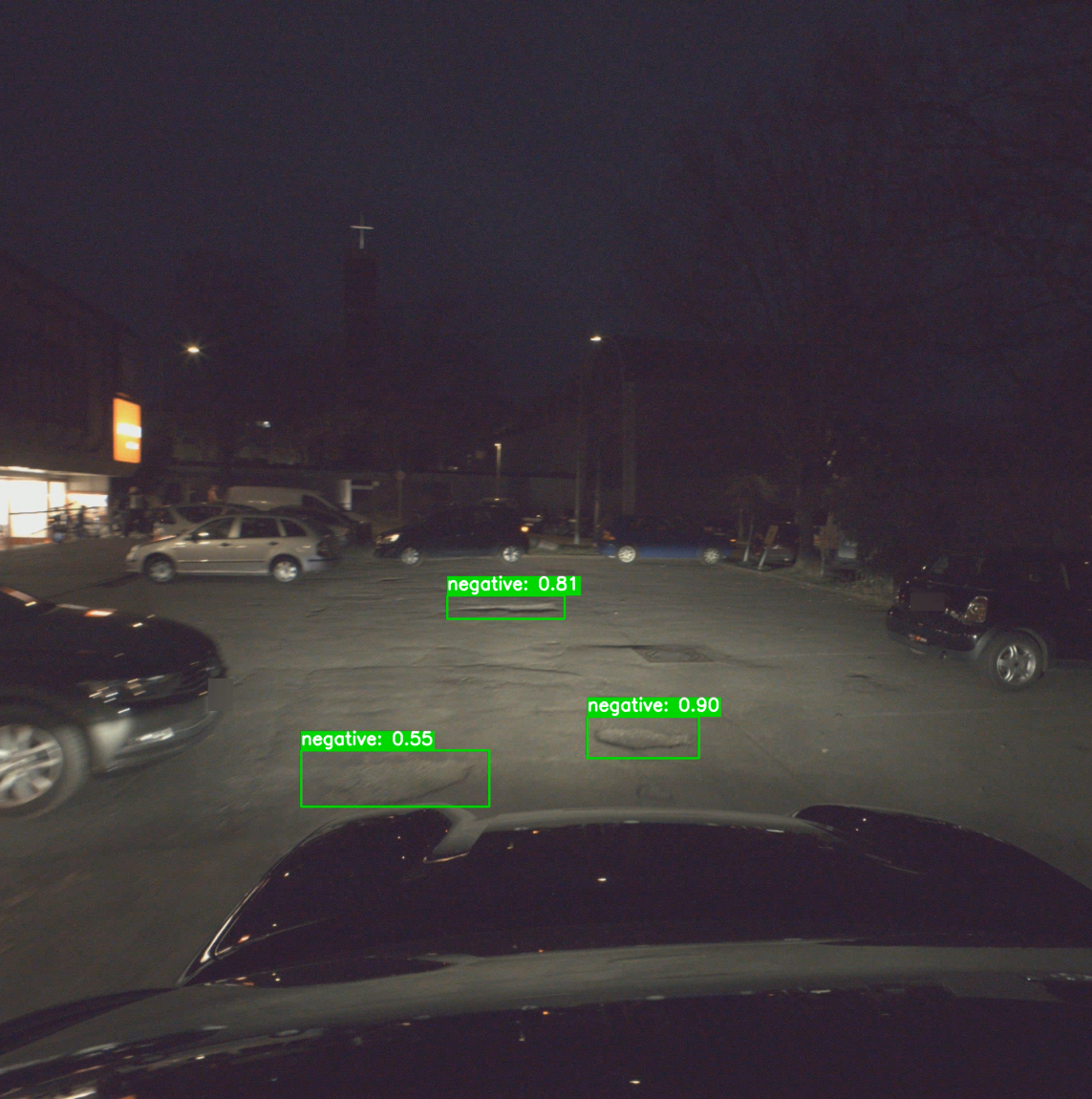}
        \caption{Potholes at night conditions.}
        \label{fig:yolo_pothole_2}
    \end{subfigure}

    \vspace{0.4em}

    % Bottom row: speed bump detections
    \begin{subfigure}[t]{0.48\columnwidth}
        \centering
        \includegraphics[width=\linewidth,trim=0 0 0 33,clip]{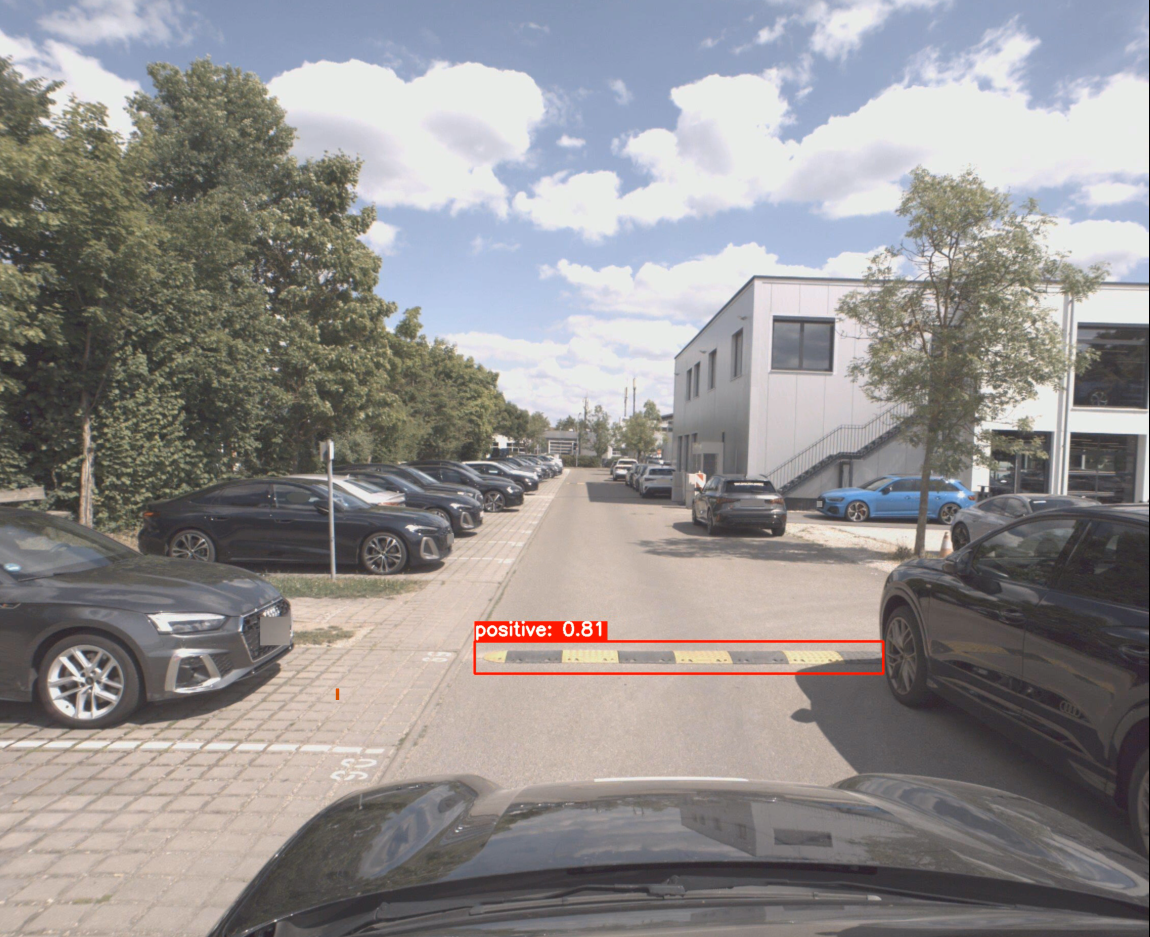}
        \caption{Marked speed bump.}
        \label{fig:yolo_bump_1}
    \end{subfigure}
    \hfill
    \begin{subfigure}[t]{0.48\columnwidth}
        \centering
        % Crop from the top again
        \includegraphics[width=\linewidth,trim=0 0 0 198,clip]{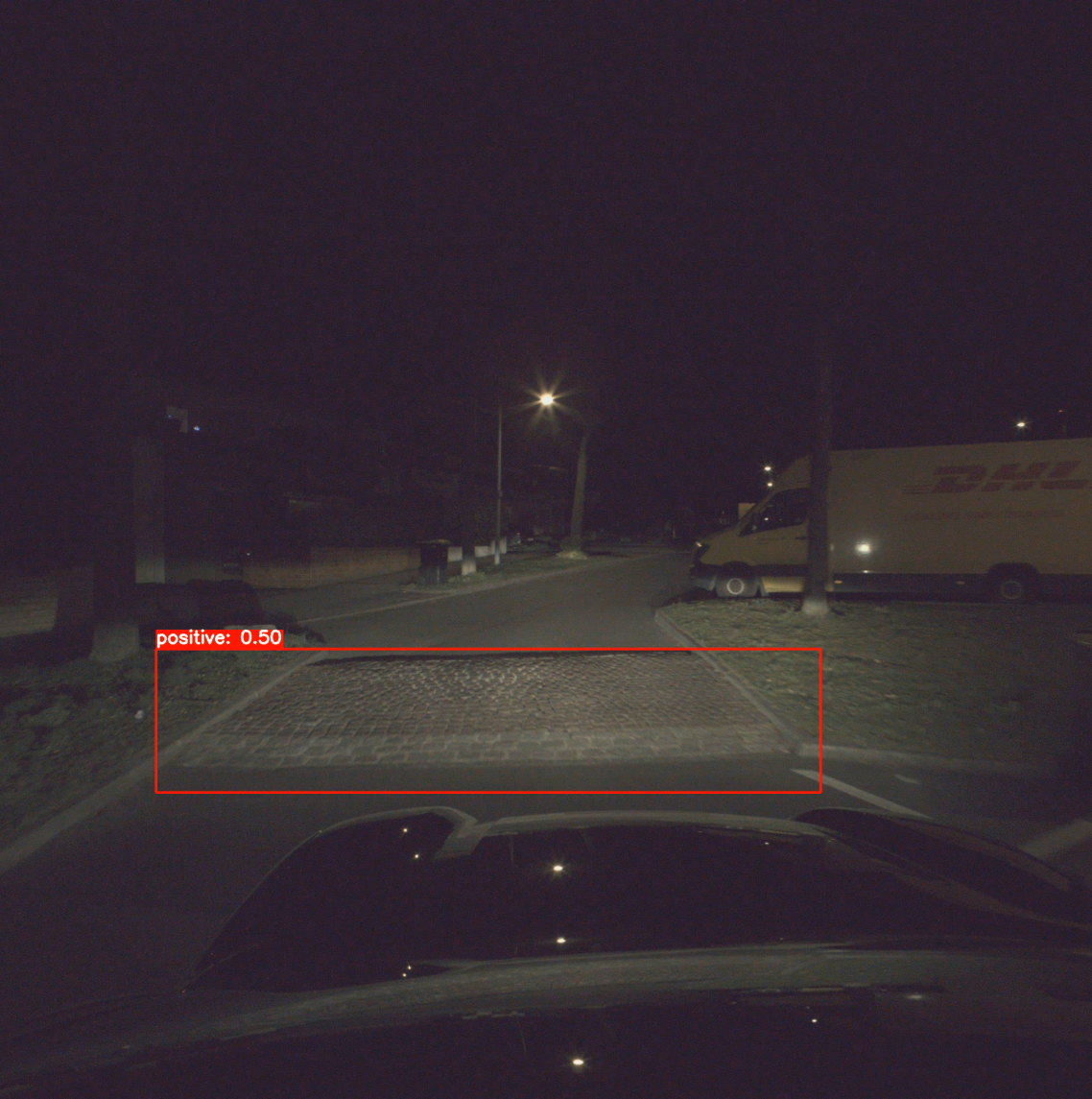}
        \caption{Brick speed bump at night.}
        \label{fig:yolo_bump_2}
    \end{subfigure}

    \caption{Qualitative YOLOv8 detections for potholes (negative class) and speed bumps (positive class) under diverse illumination conditions.}
    \label{fig:yolo_qualitative_results}
\end{figure}

\section{Quantitative Results}
\label{supplementary_sec:quantative_results}

All quantitative evaluations presented in this supplementary material are computed using the largest version of each baseline model. We report metrics for both depth and height estimation across all scenes, as well as specific per-bounding-box evaluations.

All metrics are computed over a valid pixel set $\mathcal{V}$. Let $d_i$ and $\hat{d}_i$ denote the ground truth and predicted depth for pixel $i$, respectively. A pixel $i$ is considered valid ($i \in \mathcal{V}$) if and only if both $d_i$ and $\hat{d}_i$ fall within the sensor range $[d_{\min}, d_{\max}]$, where $d_{\min}=0.1$\,m and $d_{\max}=80.0$\,m.

\subsection{Depth Evaluation}

We employ standard depth estimation metrics:

\begin{itemize}[leftmargin=*]
    \item \textbf{Absolute Relative Error (AbsRel):}
    \begin{equation}
        \text{AbsRel} = \frac{1}{|\mathcal{V}|} \sum_{i \in \mathcal{V}} \frac{|d_i - \hat{d}_i|}{d_i}
    \end{equation}

    \item \textbf{Squared Relative Error (SqRel):}
    \begin{equation}
        \text{SqRel} = \frac{1}{|\mathcal{V}|} \sum_{i \in \mathcal{V}} \frac{\|d_i - \hat{d}_i\|^2}{d_i}
    \end{equation}

    \item \textbf{Root Mean Squared Error (RMSE):}
    \begin{equation}
        \text{RMSE} = \sqrt{\frac{1}{|\mathcal{V}|} \sum_{i \in \mathcal{V}} \|d_i - \hat{d}_i\|^2}
    \end{equation}

    \item \textbf{RMSE (log):}
    \begin{equation}
        \text{RMSE}_{\log} = \sqrt{\frac{1}{|\mathcal{V}|} \sum_{i \in \mathcal{V}} \|\log d_i - \log \hat{d}_i\|^2}
    \end{equation}

    \item \textbf{Threshold Accuracy ($\delta_n$):}
    The percentage of pixels where the ratio between prediction and ground truth is below a threshold:
    \begin{equation}
        \delta_n = \frac{1}{|\mathcal{V}|} \sum_{i \in \mathcal{V}} \mathbb{I} \left( \max \left( \frac{d_i}{\hat{d}_i}, \frac{\hat{d}_i}{d_i} \right) < 1.25^n \right)
    \end{equation}
    We report results for $n \in \{1, 2, 3\}$.
\end{itemize}

\subsection{Height Evaluation Metrics}
To evaluate road irregularities, we convert absolute depth maps into height maps relative to the road surface for the current vehicle position. For each timestamp, we recover the 3D positions of the four wheels' ground contact points from the calibrated sensor rig and wheel–excitation signals (see \cref{sec:calibration,supplementary_sec:wheel_excitation}). In the vehicle frame, we construct the local road plane as the unique plane spanned by these four contact points, which serves as a physically meaningful reference surface.

Both ground-truth and predicted depth maps are back-projected into 3D and transformed into this common vehicle frame. For each 3D point, we compute its signed distance $h$ to the corresponding per-frame road plane, positive values indicate points above the road surface. This yields dense height maps that directly encode deviations of the road geometry with respect to the nominal road surface. As evaluation metrics, we report:

\begin{itemize}[leftmargin=*]
    \item \textbf{Height Absolute Difference (AbsDiff):}
    The mean absolute difference between the predicted and ground truth height:
    \begin{equation}
        \text{AbsDiff} = \frac{1}{|\mathcal{V}|} \sum_{i \in \mathcal{V}} |h_i - \hat{h}_i|
    \end{equation}

    \item \textbf{Height RMSE:}
    The root mean squared error of the height deviation:
    \begin{equation}
        \text{RMSE}_{h} = \sqrt{\frac{1}{|\mathcal{V}|} \sum_{i \in \mathcal{V}} (h_i - \hat{h}_i)^2}
    \end{equation}

    \item \textbf{Height Accuracy ($\delta @ \tau$):}
    The fraction of pixels where the absolute height error is within a specific threshold $\tau$:
    \begin{equation}
        \delta @ \tau = \frac{1}{|\mathcal{V}|} \sum_{i \in \mathcal{V}} \mathbb{I} \left( |h_i - \hat{h}_i| < \tau \right)
    \end{equation}
    We evaluate at thresholds $\tau = 10$\,cm and $\tau = 25$\,cm.
\end{itemize}

\subsection{Fine-tuning Analysis on CARD}
\label{subsec:unidepth_finetune}

\begin{table*}[t]
\centering
\small
\setlength{\tabcolsep}{4pt}
\begin{tabularx}{\linewidth}{@{} l *{14}{Y} @{}}
\toprule
\multirow{2}{*}{Method} &
\multicolumn{2}{c}{AbsRel $\downarrow$} &
\multicolumn{2}{c}{SqRel $\downarrow$} &
\multicolumn{2}{c}{RMSE $\downarrow$} &
\multicolumn{2}{c}{RMSElog $\downarrow$} &
\multicolumn{2}{c}{$\delta_1 \uparrow$} &
\multicolumn{2}{c}{$\delta_2 \uparrow$} &
\multicolumn{2}{c}{$\delta_3 \uparrow$} \\
\cmidrule(lr){2-3}\cmidrule(lr){4-5}\cmidrule(lr){6-7}\cmidrule(lr){8-9}
\cmidrule(lr){10-11}\cmidrule(lr){12-13}\cmidrule(lr){14-15}
 & F & B & F & B & F & B & F & B & F & B & F & B & F & B \\
\midrule
UniDepthV2-L$^{\dagger}$
& 0.046 & 0.029 & 0.204 & 0.022 & 1.825 & 0.382 & 0.074 & 0.032 & 0.981 & \textbf{0.992} & 0.995 & \textbf{0.993} & \textbf{0.999} & \textbf{0.993} \\
UniDepthV2-L(Fine-tuned)
& \textbf{0.042} & \textbf{0.027} & \textbf{0.183} & \textbf{0.020} & \textbf{1.749} & \textbf{0.355} & \textbf{0.069} & \textbf{0.030} & \textbf{0.989} & \textbf{0.991} & \textbf{0.997} & \textbf{0.993} & 0.998 & \textbf{0.993} \\
\bottomrule
\end{tabularx}
\caption{
{Impact of Fine-tuning on UniDepthV2 by L1 depth loss.} 
Comparisons of metrics before and after fine-tuning on the CARD training set. We report results for both the Full and per-box metrics. While fine-tuning improves global scene metrics (Full), the per-box metrics targeting road irregularities show diminishing returns, indicating that standard fine-tuning is insufficient to fully resolve local topography. Both models utilized median-scaling in testing.$\dagger$ denotes median-scaled metrics.
}
\label{tab:finetune_ablation}
\end{table*}
To probe whether the limitations of monocular methods are primarily due to domain shift, we fine-tune UniDepth-V2-L~\cite{UniDepth2025_V2} on the CARD training split. We use our dense  depth maps as metric-depth supervision and mask valid points within $[0.05, 80]$\,m. Starting from the checkpoint, we freeze the encoder backbone and optimize only the decoder parameters. The model is trained for 10 epochs with an L1 depth loss and AdamW optimizer (learning rate $1\times10^{-5}$, weight decay $10^{-2}$) with a cosine annealing schedule, using mixed-precision training with batch size $4$. We select the checkpoint with the lowest validation loss on held-out CARD sequences.

\Cref{tab:finetune_ablation} summarizes the effect of this CARD-specific fine-tuning. We observe consistent improvements in global scene metrics, as an example full-scene RMSE from $1.825$ to $1.749$, AbsRel from $0.046$ to $0.042$, indicating that our supervision benefits overall depth quality. However, on local bounding boxes around road irregularities, the gains are marginal. These structures induce only centimetre-scale depth changes, so even perceptually relevant differences correspond to small errors in depth space and remain largely negligible to global metrics. This suggests that resolving fine-grained road topography requires architectures and training strategies explicitly tailored to road geometry, beyond shallow decoder-only fine-tuning. However, as demonstrated in the main paper, incorporating the affine-invariant losses from MoGe~\cite{MoGE_V2,MoGe2025} yields substantial improvements in the topography-specific metrics. Thus, we leave the development of such specialized architectures for road geometry as future work.

% This suggests that resolving fine-grained road topography requires architectures and training strategies explicitly tailored to road geometry, rather than vanilla $L_1$ depth regression or shallow decoder fine-tuning. While incorporating affine-invariant losses from MoGe~\cite{MoGE_V2,MoGe2025} yields substantial improvements, we posit that fully capturing these features necessitates explicit design choices, such as geometric priors or surface-normal constraints. We leave the development of such specialized architectures for future work.

% and a broader exploration of such monocular models is left for future work. We therefore regard high-fidelity monocular reconstruction of road irregularities as an open problem that CARD is well suited to benchmark.

% "This suggests that resolving fine-grained road topography requires architectures and training strategies explicitly tailored to road geometry, rather than simple shallow decoder fine-tuning. However, as demonstrated in the main paper, incorporating the affine-invariant losses from MoGe~\cite{MoGE_V2,MoGe2025} yields substantial improvements in box-level metrics."

% ===== Preamble bits (ensure these exist once) =====
\begin{AutoWideTable}[t]
\captionsetup{font=small}
\scriptsize
\renewcommand{\arraystretch}{1.06}
\label{tab:card_benchmark_full}

% =========================
% A) DEPTH — NO SCALE
% =========================
% =========================
% A) DEPTH — NO SCALE
% =========================
\begin{subtable}[t]{\linewidth}
\centering
\label{tab:depth_no_scale}
\captionsetup{justification=centering,singlelinecheck=false}

\begingroup
\setlength{\tabcolsep}{3pt} % local width control

% Method (fixed) + 14 flexible centered cells fill \linewidth
\begin{tabularx}{\linewidth}{@{} >{\raggedright\arraybackslash}p{1.65cm} *{14}{Y} @{}}
\toprule
\multirow{2}{*}{Method} &
\multicolumn{2}{c}{AbsRel $\downarrow$} &
\multicolumn{2}{c}{SqRel $\downarrow$} &
\multicolumn{2}{c}{RMSE $\downarrow$} &
\multicolumn{2}{c}{RMSElog $\downarrow$} &
\multicolumn{2}{c}{$\delta_1 \uparrow$} &
\multicolumn{2}{c}{$\delta_2 \uparrow$} &
\multicolumn{2}{c}{$\delta_3 \uparrow$} \\
\cmidrule(lr){2-3}\cmidrule(lr){4-5}\cmidrule(lr){6-7}\cmidrule(lr){8-9}
\cmidrule(lr){10-11}\cmidrule(lr){12-13}\cmidrule(lr){14-15}
 & F & B & F & B & F & B & F & B & F & B & F & B & F & B \\
\midrule
DAV2~\cite{DepthAnythingV2_2025}
& 1.537 & 1.456 & 29.122 & 20.811 & 20.264 & 14.493 & 0.926 & 0.895 & 0.008 & 0.001 & 0.025 & 0.003 & 0.085 & 0.017 \\
DepthPro~\cite{DepthPro}
& 0.317 & 0.288 & 2.779 & 1.058 & 6.086 & 3.080 & 0.390 & 0.355 & 0.280 & 0.223 & 0.722 & 0.744 & 0.907 & 0.944 \\
MoGeV2~\cite{MoGE_V2}
& \secbest{0.108} & \secbest{0.092} & \secbest{0.349} & \secbest{0.143} & \secbest{2.440} & \secbest{1.013} & \secbest{0.123} & \secbest{0.089} & \secbest{0.919} & \secbest{0.963} & \secbest{0.993} & \best{0.993} & \best{0.998} & \best{0.993} \\
Metric3DV2~\cite{Metric3D_v2}
& 0.665 & 0.664 & 5.812 & 4.383 & 9.493 & 6.589 & 0.511 & 0.506 & 0.017 & 0.001 & 0.283 & 0.236 & 0.935 & 0.972 \\
UniDepthV2~\cite{UniDepth2025_V2}
& \best{0.078} & \best{0.062} & \best{0.314} & \best{0.074} & \best{2.242} & \best{0.764} & \best{0.099} & \best{0.062} & \best{0.975} & \best{0.992} & \best{0.995} & \best{0.993} & \best{0.998} & \best{0.993} \\
\cmidrule[0.3pt]{1-15}
FS (stereo)~\cite{foundationstereo_2025}
& 0.040 & 0.014 & 0.411 & 0.006 & 2.162 & 0.185 & 0.088 & 0.016 & 0.976 & 0.994 & 0.989 & 0.994 & 0.994 & 0.994 \\
\bottomrule
\end{tabularx}
\endgroup

\caption{{Depth (without median-scaling).} Raw predictions. FS~\cite{foundationstereo_2025} is a stereo model.}
\end{subtable}

\vspace{6pt}

% =========================
% A) DEPTH — WITH SCALE
% =========================
\begin{subtable}[t]{\linewidth}
\centering
\label{tab:depth_with_scale}
\captionsetup{justification=centering,singlelinecheck=false}

\begingroup
\setlength{\tabcolsep}{3pt}

\begin{tabularx}{\linewidth}{@{} >{\raggedright\arraybackslash}p{1.65cm} *{14}{Y} @{}}
\toprule
\multirow{2}{*}{Method} &
\multicolumn{2}{c}{AbsRel $\downarrow$} &
\multicolumn{2}{c}{SqRel $\downarrow$} &
\multicolumn{2}{c}{RMSE $\downarrow$} &
\multicolumn{2}{c}{RMSElog $\downarrow$} &
\multicolumn{2}{c}{$\delta_1 \uparrow$} &
\multicolumn{2}{c}{$\delta_2 \uparrow$} &
\multicolumn{2}{c}{$\delta_3 \uparrow$} \\
\cmidrule(lr){2-3}\cmidrule(lr){4-5}\cmidrule(lr){6-7}\cmidrule(lr){8-9}
\cmidrule(lr){10-11}\cmidrule(lr){12-13}\cmidrule(lr){14-15}
 & F & B & F & B & F & B & F & B & F & B & F & B & F & B \\
\midrule
DAV2$^{\dagger}$~\cite{DepthAnythingV2_2025}
& 0.096 & 0.055 & 0.548 & 0.084 & 4.107 & 0.683 & 0.154 & 0.060 & 0.895 & 0.976 & 0.969 & \secbest{0.989} & \secbest{0.990} & \secbest{0.992} \\
DepthPro$^{\dagger}$~\cite{DepthPro}
& 0.100 & 0.045 & 0.830 & 0.075 & 3.270 & 0.602 & 0.142 & 0.050 & 0.908 & 0.974 & 0.971 & \secbest{0.989} & 0.986 & \best{0.993} \\
MoGeV2$^{\dagger}$~\cite{MoGE_V2}
& \best{0.045} & \best{0.027} & \secbest{0.177} & \best{0.022} & \best{1.807} & \best{0.337} & \best{0.074} & \best{0.029} & \secbest{0.980} & 0.989 & \secbest{0.996} & \best{0.993} & \best{0.999} & \best{0.993} \\
Metric3DV2$^{\dagger}$~\cite{Metric3D_v2}
& 0.060 & 0.039 & \best{0.165} & \secbest{0.034} & 1.836 & 0.474 & 0.085 & 0.042 & 0.976 & \secbest{0.990} & \best{0.997} & \best{0.993} & \best{0.999} & \best{0.993} \\
UniDepthV2$^{\dagger}$~\cite{UniDepth2025_V2}
& \secbest{0.046} & \secbest{0.029} & 0.204 & \best{0.022} & \secbest{1.825} & \secbest{0.382} & \best{0.074} & \secbest{0.032} & \best{0.981} & \best{0.992} & 0.995 & \best{0.993} & \best{0.999} & \best{0.993} \\
\cmidrule[0.3pt]{1-15}
FS (stereo)~\cite{foundationstereo_2025}
& 0.036 & 0.009 & 0.401 & 0.003 & 2.126 & 0.137 & 0.086 & 0.011 & 0.977 & 0.994 & 0.989 & 0.994 & 0.994 & 0.994 \\
\bottomrule
\end{tabularx}
\endgroup

\caption{{Depth (with per-image scaling).} $^{\dagger}$Monocular models median-scaled.}
\end{subtable}

\vspace{8pt}

% =========================
% B) HEIGHT — WITH SCALE ONLY
% =========================
\begin{subtable}[t]{\linewidth}
\centering
\label{tab:height_with_scale}
\captionsetup{justification=centering,singlelinecheck=false}

\begingroup
\setlength{\tabcolsep}{3pt}

% Adjusted to 8 data columns (AbsDiff, RMSE, delta@10, delta@25)
\begin{tabularx}{\linewidth}{@{} >{\raggedright\arraybackslash}p{1.65cm} *{8}{Y} @{}}
\toprule
\multirow{2}{*}{Method} &
\multicolumn{2}{c}{AbsDiff $\downarrow$} &
\multicolumn{2}{c}{RMSE $\downarrow$} &
\multicolumn{2}{c}{$\delta@10$\,cm $\uparrow$} &
\multicolumn{2}{c}{$\delta@25$\,cm $\uparrow$} \\
\cmidrule(lr){2-3}\cmidrule(lr){4-5}\cmidrule(lr){6-7}\cmidrule(lr){8-9}
 & F & B & F & B & F & B & F & B \\
\midrule
DAV2$^{\dagger}$~\cite{DepthAnythingV2_2025}
& 0.181 & 0.106 & \secbest{0.371} & 0.113 & 0.578 & 0.620 & 0.836 & 0.929 \\
DepthPro$^{\dagger}$~\cite{DepthPro}
& 0.249 & 0.086 & 0.817 & 0.092 & 0.630 & 0.726 & 0.839 & 0.944 \\
MoGeV2$^{\dagger}$~\cite{MoGE_V2}
& \secbest{0.119} & \best{0.051} & 0.633 & \best{0.056} & \secbest{0.801} & \best{0.893} & \secbest{0.936} & \secbest{0.987} \\
Metric3DV2$^{\dagger}$~\cite{Metric3D_v2}
& \best{0.117} & 0.074 & \best{0.356} & 0.081 & 0.675 & 0.754 & 0.911 & 0.969 \\
UniDepthV2~\cite{UniDepth2025_V2}
& \best{0.117} & \secbest{0.055} & 0.456 & \secbest{0.061} & \best{0.807} & \secbest{0.860} & \best{0.948} & \best{0.989} \\
\cmidrule[0.3pt]{1-9}
FS (stereo)~\cite{foundationstereo_2025}
& 0.169 & 0.017 & 1.508 & 0.021 & 0.907 & 0.992 & 0.958 & 0.999 \\
\bottomrule
\end{tabularx}
\endgroup

\caption{{Height (with scale).} We report only scaled height metrics (Full vs Boxes). $^{\dagger}$Monocular values use the same per-image scaling as in depth. FS is stereo.}
\end{subtable}

\caption{Comprehensive CARD benchmark. Each subtable reports \emph{Full} (F) and \emph{Boxes} (B) side by side.}
\label{TABLE:FULL_TABLE_BENCHMARK_ALL}
\vspace{-1mm}
\end{AutoWideTable}

\section{Qualitative Results}
\label{supplementary_sec:qualitative_results}

\begin{figure*}[p] 
    \centering
    \setlength{\abovecaptionskip}{3pt}
    \setlength{\belowcaptionskip}{10pt}

    % --- ROW 1 ---
    \begin{subfigure}[t]{0.48\linewidth}
        \centering
        \includegraphics[width=\linewidth]{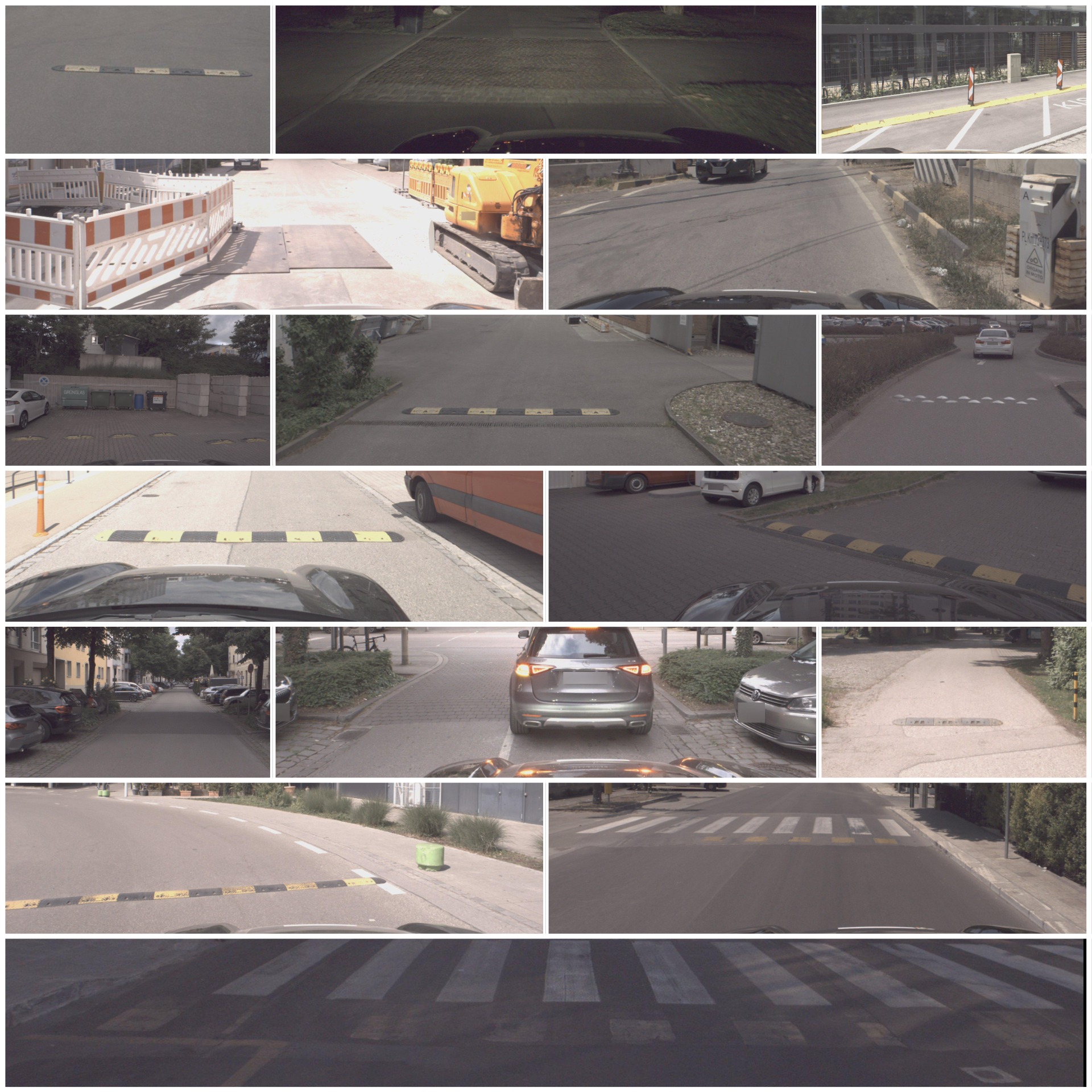}
        \caption{{Positive Irregularities:} Speed bumps, raised manholes, and plateau bumps.}
        \label{fig:gallery_positive}
    \end{subfigure}
    \hfill 
    \begin{subfigure}[t]{0.48\linewidth}
        \centering
        \includegraphics[width=\linewidth]{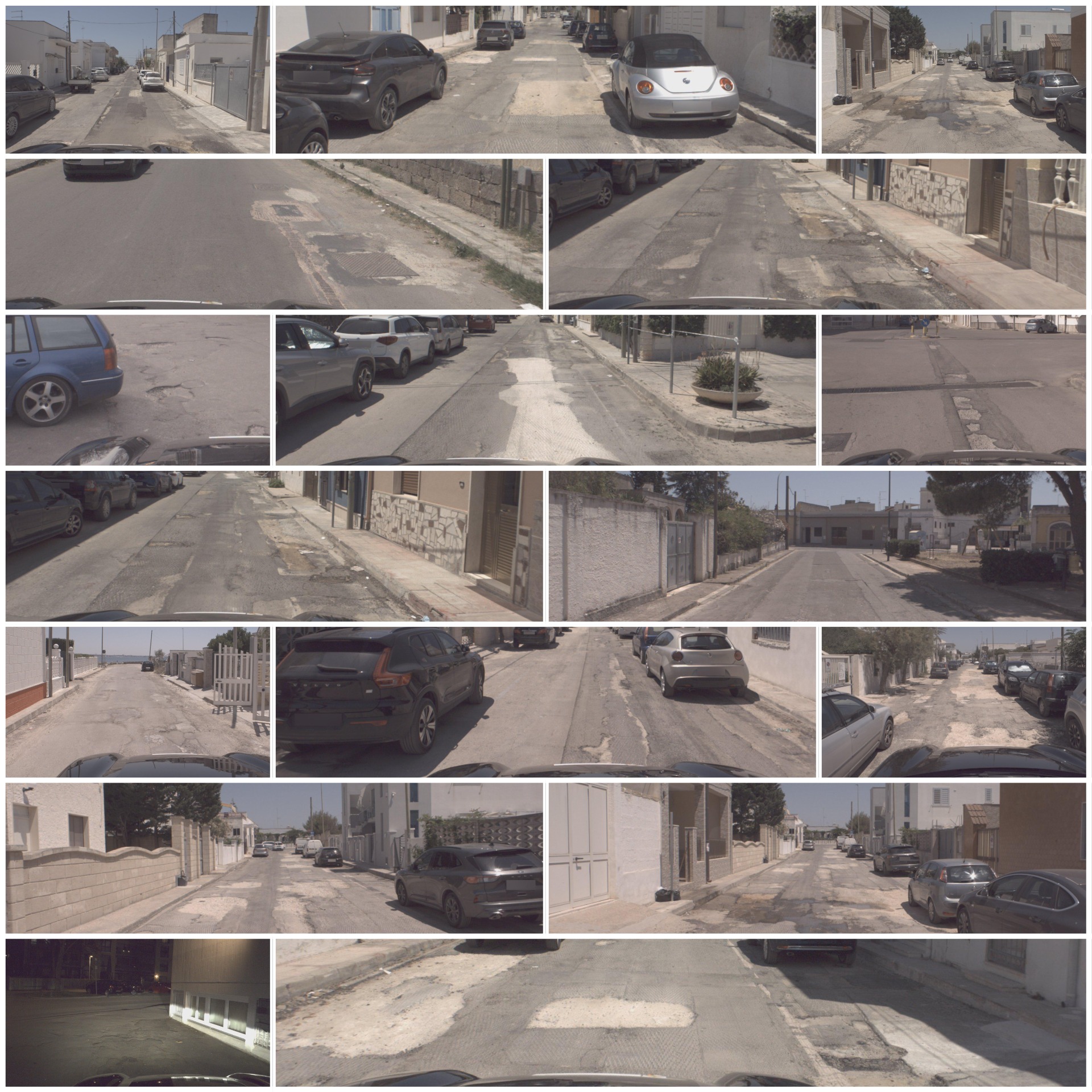}
        \caption{{Negative Irregularities:} Potholes, subsidence, and road cracks.}
        \label{fig:gallery_negative}
    \end{subfigure}

    \vspace{1em} % Vertical gap

    % --- ROW 2 ---
    \begin{subfigure}[t]{0.48\linewidth}
        \centering
        \includegraphics[width=\linewidth]{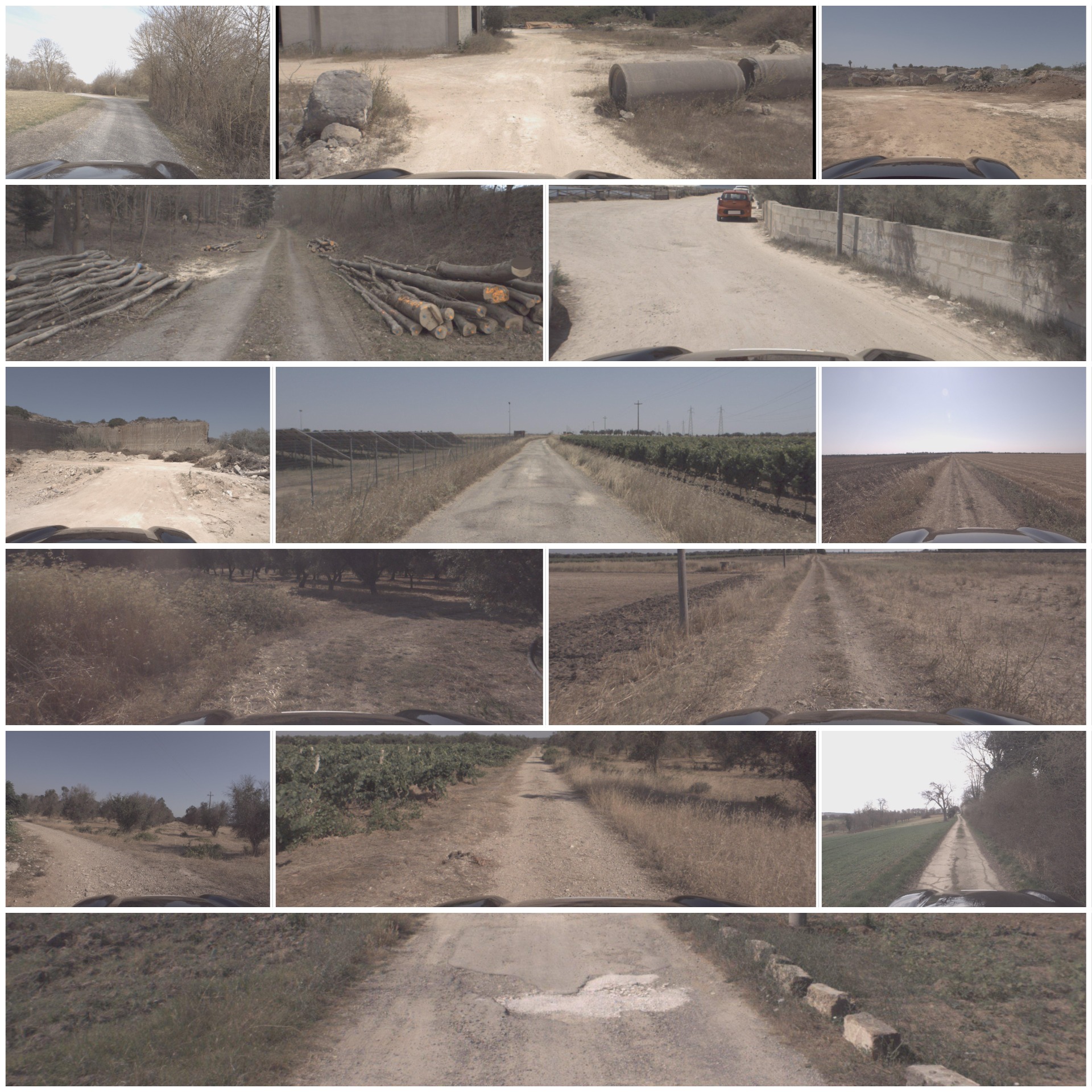}
        \caption{{Off-Road Terrain:} Unpaved surfaces, gravel, and unstructured geometry.}
        \label{fig:gallery_offroad}
    \end{subfigure}
    \hfill 
    \begin{subfigure}[t]{0.48\linewidth}
        \centering
        % This is the "random roads" collage
        \includegraphics[width=\linewidth]{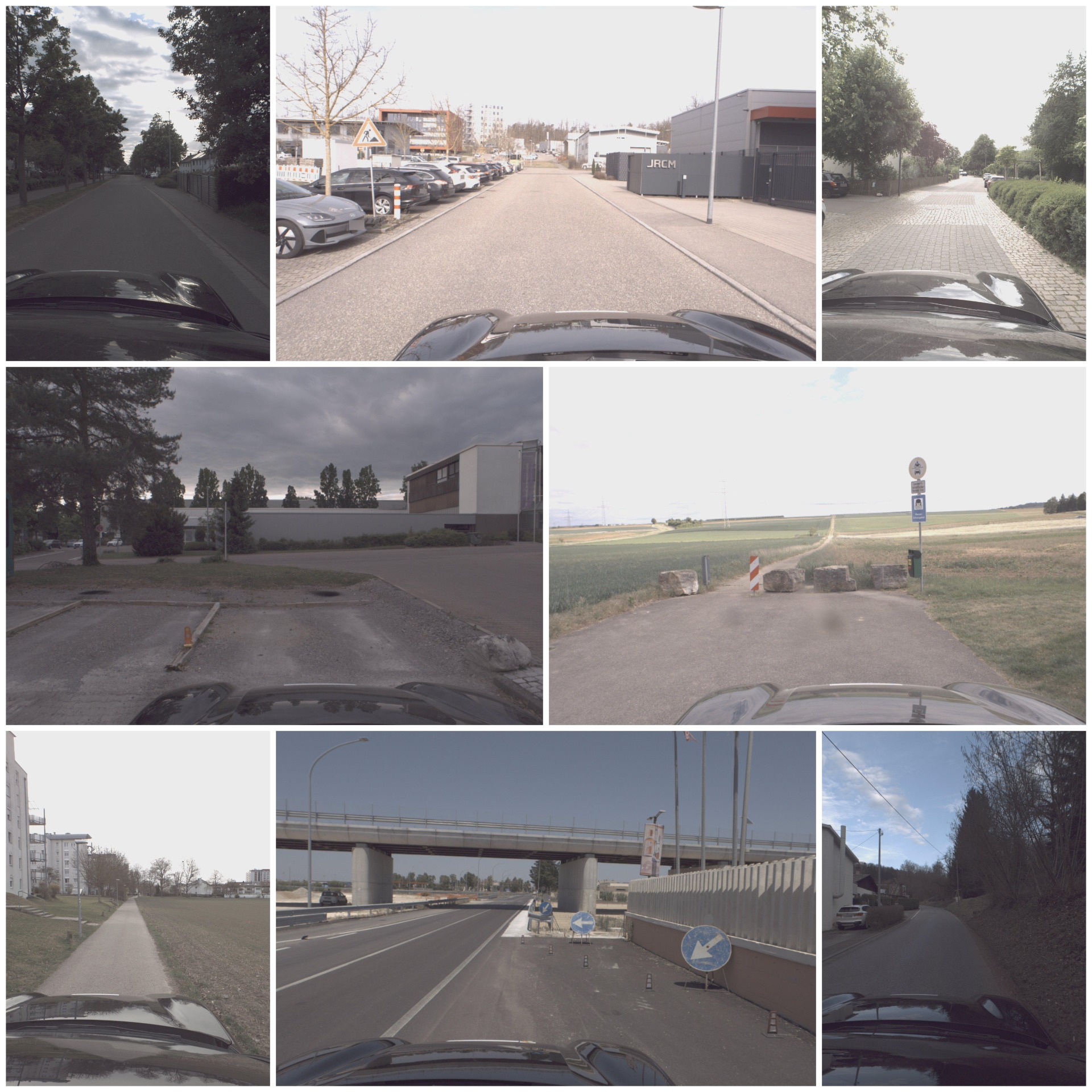}
        \caption{{General Road Scenes:} diverse pavement types and nominal driving conditions.}
        \label{fig:gallery_general}
    \end{subfigure}

    \caption{
    {Diversity of the CARD Dataset.} 
    We display random image crops sampled across the dataset to illustrate variability in geometry and environment. The samples cover (a) positive vertical obstacles, (b) negative surface defects, (c) unstructured off-road terrain, and (d) general road contexts.
    }
    \label{fig:dataset_gallery}
\end{figure*}

\Cref{fig:dataset_gallery} illustrates the visual diversity of the CARD dataset through random image crops, categorizing scenes into positive irregularities, negative irregularities, off-road terrain, and general road contexts.

\begin{figure}[t!] % [t!] places it at the top of the single column
    \centering
    \setlength{\abovecaptionskip}{3pt}
    \setlength{\belowcaptionskip}{-10pt}

    % Replace 'combined_gt_quality.jpg' with your actual single file name
    \includegraphics[width=\linewidth]{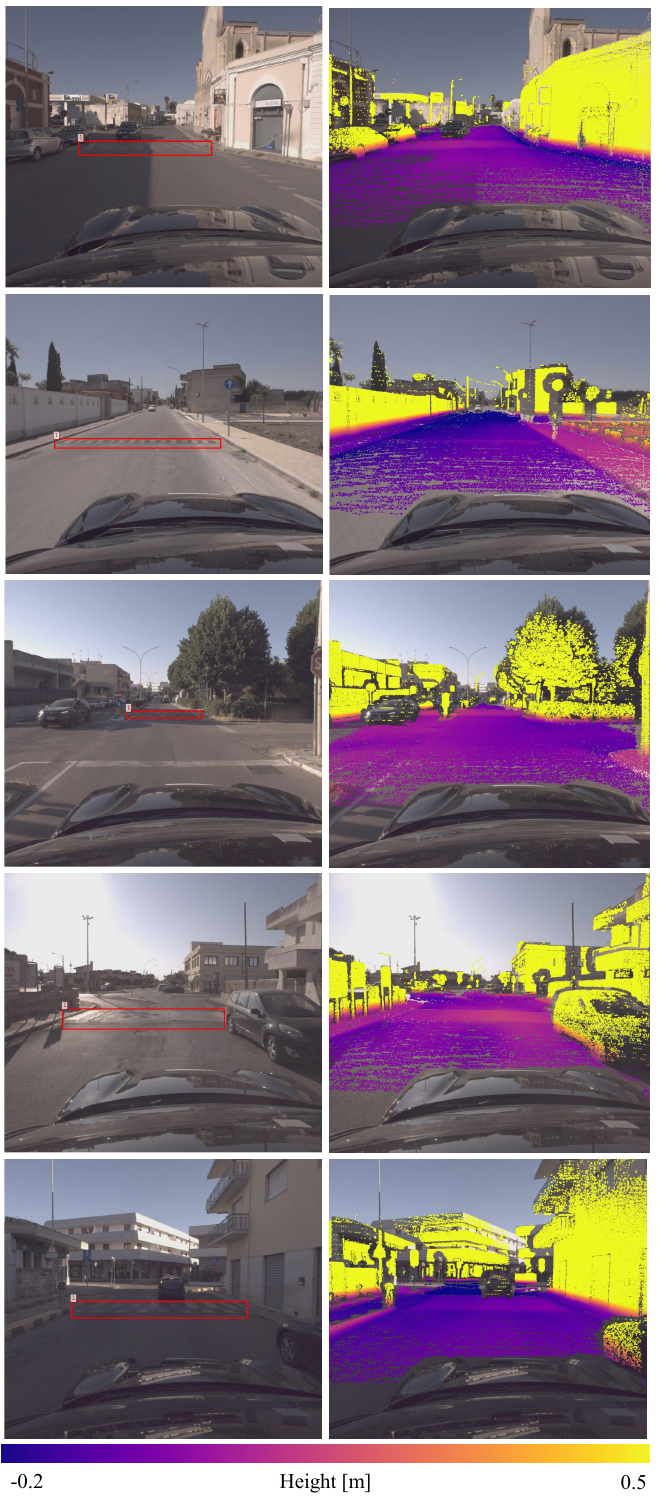}

    \caption{
    {Projected Densified Ground Truth.} 
    We show input RGB images (left) alongside their corresponding projected dense height maps (right).
    }
    \label{fig:gt_quality}
\end{figure}

To highlight the quality of our data, \cref{fig:gt_quality} presents examples of our densified ground truth. As shown, the generated height maps are dense and clearly resolve fine-grained road features, such as speed bumps. Crucially, the aggregation pipeline effectively mitigates artifacts from dynamic objects, resulting in clean representations of the static road surface.

\begin{figure*}[t!]
    \centering
    \setlength{\tabcolsep}{0pt} 
    \setlength{\abovecaptionskip}{4pt} 
    \setlength{\belowcaptionskip}{-5pt} 
    
    % ==========================================
    % SECTION A: CONTEXT (Distributed Full Width)
    % ==========================================
    {Positive Example} \\
    
    % We use 0.97\linewidth to match the width of the rows below
    \begin{minipage}{0.97\linewidth} 
        \centering
        % 3 images spread evenly (approx 0.32 width each)
        \includegraphics[width=0.240\linewidth,trim=0 0 0 50,clip]{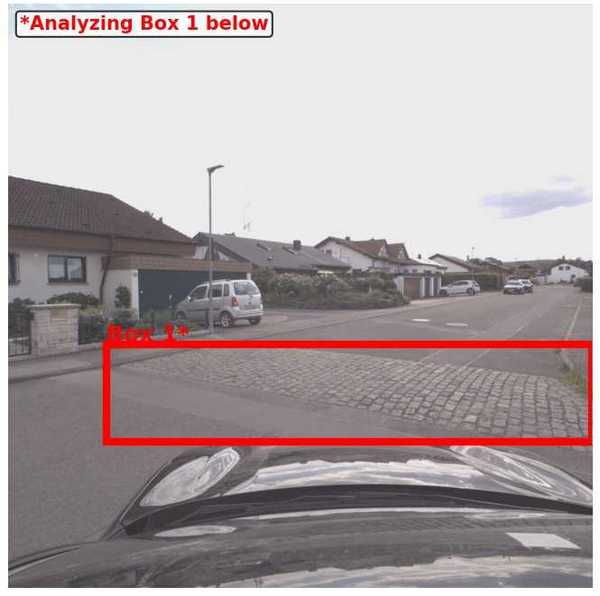} \hfill
        \includegraphics[width=0.24\linewidth]{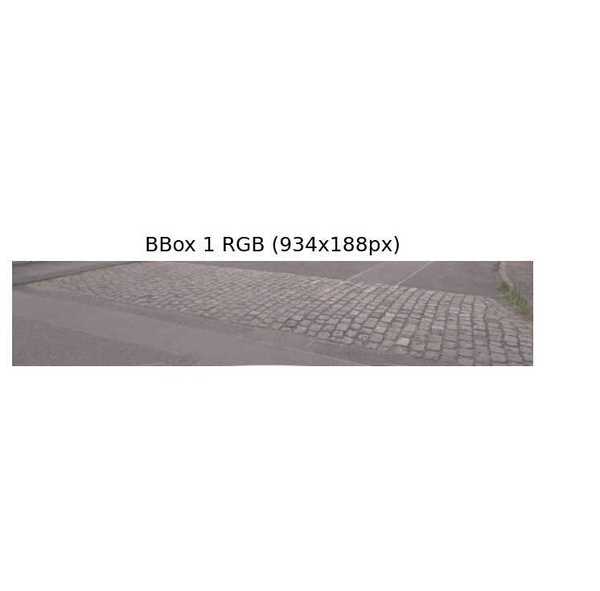} \hfill
        \includegraphics[width=0.24\linewidth]{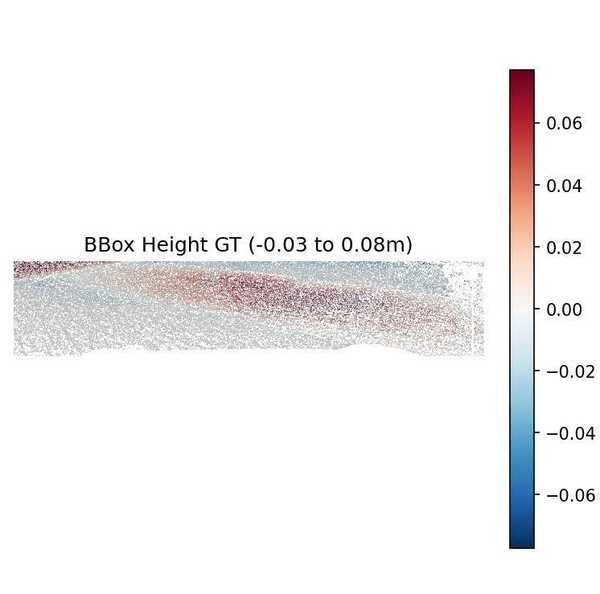} \\
        % Captions spread evenly
        \footnotesize \makebox[0.24\linewidth]{(a) Input RGB} \hfill 
        \makebox[0.24\linewidth]{(b) Box Crop} \hfill 
        \makebox[0.24\linewidth]{(c) GT Height}
    \end{minipage}

    \vspace{1pt} \hrule \vspace{4pt}

    % ==========================================
    % HEADERS
    % ==========================================
    \begin{minipage}{\linewidth}
        \centering \footnotesize \bfseries
        \makebox[0.2\linewidth]{Predicted Depth} \hfill
        \makebox[0.2\linewidth]{Predicted Height} \hfill
        \makebox[0.2\linewidth]{Depth Error} \hfill
        \makebox[0.2\linewidth]{Height Error}
    \end{minipage}
    
    % ==========================================
    % THE GRID (Metric3D, MoGe, UniDepth, FS)
    % ==========================================

    % --- Metric3D ---
    % CHANGE -15pt HERE TO MOVE UP/DOWN
    \raisebox{-15pt}{\rotatebox{90}{\tiny \textbf{Metric3DV2}}} \hspace{2pt}
    \begin{minipage}{0.97\linewidth}
        \includegraphics[width=0.2\linewidth]{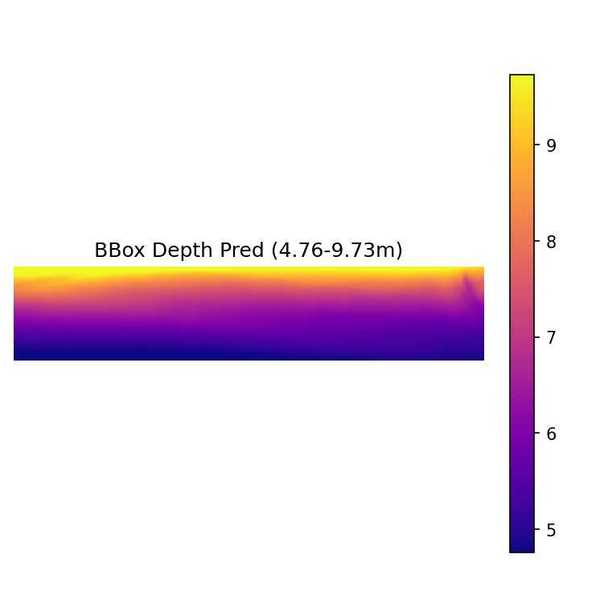} \hfill
        \includegraphics[width=0.2\linewidth]{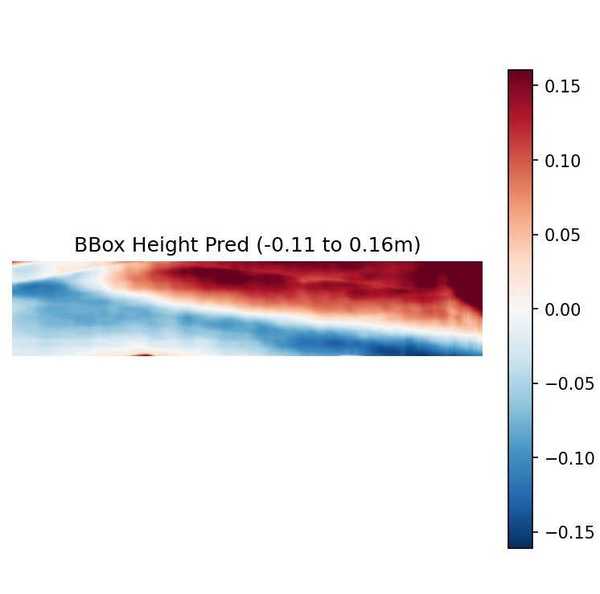} \hfill
        \includegraphics[width=0.2\linewidth]{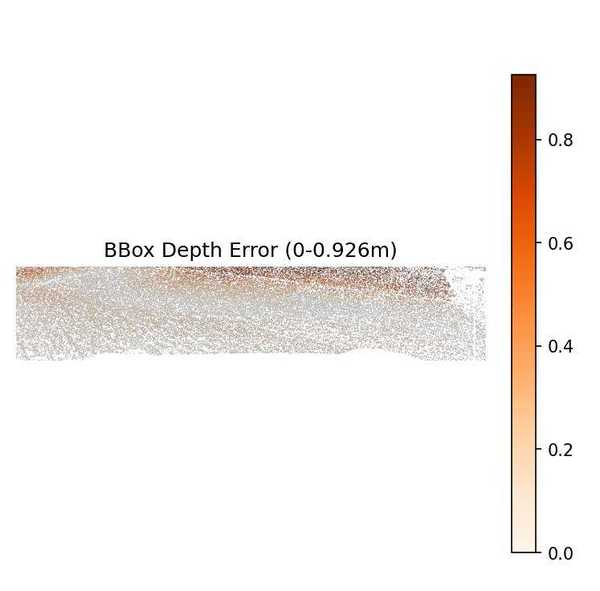} \hfill
        \includegraphics[width=0.2\linewidth]{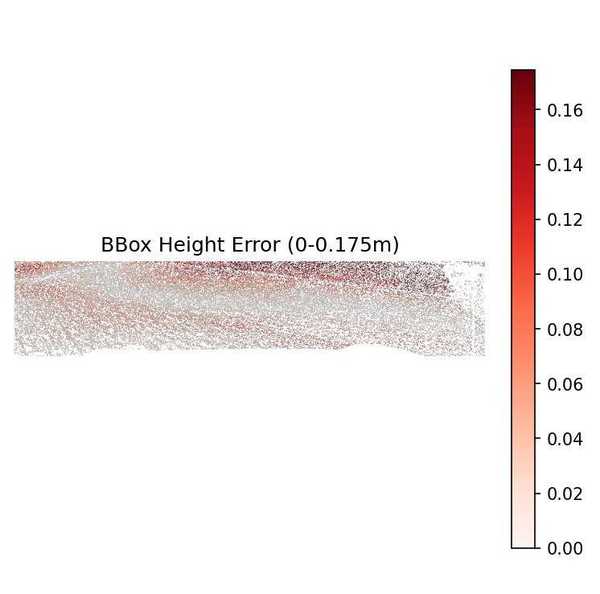}
    \end{minipage}

    % --- MoGe ---
    % CHANGE -15pt HERE TO MOVE UP/DOWN
    \raisebox{-15pt}{\rotatebox{90}{\tiny \textbf{MoGeV2}}} \hspace{2pt}
    \begin{minipage}{0.97\linewidth}
        \includegraphics[width=0.2\linewidth]{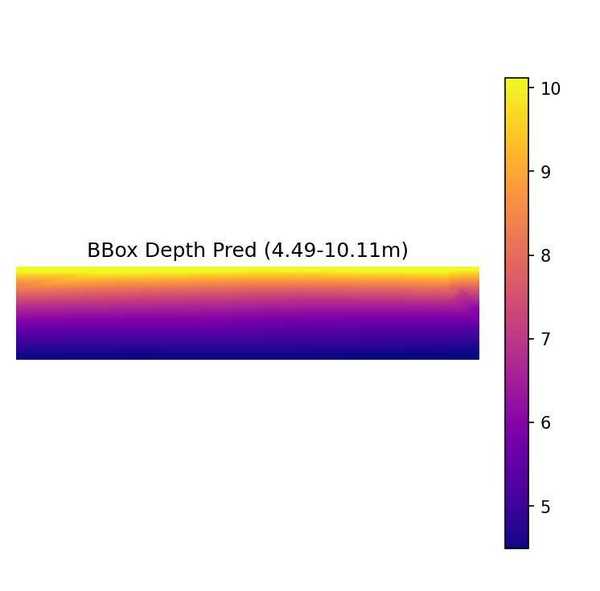} \hfill
        \includegraphics[width=0.2\linewidth]{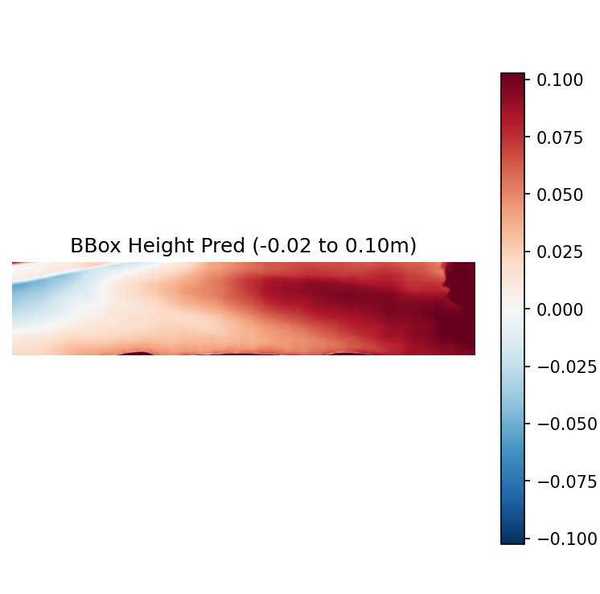} \hfill
        \includegraphics[width=0.2\linewidth]{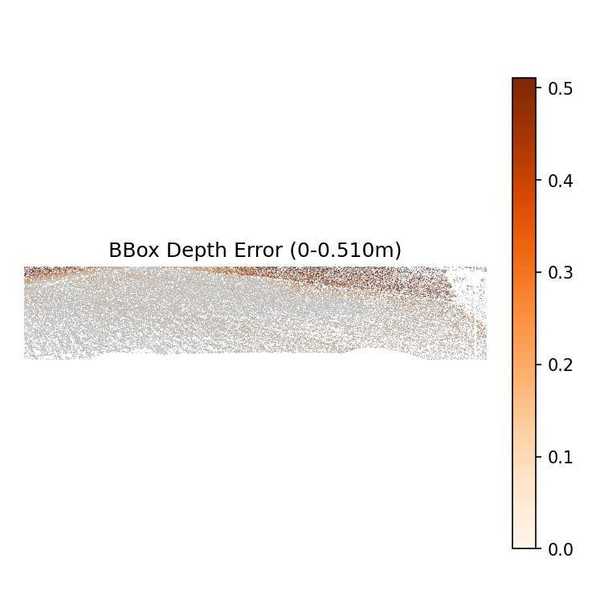} \hfill
        \includegraphics[width=0.2\linewidth]{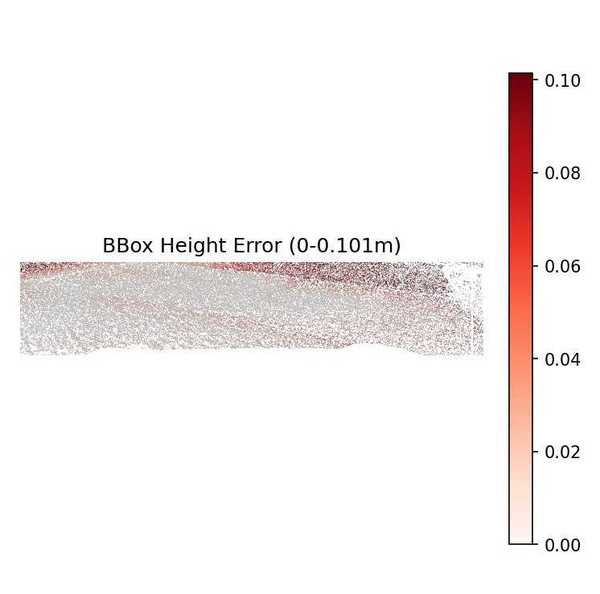}
    \end{minipage}

    % --- UniDepth ---
    % CHANGE -15pt HERE TO MOVE UP/DOWN
    \raisebox{-15pt}{\rotatebox{90}{\tiny \textbf{UniDepthV2}}} \hspace{2pt}
    \begin{minipage}{0.97\linewidth}
        \includegraphics[width=0.2\linewidth]{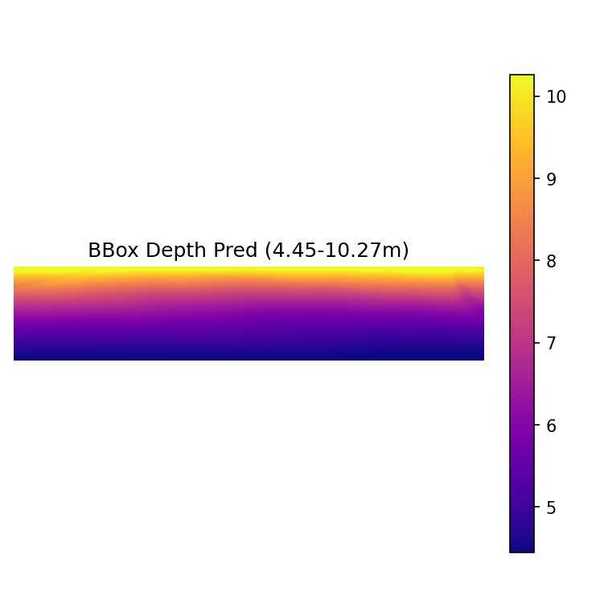} \hfill
        \includegraphics[width=0.2\linewidth]{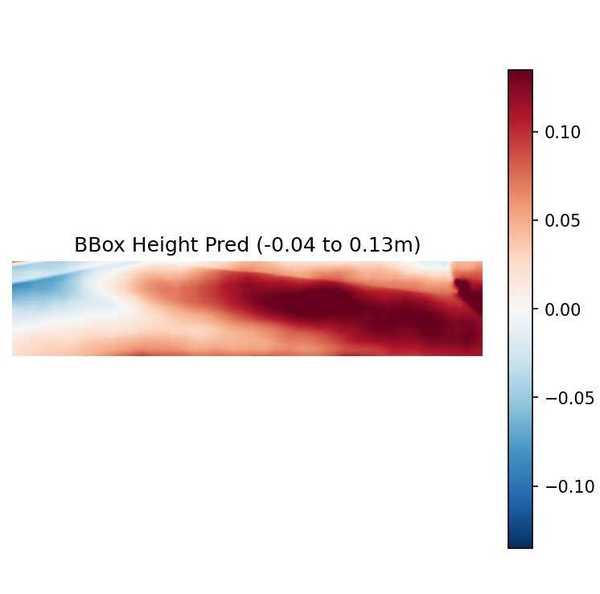} \hfill
        \includegraphics[width=0.2\linewidth]{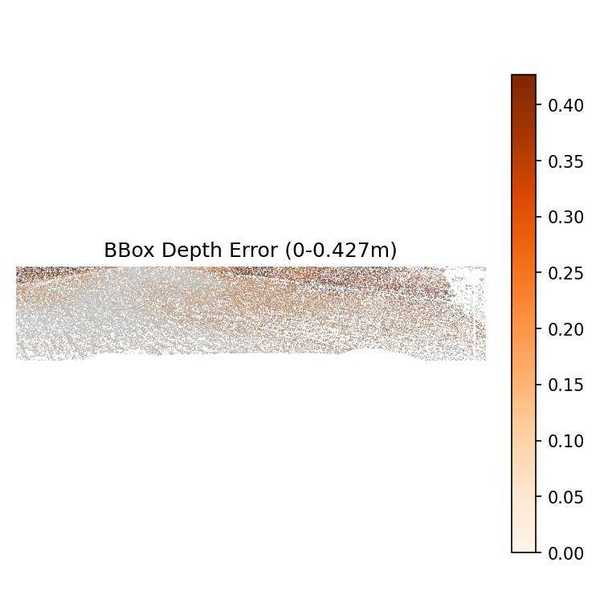} \hfill
        \includegraphics[width=0.2\linewidth]{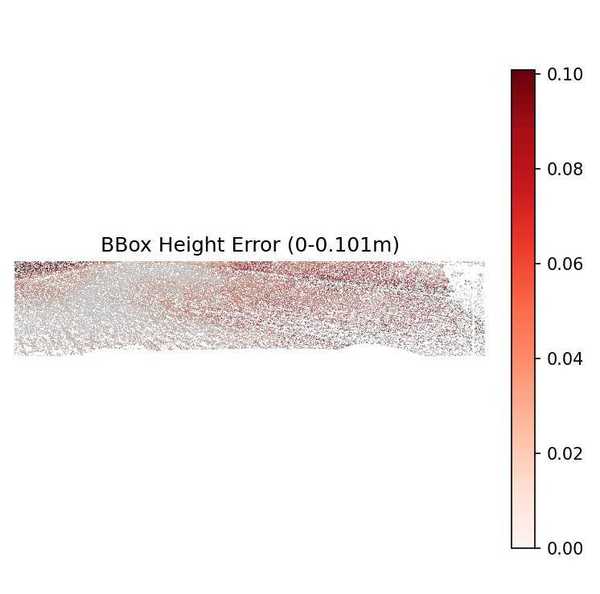}
    \end{minipage}
    
    % --- STEREO ---
    % CHANGE -15pt HERE TO MOVE UP/DOWN
    \raisebox{-15pt}{\rotatebox{90}{\tiny \textbf{FoundationStereo}}} \hspace{2pt}
    \begin{minipage}{0.97\linewidth}
        \includegraphics[width=0.2\linewidth]{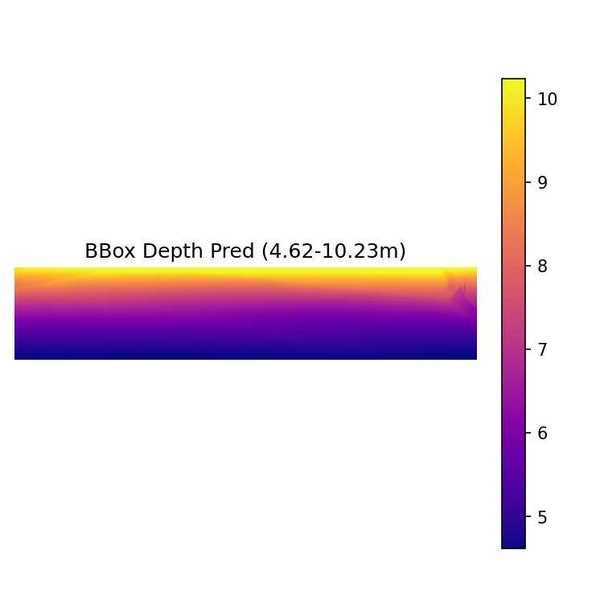} \hfill
        \includegraphics[width=0.2\linewidth]{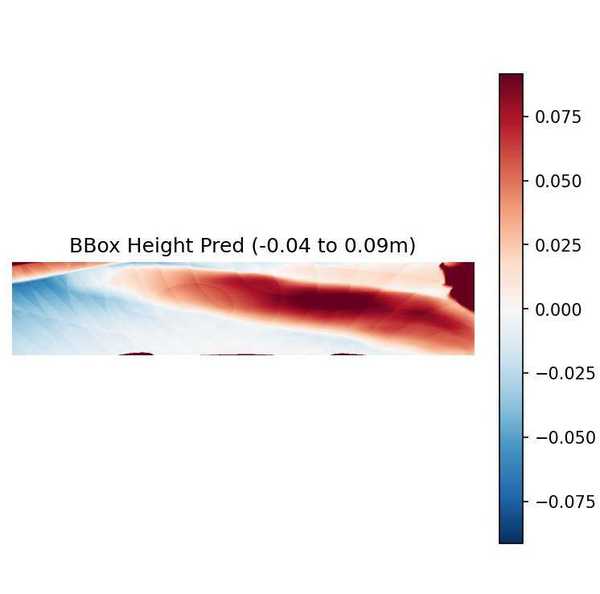} \hfill
        \includegraphics[width=0.2\linewidth]{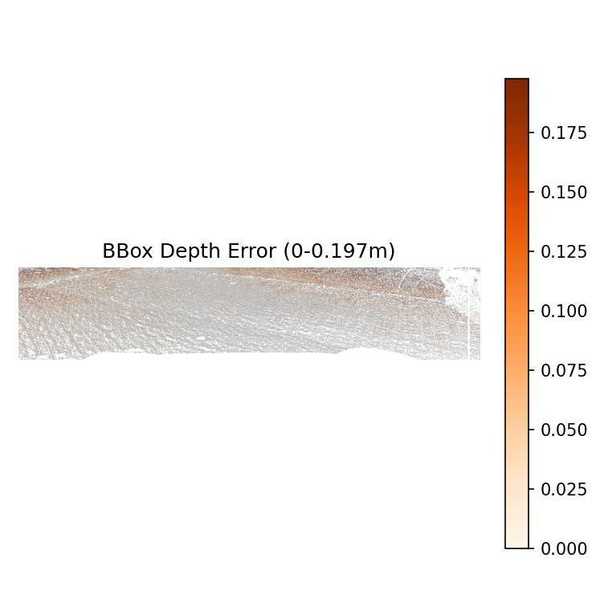} \hfill
        \includegraphics[width=0.2\linewidth]{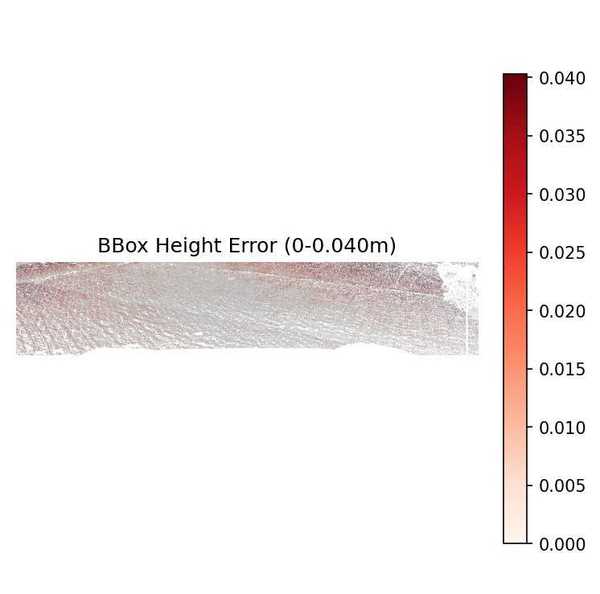}
    \end{minipage}

    \caption{
    \textbf{Qualitative Comparison of a positive example of a speed bump.} 
    Top: Input context and GT geometry. Bottom: Model predictions.
    Monocular baselines~\cite{Metric3D_v2,MoGE_V2,UniDepth2025_V2} (Rows 1--3). FoundationStereo~\cite{foundationstereo_2025} (Bottom Row).
    }
    \label{fig:qual_sample_5777}
\end{figure*}

\begin{figure*}[t!]
    \centering
    \setlength{\tabcolsep}{0pt} 
    \setlength{\abovecaptionskip}{4pt} 
    \setlength{\belowcaptionskip}{-5pt} 
    
    % ==========================================
    % SECTION A: CONTEXT (Distributed Full Width)
    % ==========================================
    \textbf{Negative Example} \\
    
    % We use 0.97\linewidth to match the width of the rows below
    \begin{minipage}{0.97\linewidth} 
        \centering
        % 3 images spread evenly. 
        % Note: Adjusted width to 0.32 to match the makebox text width below
        \includegraphics[width=0.22\linewidth,trim=0 0 0 50,clip]{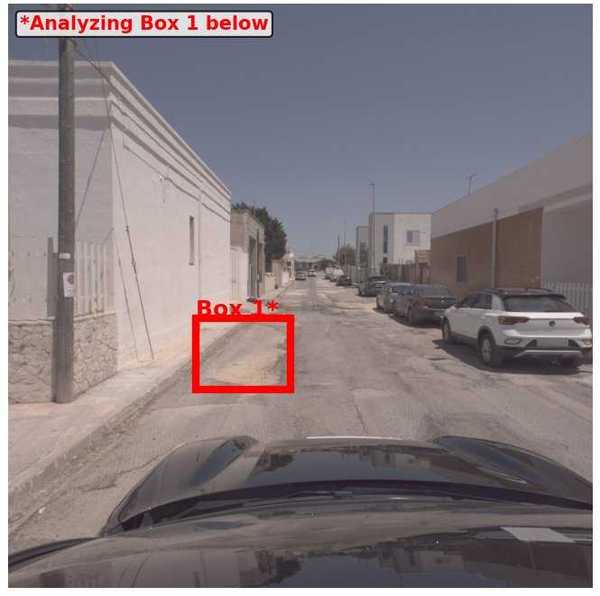} \hfill
        \includegraphics[width=0.22\linewidth]{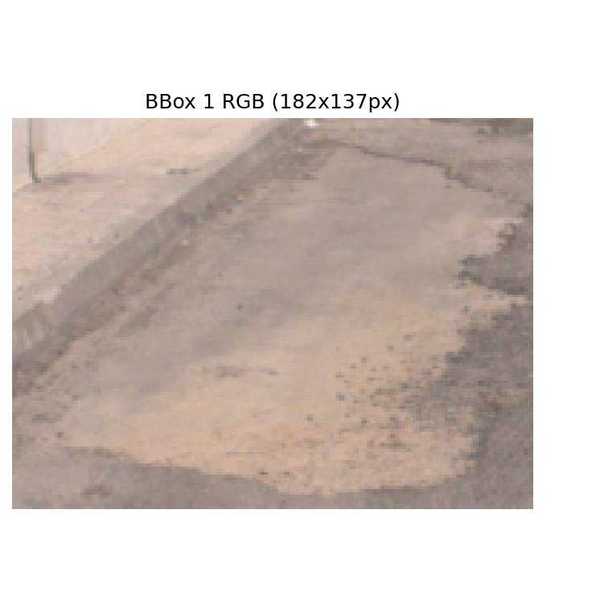} \hfill
        \includegraphics[width=0.22\linewidth]{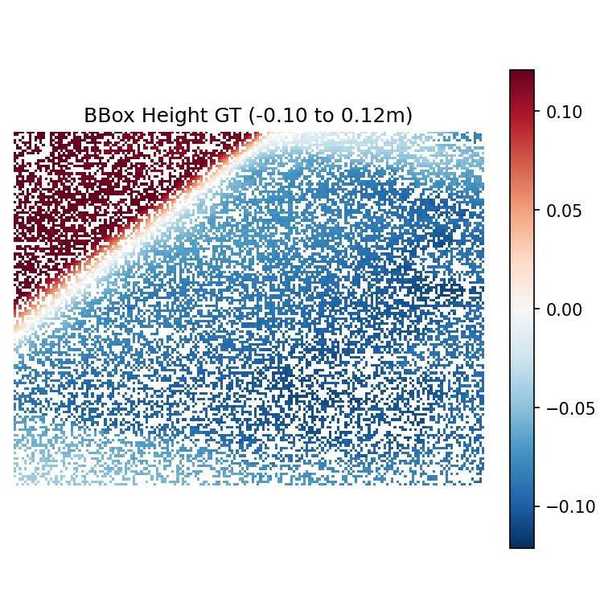} \\
        % Captions spread evenly
        \footnotesize \makebox[0.22\linewidth]{(a) Input RGB} \hfill 
        \makebox[0.22\linewidth]{(b) Box Crop} \hfill 
        \makebox[0.22\linewidth]{(c) GT Height}
    \end{minipage}

    \vspace{1pt} \hrule \vspace{4pt}

    % ==========================================
    % HEADERS
    % ==========================================
    \begin{minipage}{\linewidth}
        \centering \footnotesize \bfseries
        \makebox[0.2\linewidth]{Predicted Depth} \hfill
        \makebox[0.2\linewidth]{Predicted Height} \hfill
        \makebox[0.2\linewidth]{Depth Error} \hfill
        \makebox[0.2\linewidth]{Height Error}
    \end{minipage}
    
    % ==========================================
    % THE GRID (Metric3D, MoGe, UniDepth, FS)
    % ==========================================

    % --- Metric3D ---
    % CHANGE -15pt HERE TO MOVE UP/DOWN
    \raisebox{-15pt}{\rotatebox{90}{\tiny \textbf{Metric3DV2}}} \hspace{2pt}
    \begin{minipage}{0.97\linewidth}
        \includegraphics[width=0.2\linewidth]{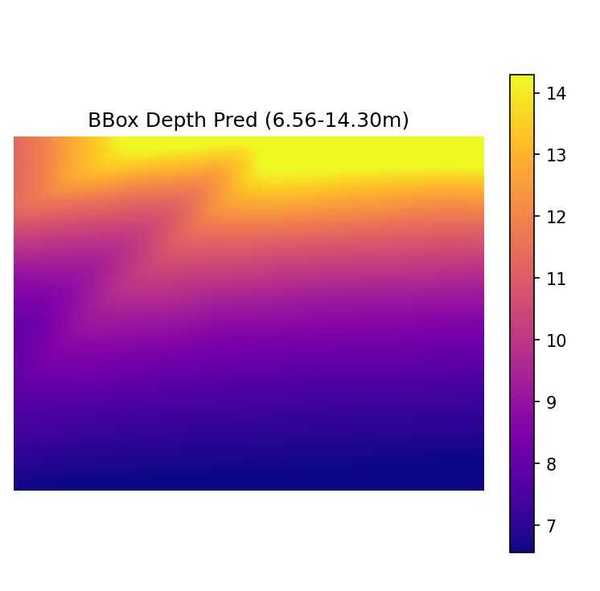} \hfill
        \includegraphics[width=0.2\linewidth]{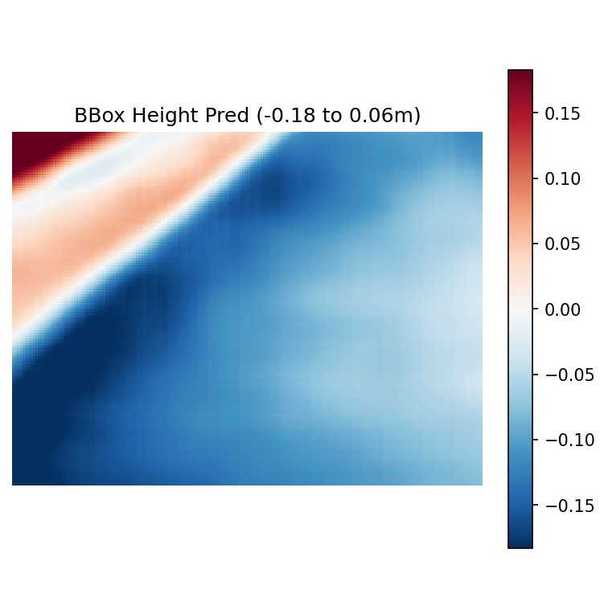} \hfill
        \includegraphics[width=0.2\linewidth]{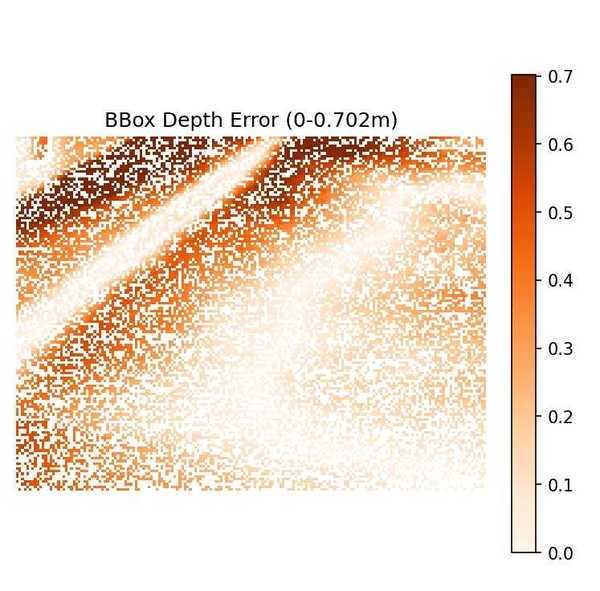} \hfill
        \includegraphics[width=0.2\linewidth]{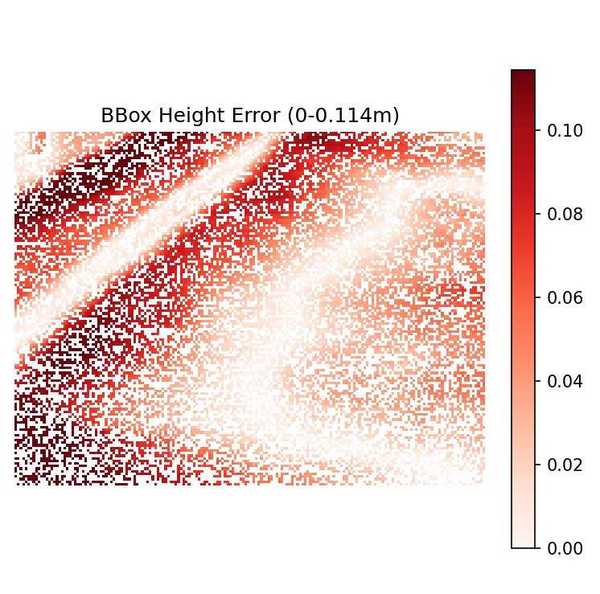}
    \end{minipage}

    % --- MoGe ---
    % CHANGE -15pt HERE TO MOVE UP/DOWN
    \raisebox{-15pt}{\rotatebox{90}{\tiny \textbf{MoGeV2}}} \hspace{2pt}
    \begin{minipage}{0.97\linewidth}
        \includegraphics[width=0.2\linewidth]{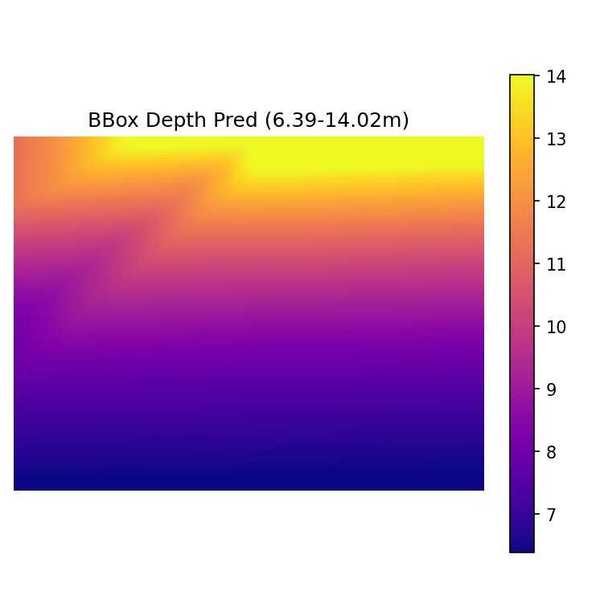} \hfill
        \includegraphics[width=0.2\linewidth]{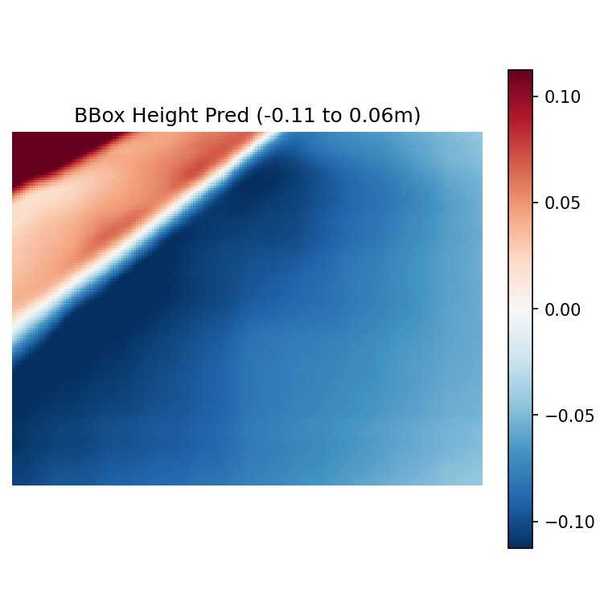} \hfill
        \includegraphics[width=0.2\linewidth]{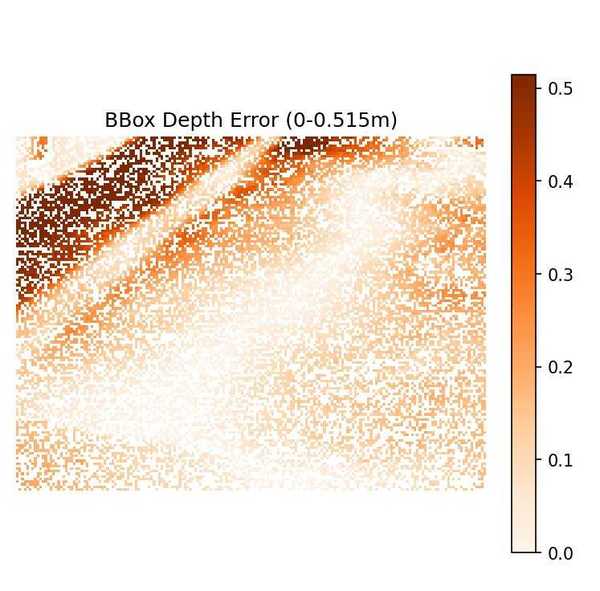} \hfill
        \includegraphics[width=0.2\linewidth]{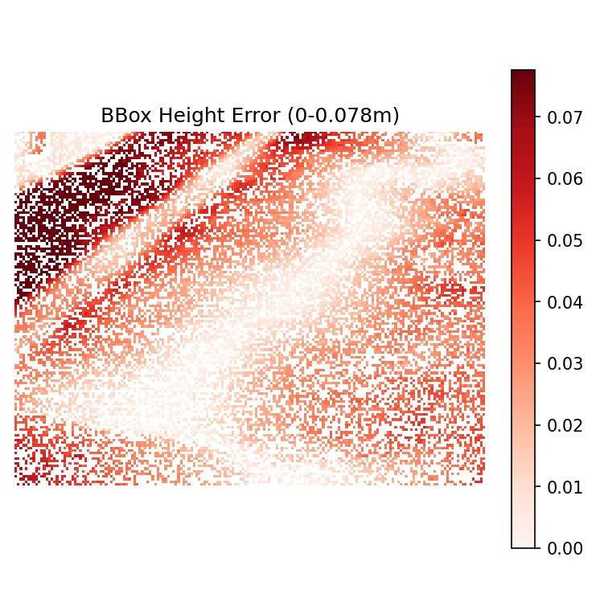}
    \end{minipage}

    % --- UniDepth ---
    % CHANGE -15pt HERE TO MOVE UP/DOWN
    \raisebox{-15pt}{\rotatebox{90}{\tiny \textbf{UniDepthV2}}} \hspace{2pt}
    \begin{minipage}{0.97\linewidth}
        \includegraphics[width=0.2\linewidth]{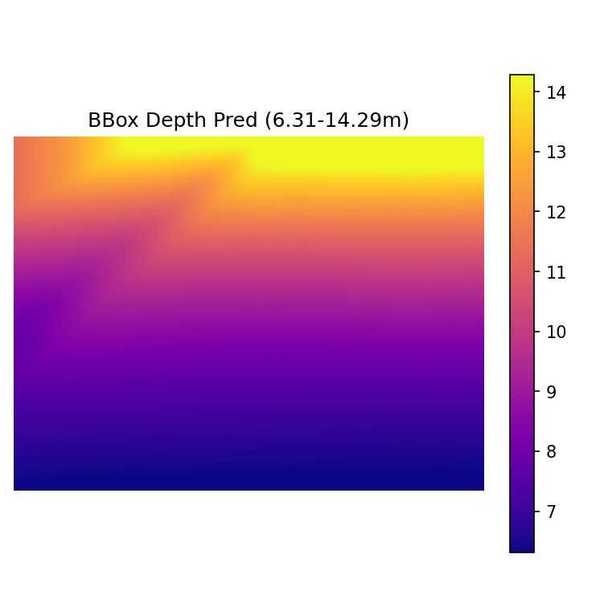} \hfill
        \includegraphics[width=0.2\linewidth]{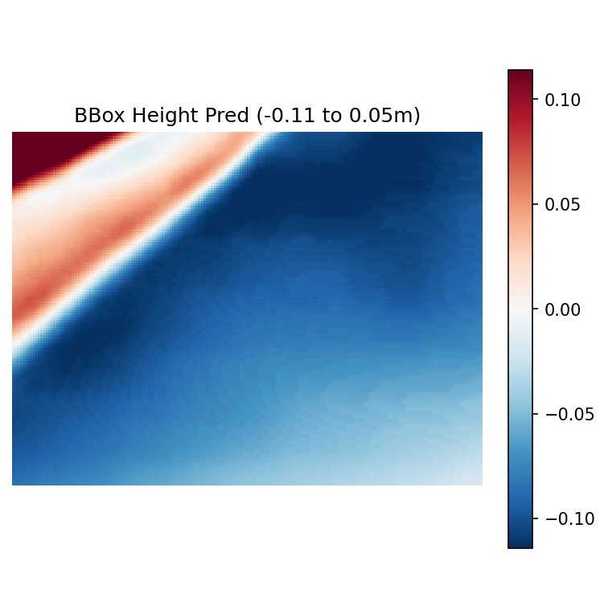} \hfill
        \includegraphics[width=0.2\linewidth]{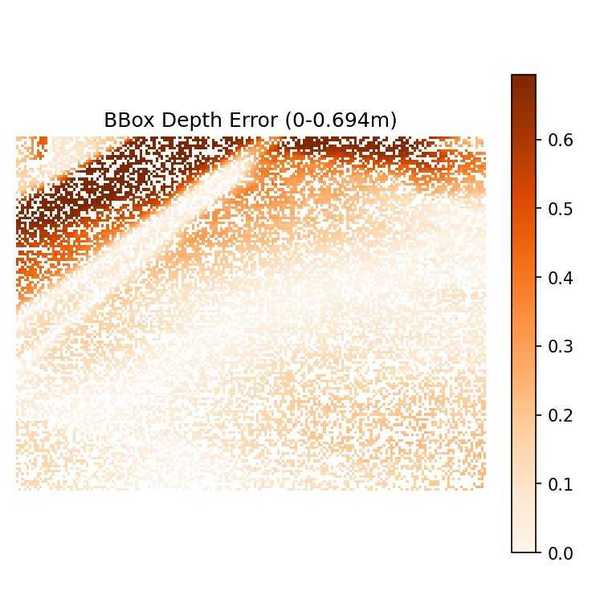} \hfill
        \includegraphics[width=0.2\linewidth]{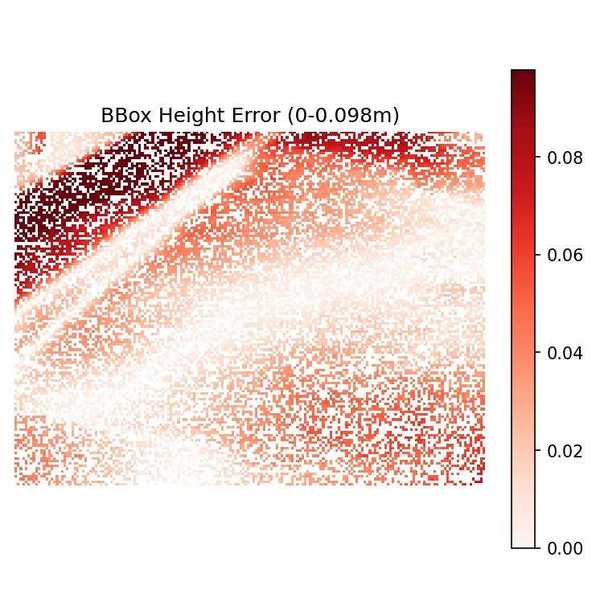}
    \end{minipage}
    
    % --- STEREO ---
    % CHANGE -15pt HERE TO MOVE UP/DOWN
    \raisebox{-15pt}{\rotatebox{90}{\tiny \textbf{FoundationStereo}}} \hspace{2pt}
    \begin{minipage}{0.97\linewidth}
        \includegraphics[width=0.2\linewidth]{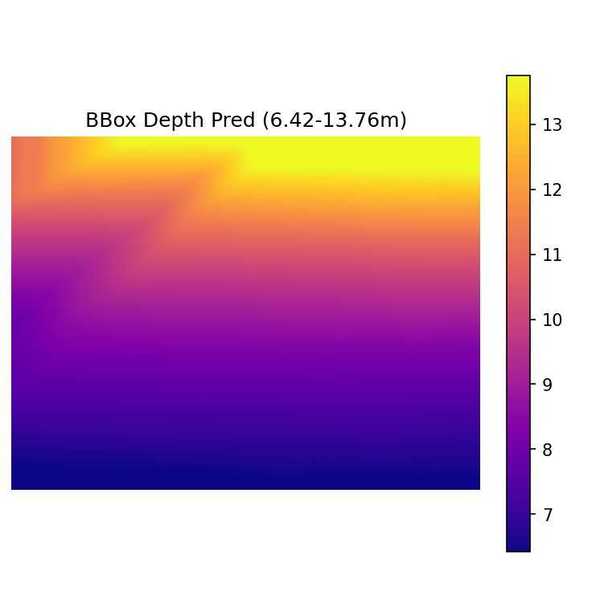} \hfill
        \includegraphics[width=0.2\linewidth]{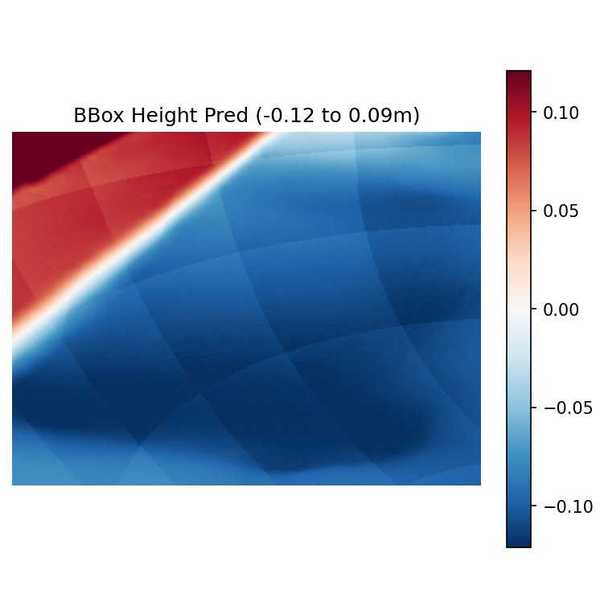} \hfill
        \includegraphics[width=0.2\linewidth]{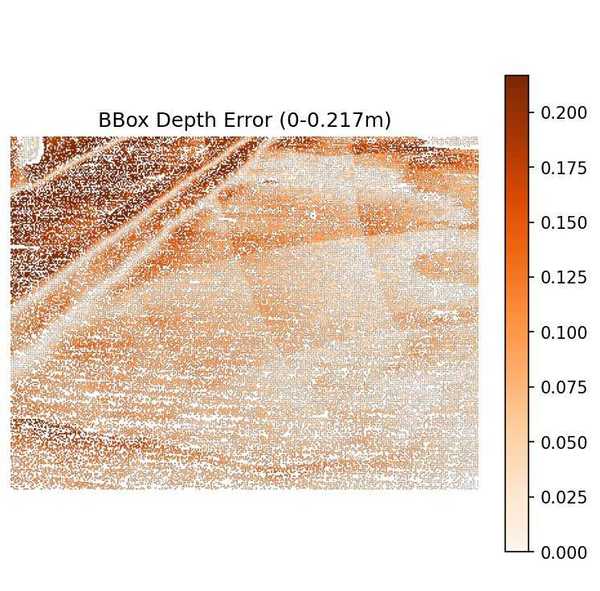} \hfill
        \includegraphics[width=0.2\linewidth]{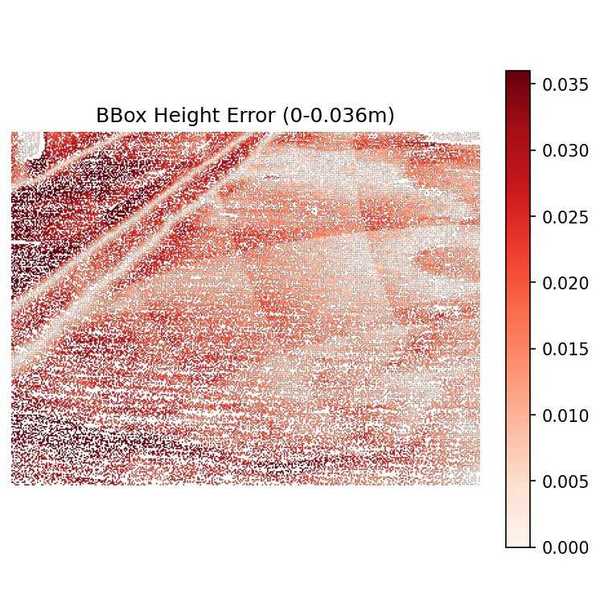}
    \end{minipage}

    \caption{
    \textbf{Qualitative Comparison of a pothole example.} 
    Top: Input context and GT geometry. Bottom: Model predictions.
    Monocular baselines~\cite{Metric3D_v2,MoGE_V2,UniDepth2025_V2} (Rows 1--3). FoundationStereo~\cite{foundationstereo_2025} (Bottom Row) accurately recovers the geometry.
    }
    \label{fig:qual_sample_10675}
\end{figure*}

\begin{figure*}[t!]
    \centering
    \setlength{\tabcolsep}{0pt} 
    \setlength{\abovecaptionskip}{4pt} 
    \setlength{\belowcaptionskip}{-5pt} 
    
    % ==========================================
    % SECTION A: CONTEXT (Distributed Full Width)
    % ==========================================
    \textbf{Positive Road Irregularity} \\
    
    % We use 0.97\linewidth to match the width of the rows below
    \begin{minipage}{0.97\linewidth} 
        \centering
        % 3 images spread evenly
        \includegraphics[width=0.2\linewidth,trim=0 0 0 50,clip]{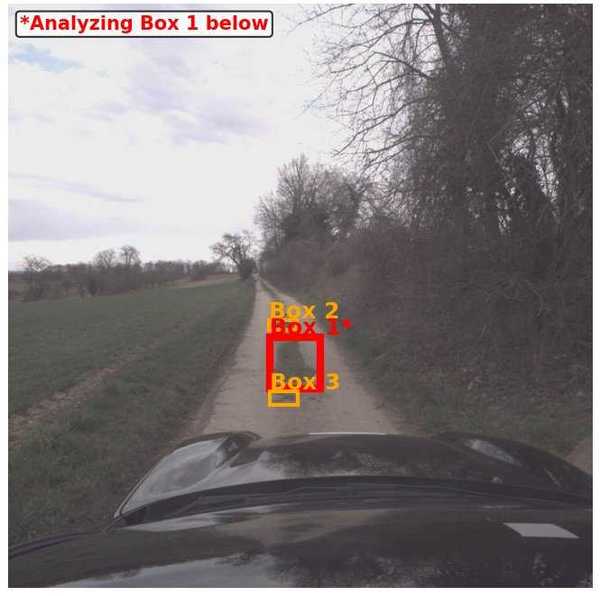} \hfill
        \includegraphics[width=0.2\linewidth]{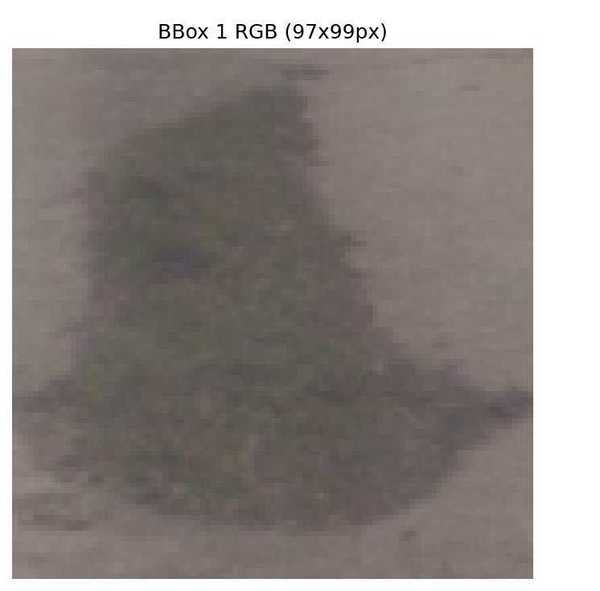} \hfill
        \includegraphics[width=0.2\linewidth]{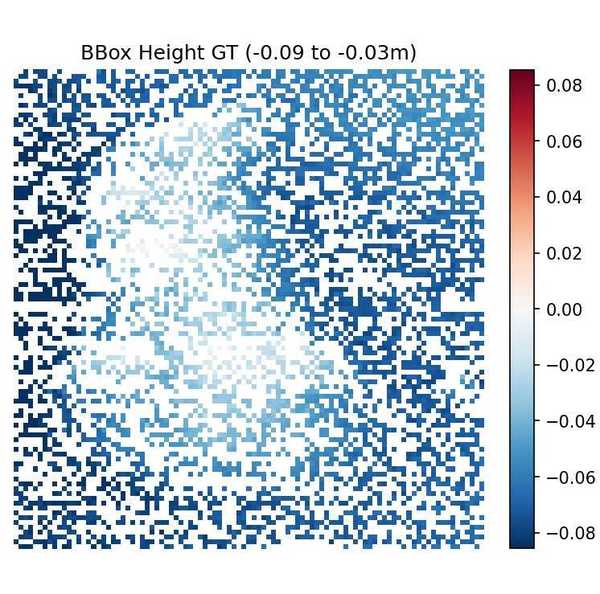} \\
        % Captions spread evenly
        \footnotesize \makebox[0.2\linewidth]{(a) Input RGB} \hfill 
        \makebox[0.2\linewidth]{(b) Box Crop} \hfill 
        \makebox[0.2\linewidth]{(c) GT Height}
    \end{minipage}

    \vspace{1pt} \hrule \vspace{4pt}

    % ==========================================
    % HEADERS
    % ==========================================
    \begin{minipage}{\linewidth}
        \centering \footnotesize \bfseries
        \makebox[0.24\linewidth]{Predicted Depth} \hfill
        \makebox[0.24\linewidth]{Predicted Height} \hfill
        \makebox[0.24\linewidth]{Depth Error} \hfill
        \makebox[0.24\linewidth]{Height Error}
    \end{minipage}
    
    % ==========================================
    % THE GRID (Metric3D, DepthPro, UniDepth, FS)
    % ==========================================

    % --- Metric3D ---
    % CHANGE -15pt HERE TO MOVE UP/DOWN
    \raisebox{-15pt}{\rotatebox{90}{\tiny \textbf{Metric3DV2}}} \hspace{2pt}
    \begin{minipage}{0.97\linewidth}
        \includegraphics[width=0.195\linewidth]{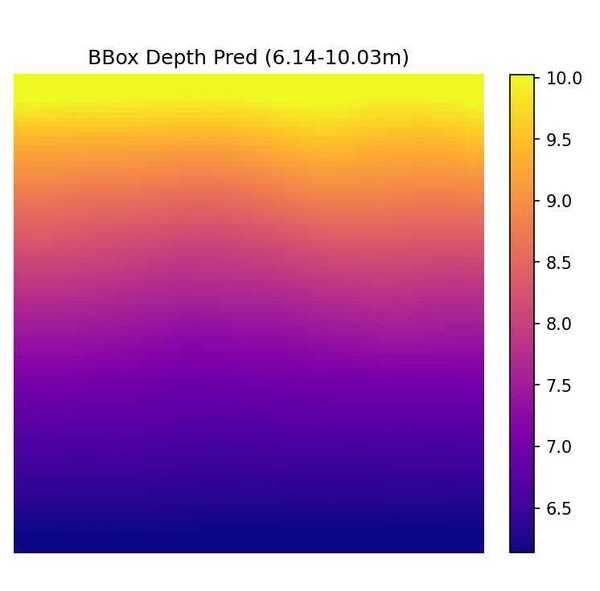} \hfill
        \includegraphics[width=0.195\linewidth]{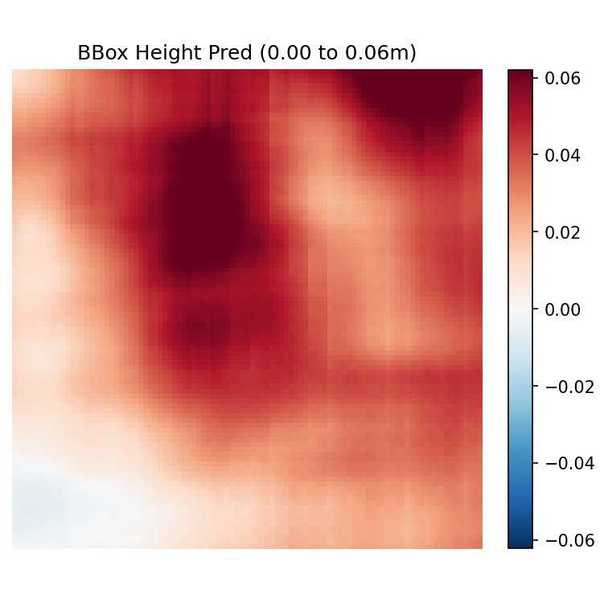} \hfill
        \includegraphics[width=0.195\linewidth]{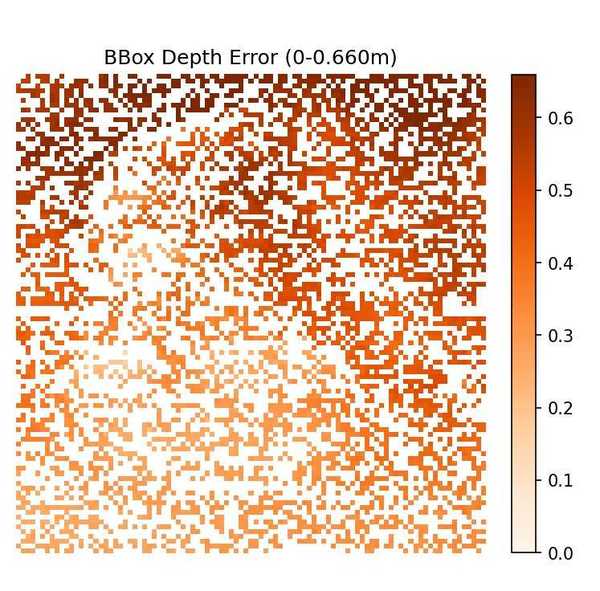} \hfill
        \includegraphics[width=0.195\linewidth]{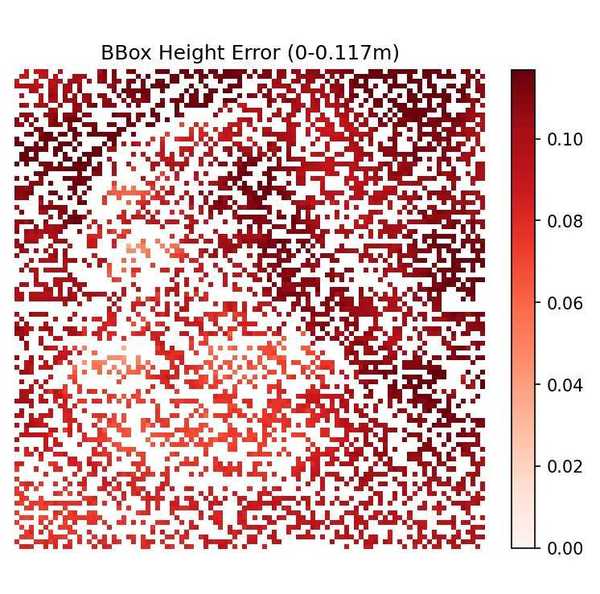}
    \end{minipage}

    % --- DepthPro (REPLACED MoGE) ---
    % CHANGE -15pt HERE TO MOVE UP/DOWN
    \raisebox{-15pt}{\rotatebox{90}{\tiny \textbf{DepthPro}}} \hspace{2pt}
    \begin{minipage}{0.97\linewidth}
        \includegraphics[width=0.195\linewidth]{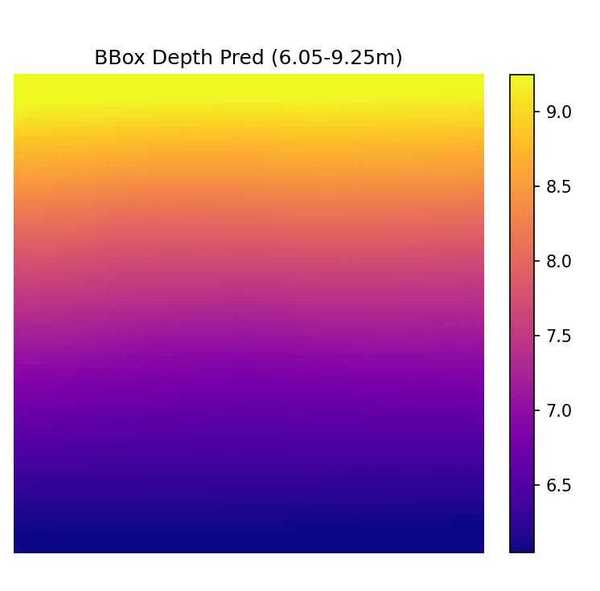} \hfill
        \includegraphics[width=0.195\linewidth]{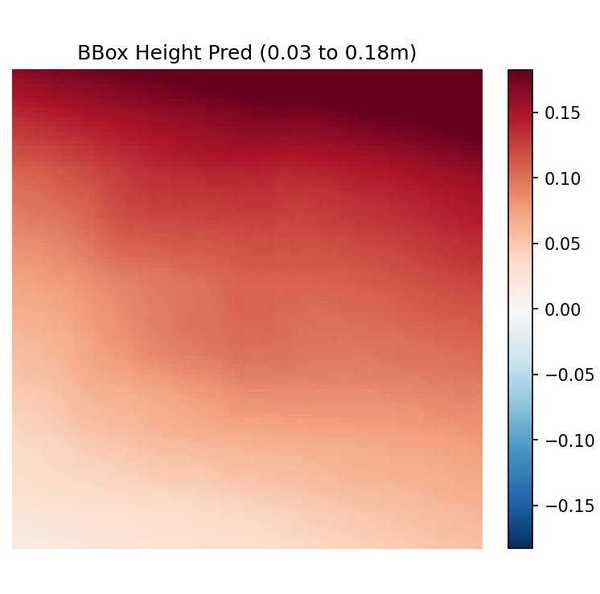} \hfill
        \includegraphics[width=0.195\linewidth]{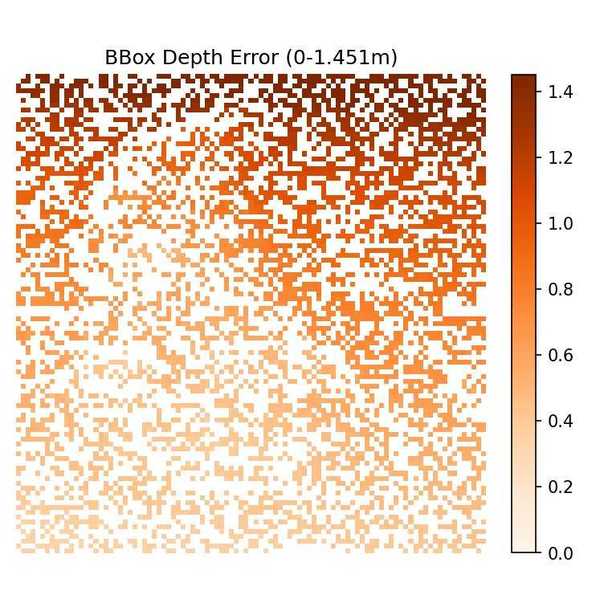} \hfill
        \includegraphics[width=0.195\linewidth]{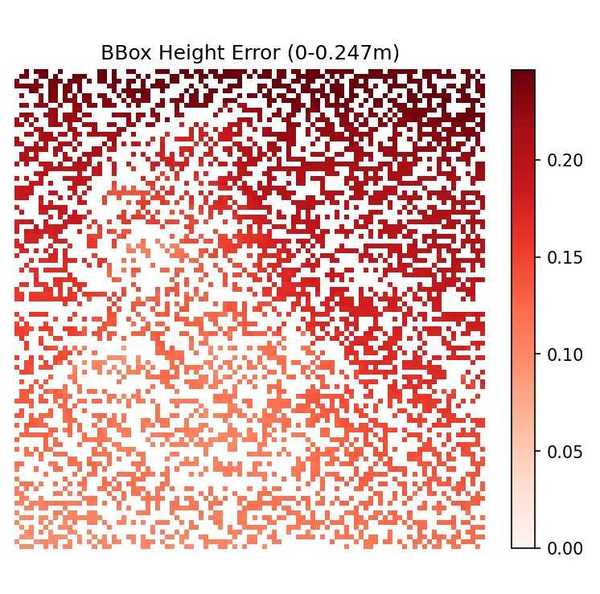}
    \end{minipage}

    % --- UniDepth ---
    % CHANGE -15pt HERE TO MOVE UP/DOWN
    \raisebox{-15pt}{\rotatebox{90}{\tiny \textbf{UniDepthV2}}} \hspace{2pt}
    \begin{minipage}{0.97\linewidth}
        \includegraphics[width=0.195\linewidth]{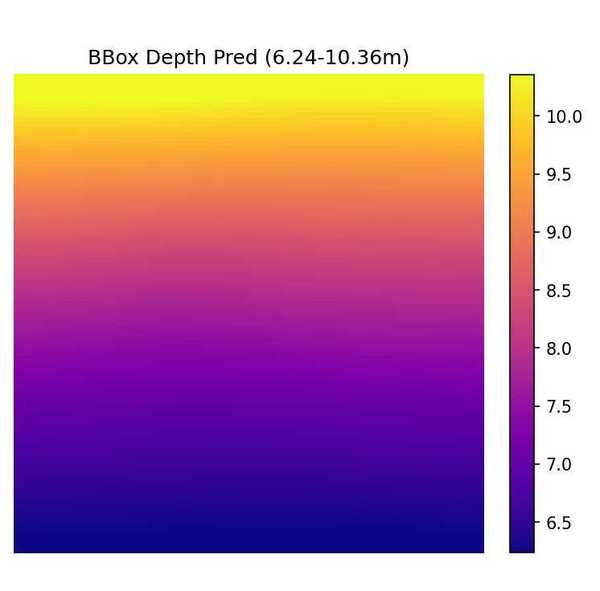} \hfill
        \includegraphics[width=0.195\linewidth]{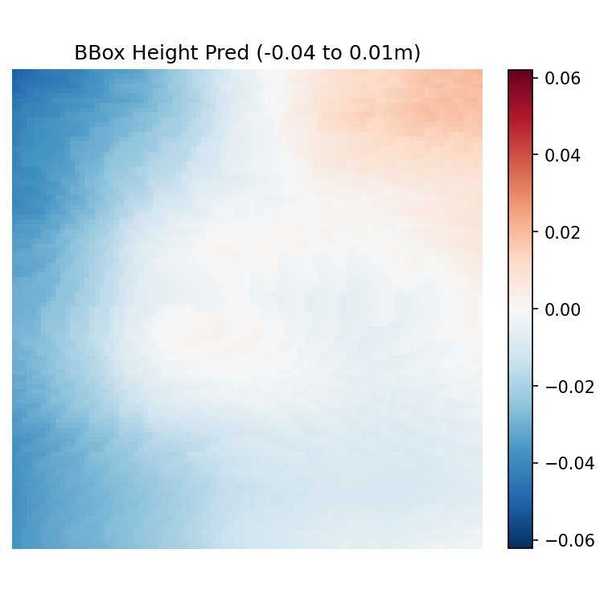} \hfill
        \includegraphics[width=0.195\linewidth]{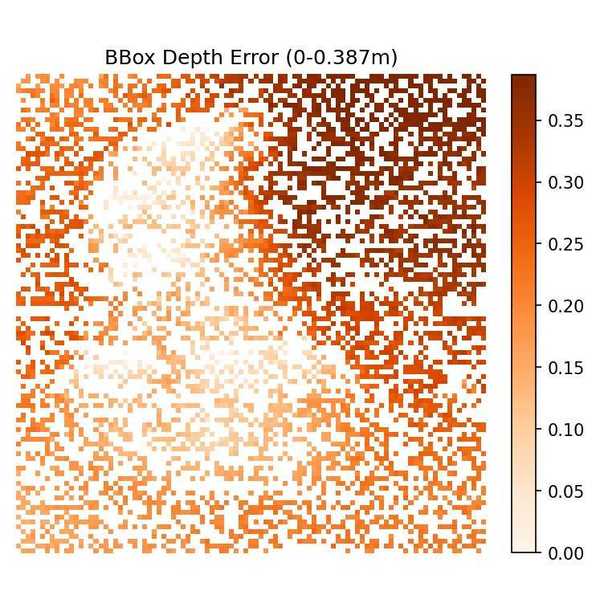} \hfill
        \includegraphics[width=0.195\linewidth]{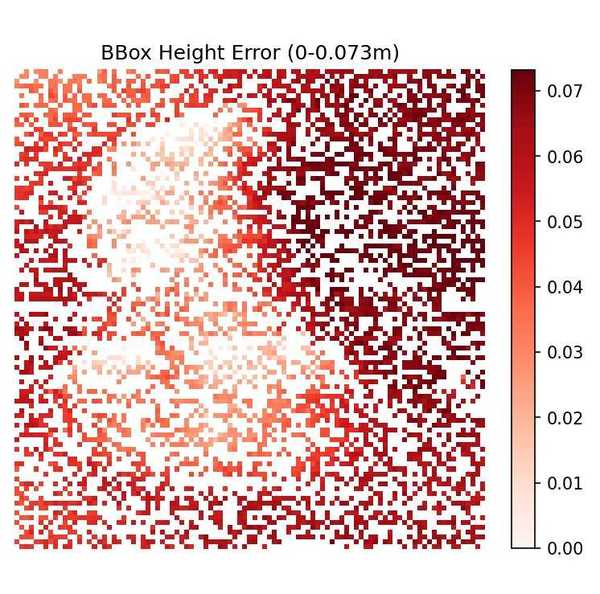}
    \end{minipage}
    
    % --- STEREO ---
    % CHANGE -15pt HERE TO MOVE UP/DOWN
    \raisebox{-15pt}{\rotatebox{90}{\tiny \textbf{FoundationStereo}}} \hspace{2pt}
    \begin{minipage}{0.97\linewidth}
        \includegraphics[width=0.195\linewidth]{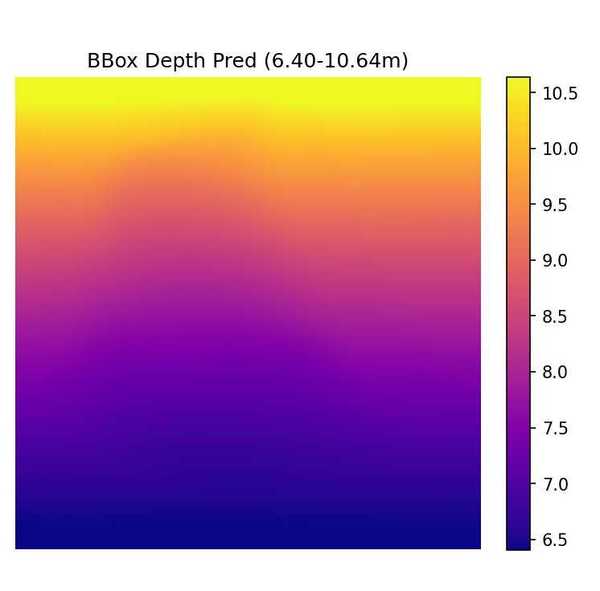} \hfill
        \includegraphics[width=0.195\linewidth]{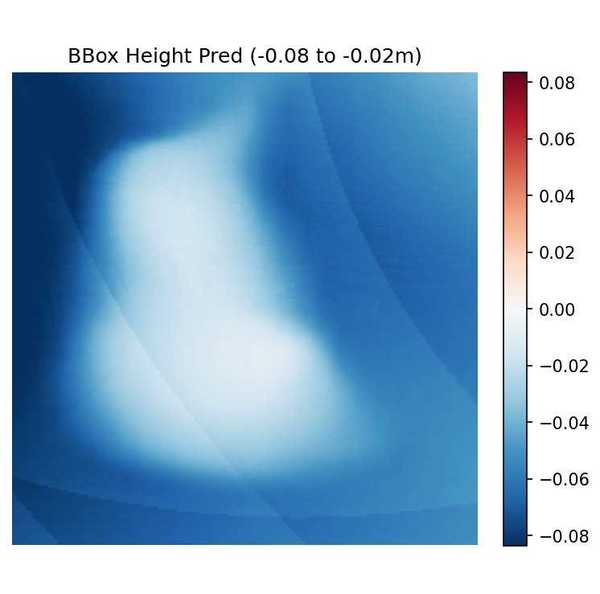} \hfill
        \includegraphics[width=0.195\linewidth]{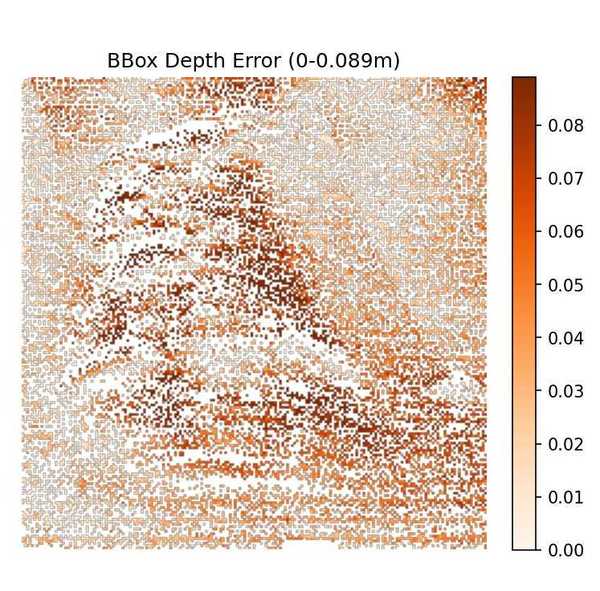} \hfill
        \includegraphics[width=0.195\linewidth]{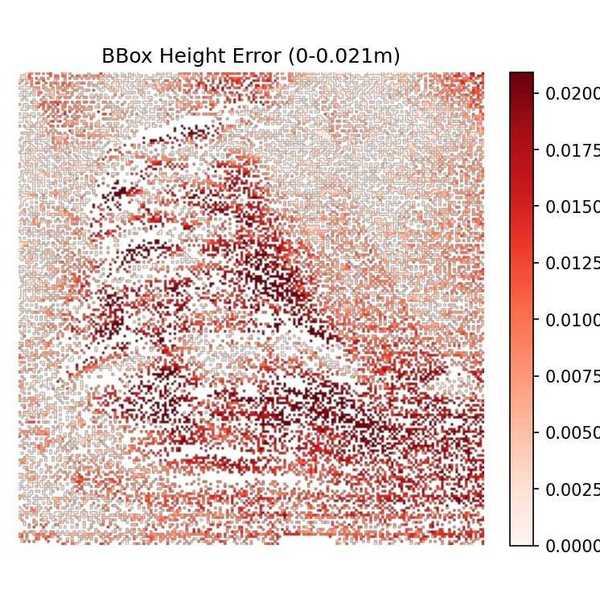}
    \end{minipage}

    \caption{
    \textbf{Qualitative Comparison of a positive road irregularity.} 
    Top: Input context and GT geometry. Bottom: Model predictions.
    Monocular baselines~\cite{Metric3D_v2,DepthPro,UniDepth2025_V2} (Rows 1--3). FoundationStereo~\cite{foundationstereo_2025} (Bottom Row).
    }
    \label{fig:qual_sample_3376}
\end{figure*}

\begin{figure*}[t!]
    \centering
    \setlength{\tabcolsep}{0pt} 
    \setlength{\abovecaptionskip}{4pt} 
    \setlength{\belowcaptionskip}{-5pt} 
    
    % ==========================================
    % SECTION A: CONTEXT (Distributed Full Width)
    % ==========================================
    \textbf{Positive Road Irregularity} \\
    
    % We use 0.97\linewidth to match the width of the rows below
    \begin{minipage}{0.97\linewidth} 
        \centering
        % 3 images spread evenly
        \includegraphics[width=0.2\linewidth]{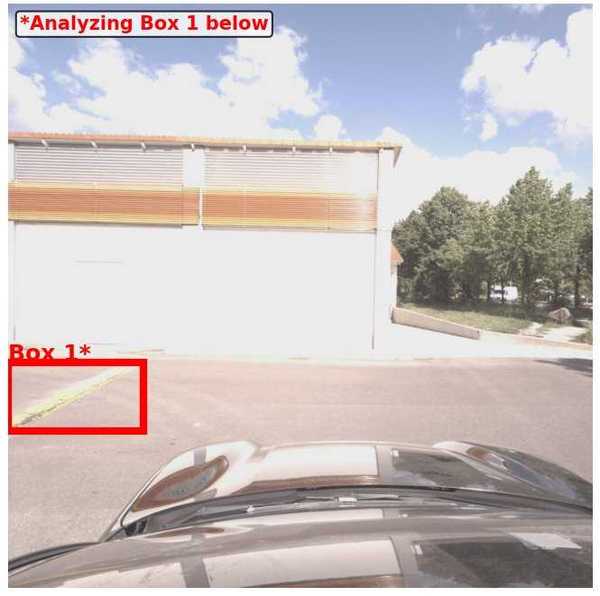} \hfill
        \includegraphics[width=0.2\linewidth]{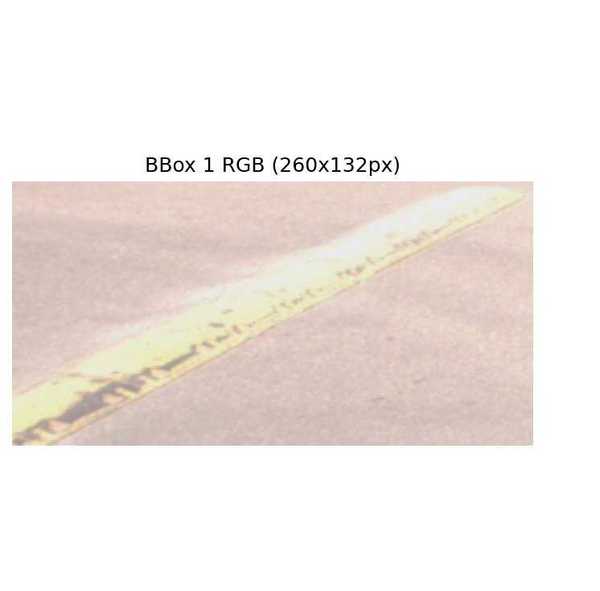} \hfill
        \includegraphics[width=0.2\linewidth]{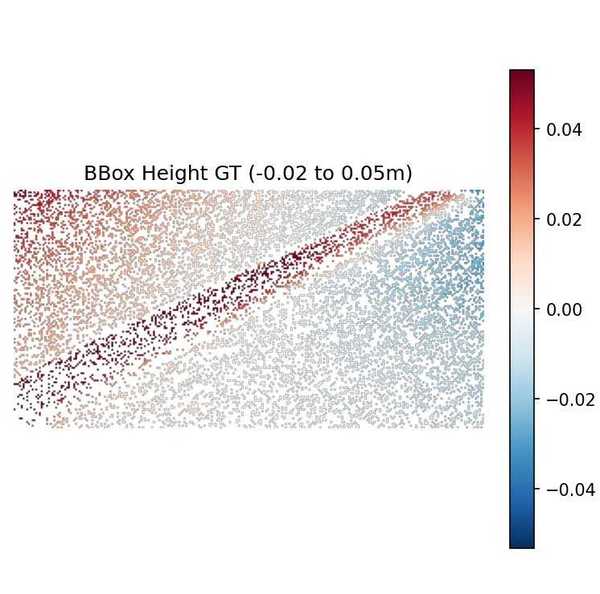} \\
        % Captions spread evenly
        \footnotesize \makebox[0.2\linewidth]{(a) Input RGB} \hfill 
        \makebox[0.2\linewidth]{(b) Box Crop} \hfill 
        \makebox[0.2\linewidth]{(c) GT Height}
    \end{minipage}

    \vspace{1pt} \hrule \vspace{4pt}

    % ==========================================
    % HEADERS
    % ==========================================
    \begin{minipage}{\linewidth}
        \centering \footnotesize \bfseries
        \makebox[0.24\linewidth]{Predicted Depth} \hfill
        \makebox[0.24\linewidth]{Predicted Height} \hfill
        \makebox[0.24\linewidth]{Depth Error} \hfill
        \makebox[0.24\linewidth]{Height Error}
    \end{minipage}
    
    % ==========================================
    % THE GRID (Metric3D, DepthAnythingV2, UniDepth, FS)
    % ==========================================

    % --- Metric3D ---
    % CHANGE -15pt HERE TO MOVE UP/DOWN
    \raisebox{-15pt}{\rotatebox{90}{\tiny \textbf{Metric3DV2}}} \hspace{2pt}
    \begin{minipage}{0.97\linewidth}
        \includegraphics[width=0.195\linewidth]{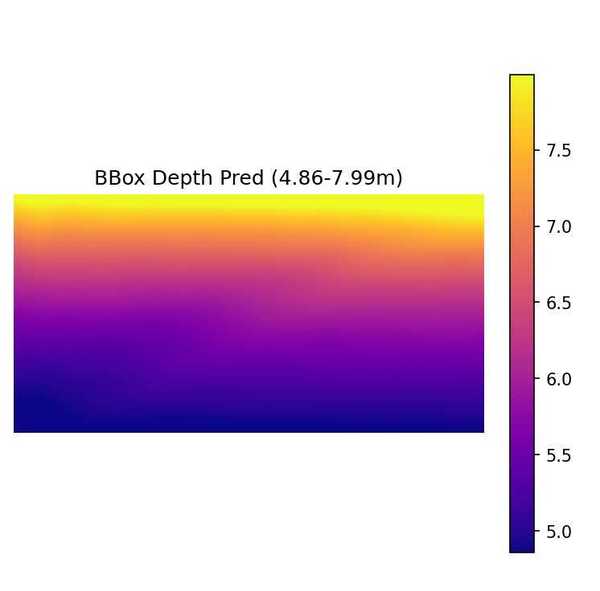} \hfill
        \includegraphics[width=0.195\linewidth]{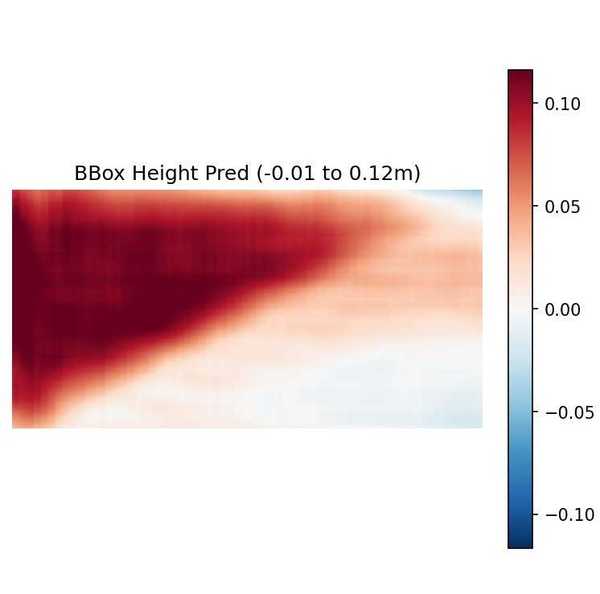} \hfill
        \includegraphics[width=0.195\linewidth]{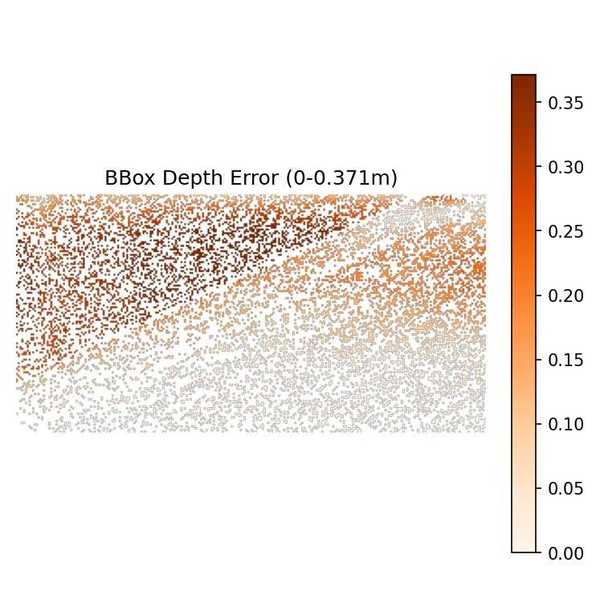} \hfill
        \includegraphics[width=0.195\linewidth]{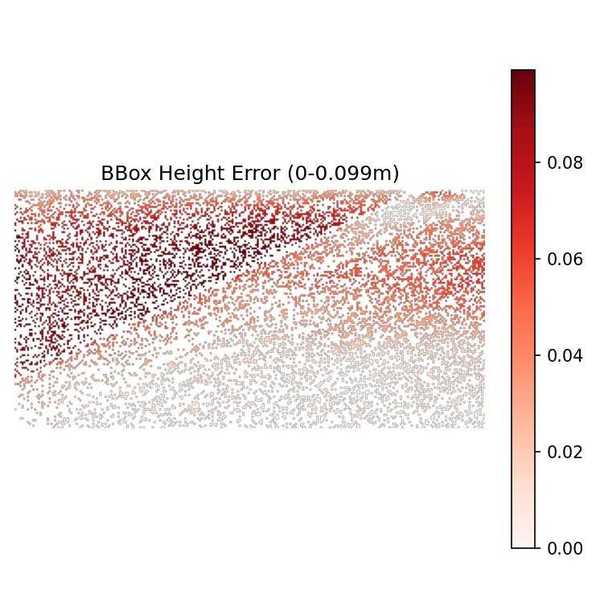}
    \end{minipage}

    % --- DepthAnything V2 (REPLACED DepthPro) ---
    % CHANGE -15pt HERE TO MOVE UP/DOWN
    \raisebox{-15pt}{\rotatebox{90}{\tiny \textbf{DepthAnythingV2}}} \hspace{2pt}
    \begin{minipage}{0.97\linewidth}
        \includegraphics[width=0.195\linewidth]{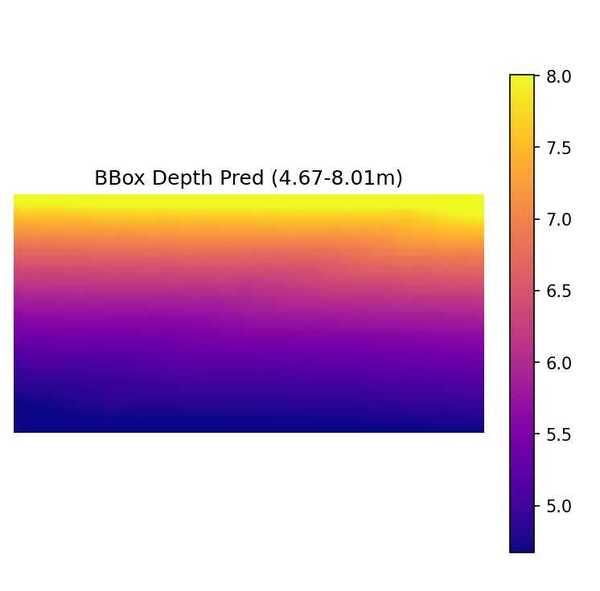} \hfill
        \includegraphics[width=0.195\linewidth]{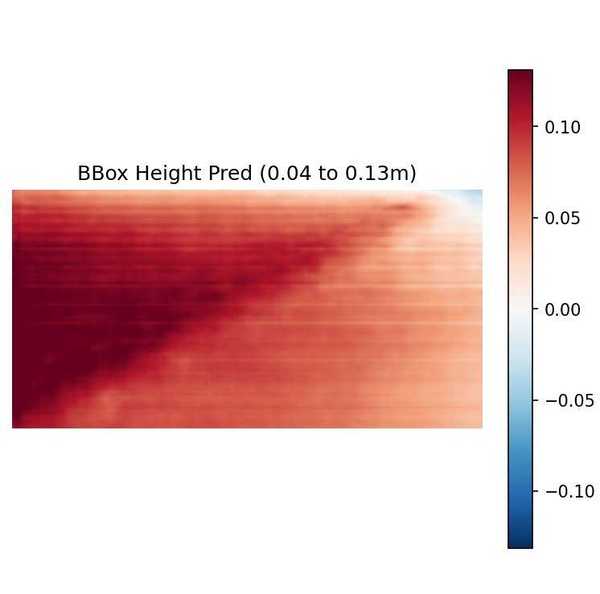} \hfill
        \includegraphics[width=0.195\linewidth]{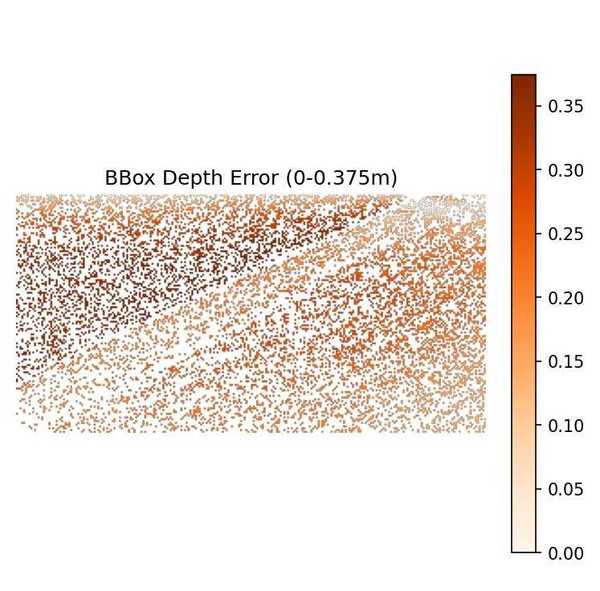} \hfill
        \includegraphics[width=0.195\linewidth]{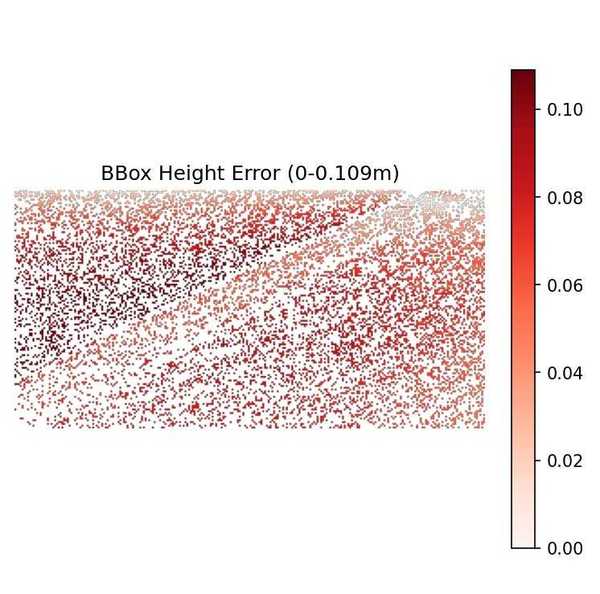}
    \end{minipage}

    % --- UniDepth ---
    % CHANGE -15pt HERE TO MOVE UP/DOWN
    \raisebox{-15pt}{\rotatebox{90}{\tiny \textbf{UniDepthV2}}} \hspace{2pt}
    \begin{minipage}{0.97\linewidth}
        \includegraphics[width=0.195\linewidth]{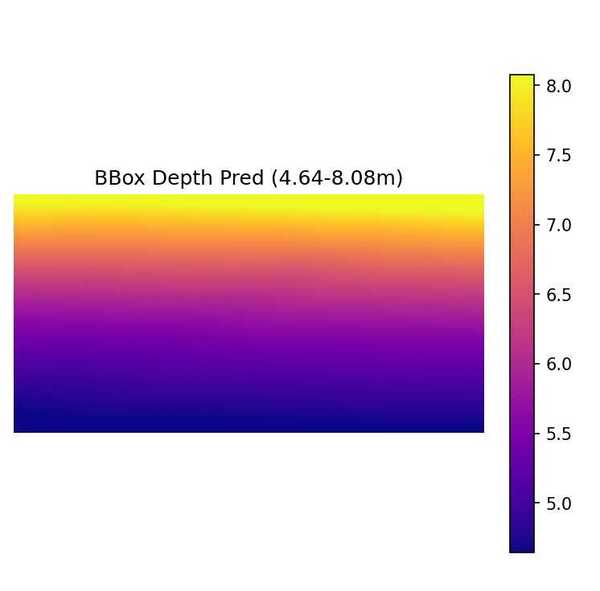} \hfill
        \includegraphics[width=0.195\linewidth]{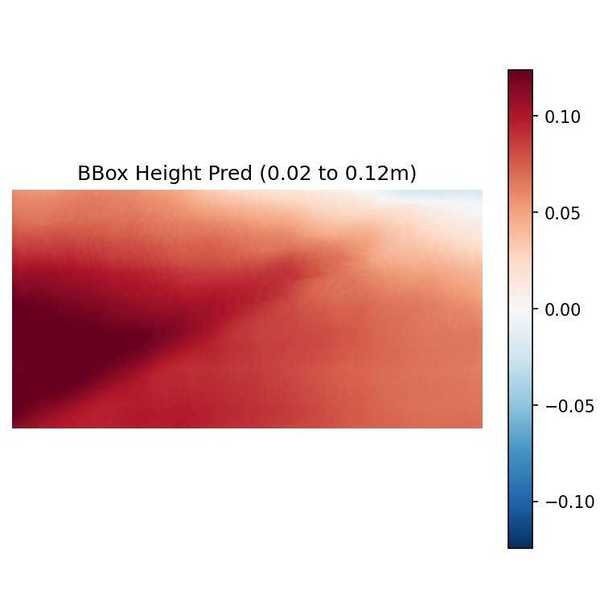} \hfill
        \includegraphics[width=0.195\linewidth]{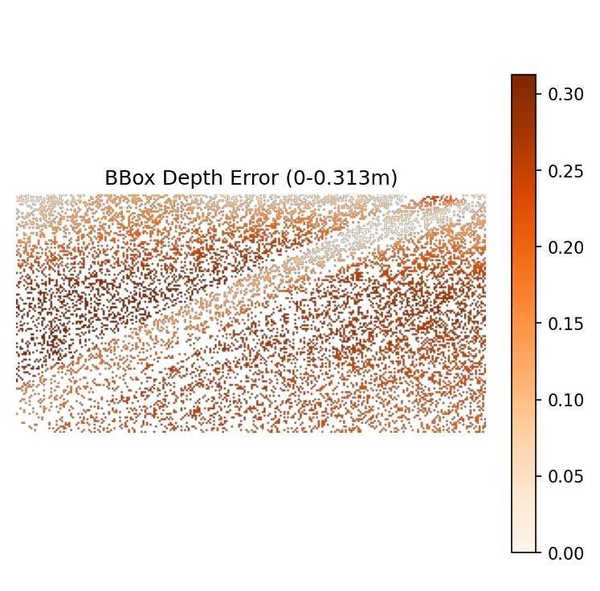} \hfill
        \includegraphics[width=0.195\linewidth]{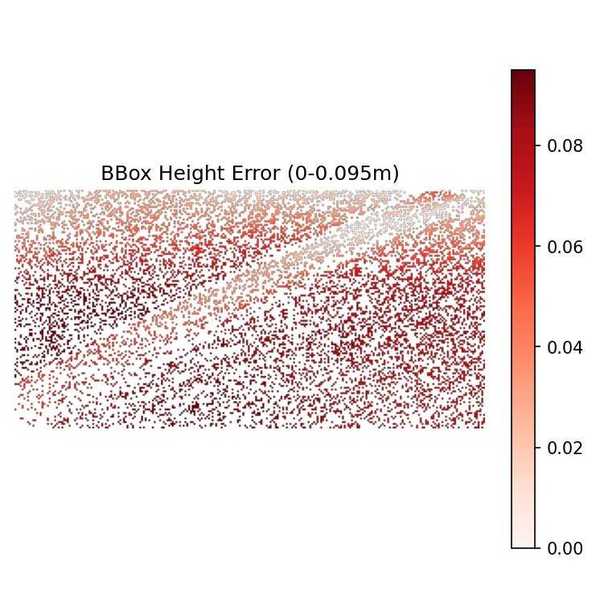}
    \end{minipage}
    
    % --- STEREO ---
    % CHANGE -15pt HERE TO MOVE UP/DOWN
    \raisebox{-15pt}{\rotatebox{90}{\tiny \textbf{FoundationStereo}}} \hspace{2pt}
    \begin{minipage}{0.97\linewidth}
        \includegraphics[width=0.195\linewidth]{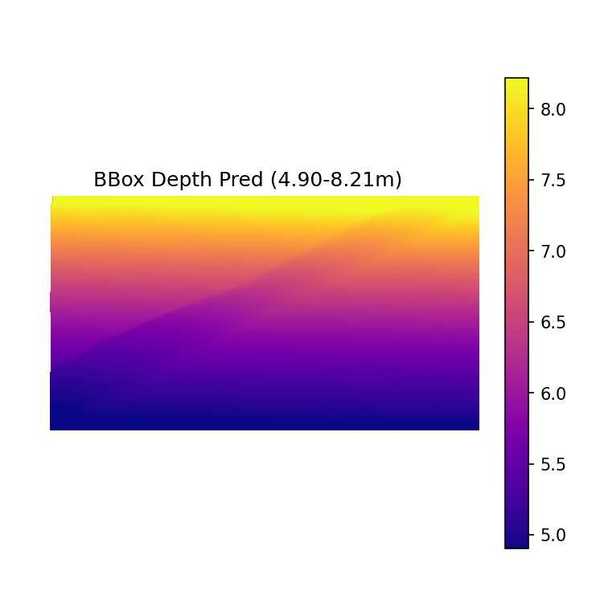} \hfill
        \includegraphics[width=0.195\linewidth]{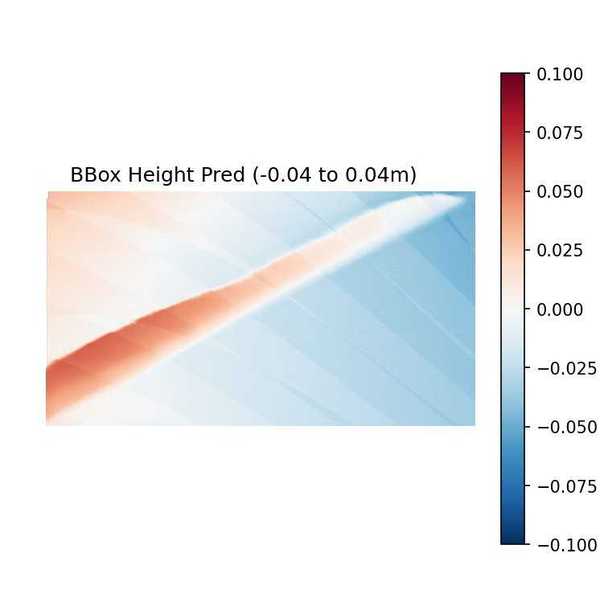} \hfill
        \includegraphics[width=0.195\linewidth]{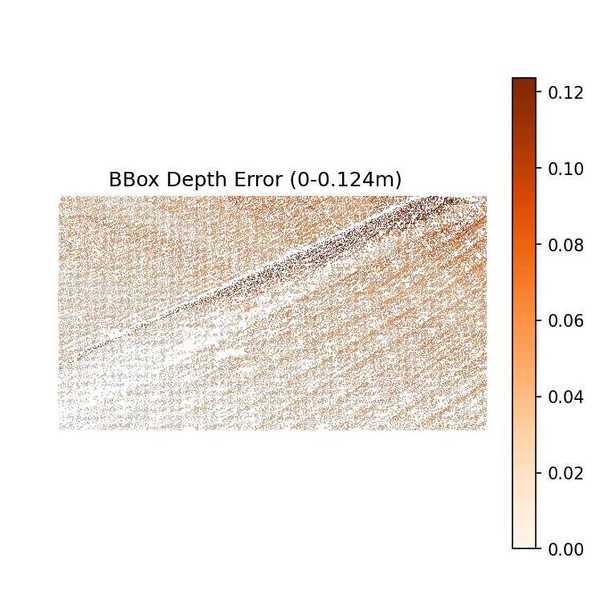} \hfill
        \includegraphics[width=0.195\linewidth]{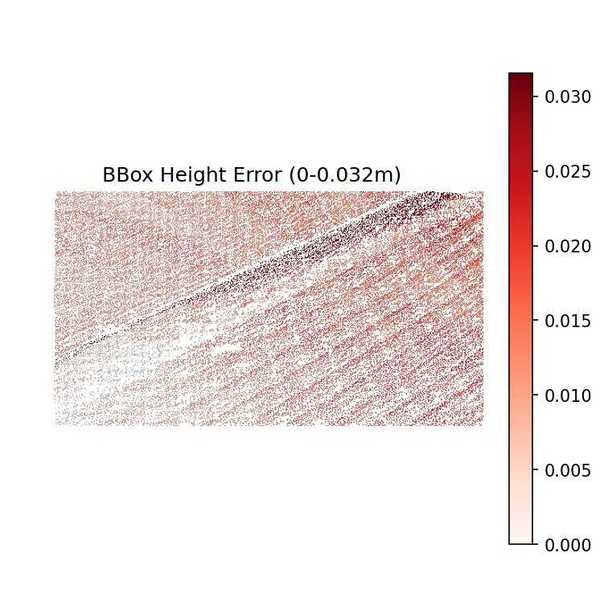}
    \end{minipage}

    \caption{
    {Qualitative Comparison of a positive road irregularity.} 
    Top: Input context and GT geometry. Bottom: Model predictions.
    Monocular baselines~\cite{Metric3D_v2,DepthAnythingV2_2025,UniDepth2025_V2} (Rows 1--3). FoundationStereo~\cite{foundationstereo_2025} (Bottom Row).
    }
    \label{fig:qual_sample_6327}
\end{figure*}

Additionally, \cref{fig:qual_sample_5777,fig:qual_sample_10675,fig:qual_sample_3376,fig:qual_sample_6327} present specific qualitative comparisons of surface reconstruction. In these examples, we visualize the input context, ground truth height, and predictions from monocular baselines alongside the stereo reference. The error maps indicate that monocular estimators often oversmooth high-frequency geometry, leading to height estimation errors on potholes and bumps, whereas the stereo baseline better preserves these structural details.

